\documentclass[final]{cvpr}

\usepackage{times}
\usepackage{epsfig}
\usepackage{graphicx}
\usepackage{amsmath}
\usepackage{amssymb}
\usepackage[caption=false]{subfig}
\usepackage{appendix}

\usepackage{cite}
\usepackage[pagebackref=true,breaklinks=true,colorlinks,bookmarks=false]{hyperref}

\newcommand{\PAR}[1]{\vskip4pt \noindent{\bf #1~}}

\DeclareMathOperator*{\argmax}{argmax}
\makeatother

\def\cvprPaperID{2057} % *** Enter the CVPR Paper ID here
\def\confYear{CVPR 2021}

\begin{document}

%%%%%%%%% TITLE
\title{How Privacy-Preserving are Line Clouds?\\ Recovering Scene Details from 3D Lines}

\author{Kunal Chelani$^1$ \quad Fredrik Kahl$^1$ \quad Torsten Sattler$^{1,2}$\\
$^1$Chalmers University of Technology \quad $^2$Czech Technical University in Prague}

% \author{Kunal Chelani\\
% Chalmers Institute of Technology\\
% % For a paper whose authors are all at the same institution,
% % omit the following lines up until the closing ``}''.
% % Additional authors and addresses can be added with ``\and'',
% % just like the second author.
% % To save space, use either the email address or home page, not both
% \and
% Fredrik Kahl\\
% Chalmers Institute of Technology\\
% \and
% Torsten Sattler\\
% Chalmers Institute of Technology\\
% Czech Technical University\\
% {\tt\small \{chelani, fredrik.kahl, torsat\}@chalmers.se}
% }
\maketitle
\thispagestyle{empty}

%%%%%%%%% ABSTRACT
\begin{abstract}
\vspace{-6pt}
Visual localization is the problem of estimating the camera pose of a given image with respect to a known scene. Visual localization algorithms are a fundamental building block in advanced computer vision applications, including Mixed and Virtual Reality systems. Many algorithms used in practice represent the scene through a Structure-from-Motion (SfM) point cloud and use 2D-3D matches between a query image and the 3D points for camera pose estimation. As recently shown, image details can be accurately recovered from SfM point clouds by translating renderings of the sparse point clouds to images. To address the resulting potential privacy risks for user-generated content, it was recently proposed to lift point clouds to line clouds by replacing 3D points by randomly oriented 3D lines passing through these points. The resulting representation is unintelligible to humans and effectively prevents point cloud-to-image translation. This paper shows that a significant amount of information about the 3D scene geometry is preserved in these line clouds, allowing us to (approximately) recover the 3D point positions and thus to (approximately) recover image content. Our approach is based on the observation that the closest points between lines can yield a good approximation to the original 3D points. Code is available at \href{https://github.com/kunalchelani/Line2Point}{https://github.com/kunalchelani/Line2Point}.
\end{abstract}

\begin{figure*}[!t]
    \centering
    \includegraphics[width=1\linewidth]{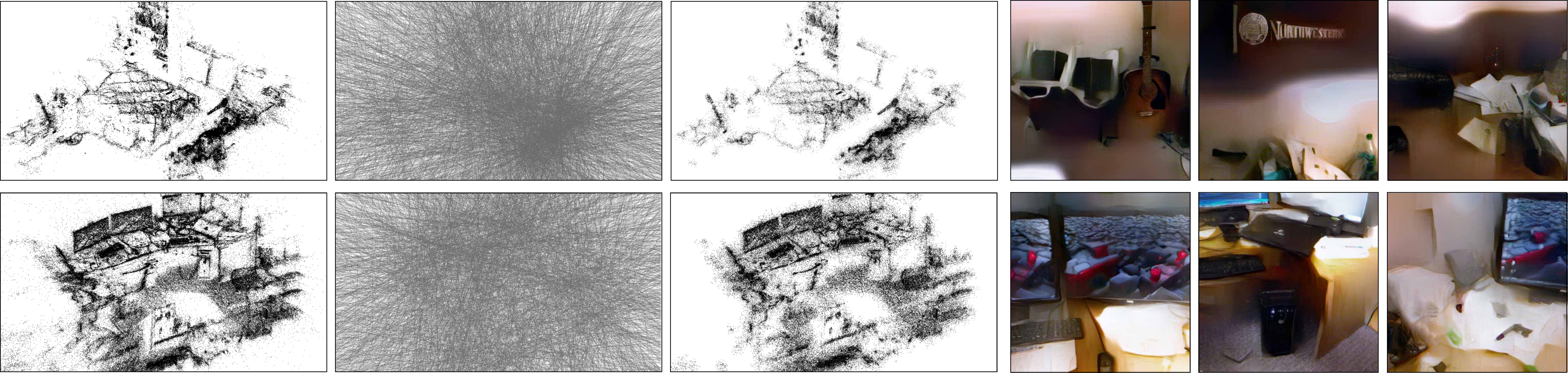}
    \caption{In order to preserve privacy, \cite{Speciale2019CVPR} proposed to store line clouds instead of point clouds for visual localization (left). While unintelligible to the human eye, we show that it is possible to recover the underlying 3D point clouds (middle). Applying a point cloud-to-image translation approach~\cite{Pittaluga2019CVPR} then allows us to recover image details (right), showing that lifting point clouds to line clouds can still preserve privacy critical information that can later be extracted from the line clouds.} 
    \label{fig:teaser}
\end{figure*}

\vspace{-12pt}
\section{Introduction}
Visual localization is the problem of estimating the position and orientation from which an image %or a sequence of images 
was taken in a known scene. 
Visual localization is a fundamental part of computer vision systems such as self-driving cars~\cite{Schwesinger2016IV,Heng2019ICRA}, Augmented and Mixed Reality applications~\cite{Castle08ISWC,Arth09ISMAR}, Structure-from-Motion (SfM)~\cite{Schoenberger2016CVPR,kahl-hartley-pami-2008,Snavely08IJCV,Heinly2015CVPR}, and Simultaneous Localization and Mapping (SLAM)~\cite{Mur2017TRO,SalasMoreno2013CVPR,Cummins08IJRR}.

Classical approaches to visual localization~\cite{Li2010ECCV,Li2012ECCV,Lim12CVPR,Irschara09CVPR,Sattler2011ICCV,Sattler2017CVPR,Zeisl2015ICCV,Choudhary12ECCV,Robertson04BMVC,Zhang06TDPVT} are based on local features such as SIFT~\cite{Lowe04IJCV}. 
They use SfM to construct a sparse 3D point cloud of the scene, where each point is associated with the local image features it was triangulated from. 
Descriptor matching between local features extracted in a test / query image and the 3D points then yields a set of 2D-3D matches that can be used for RANSAC-based camera pose estimation~\cite{Fischler81CACM,Chum08PAMI,Lebeda2012BMVC,Kukelova13ICCV,Bujnak08CVPR,Kukelova2016CVPR,Larsson2019ICCV}. 
To the best of our knowledge, visual localization approaches used in practice by companies such as Google~\cite{Lynen2019ARXIV,GoogleVPS}, Microsoft~\cite{Speciale2019CVPR,Speciale2019ICCV,MicrosoftSpatialAnchors}, or Scape Technologies~\cite{ScapeMedium} all follow this classical approach. %feature-based approach. 

Traditionally, work on visual localization has focused on accurate and scalable algorithms able to cover large areas~\cite{Li2012ECCV,Zeisl2015ICCV,Chen2011CVPR,Svarm2017PAMI,Sarlin2019CVPR,Irschara09CVPR,Toft2021PAMI} or to run in real-time on mobile devices with limited memory and compute capabilities~\cite{Lim12CVPR,Arth09ISMAR,Middelberg2014ECCV,Lynen2015RSS,Lynen2019ARXIV}. 
Thus, the underlying scene representations have been designed to enable efficient 2D-3D matching~\cite{Irschara09CVPR,Sattler2012BMVC,Sarlin2019CVPR,Sattler2015ICCV,Lynen2019ARXIV} and / or to limit memory consumption~\cite{Cao2014CVPR,Li2010ECCV,Camposeco2019CVPR,Lynen2015RSS,Lynen2019ARXIV,Sattler2015ICCV}. 
Privacy aspects such as avoiding user generated content from being recovered either through 3D models stored in the cloud or through query images sent to a server have traditionally not been taken into account.

Recently, \cite{Pittaluga2019CVPR,Song2020ECCV} showed that it is possible to recover images from SfM point clouds. 
Given a rendering of the point cloud (and the feature descriptors associated with the 3D points), \cite{Pittaluga2019CVPR} uses a CNN to translate the rendering to a complete image. 
Their work clearly demonstrates that storing SfM point clouds creates potential privacy risks as an attacker could recover details from user-uploaded content stored in the cloud. 
To prevent such attacks, \cite{Speciale2019CVPR} proposed to replace each SfM point through a random line passing through this point (\cf Fig.~\ref{fig:teaser}(left)). 
They showed that the resulting representation is unintelligible to humans, prevents a direct application of \cite{Pittaluga2019CVPR}, and still enables accurate camera pose estimation. 
This idea of lifting points to lines was later adapted for %privacy-preserving SfM~\cite{Geppert2020ECCV} and 
privacy-perserving SLAM~\cite{Shibuya2020ECCV}. 
However, this paper shows that it is possible to (approximately) recover the original 3D point positions from a line cloud (\cf Fig.~\ref{fig:teaser}(middle)), again enabling us to use~\cite{Pittaluga2019CVPR,Song2020ECCV} %the approach from \cite{Pittaluga2019CVPR} 
to obtain images (\cf Fig.~\ref{fig:teaser}(right)). % details. 

In detail, this paper makes the following \textbf{contributions}: 
(\textbf{i}) in the case that the line directions are chosen uniformly at random, as is the case in~\cite{Speciale2019CVPR}, we show that knowledge about local neighborhoods allows us to (approximately) recover the original 3D points based on the closest points between pairs of lines. % provide an approximation to the original point positions. 
(\textbf{ii}) based on this insight, we propose a two-stage approach that first recovers these neighborhoods and then estimates the 3D points corresponding to the input lines. %we propose a coarse-level point recovery approach to obtain initial estimates for the original points. 
% In a next step, we refine our initial estimates. 
(\textbf{iii}) detailed experiments on both indoor and outdoor datasets show that our approach allows us to faithfully reproduce the original point clouds. % using only the geometry of the line cloud. 
In addition, applying~\cite{Pittaluga2019CVPR} on the resulting point clouds enables us to recover image details (\cf Fig.~\ref{fig:teaser}(right)). 
% While taking the associated descriptors into account could further improve results, we show that purely geometric reasoning is already sufficient for the recovery process. 
(\textbf{iv}) while using line clouds alone is not effective in obfuscating the underlying 3D scene geometry, we show that using (very) sparse line clouds effectively prevents our approach from recovering image details. %e analyze the accuracy of our approach and discuss its failure cases. 
Our results clearly show that further research on privacy-preserving scene representations is needed. 

\vspace{-4pt}
\section{Related Work}
% \TODO{Discuss the privacy-preserving SfM and SLAM papers from ECCV 2020 and anything else that we might have forgotten / that was published since the ECCV 2020 deadline}
% \TODO{Target length: 1 page}
Privacy-preserving methods in computer vision are not new. For instance, it is standard to blur faces and license plates to prevent the identification of persons and cars, respectively. Yet, it has been shown that this process is not privacy preserving \cite{oh2016ECCV} as one can still train a recognition system on obfuscated images with reasonable accuracy rates. Privacy preserving image degradation with adversarial training has been proposed for  visual recognition \cite{wu2018ECCV}. In this paper, we focus on privacy preserving localization.

\PAR{Visual localization.} Traditionally, visual localization approaches have relied on local features such as SIFT~\cite{Lowe04IJCV} or its learned alternatives~\cite{Dusmanu2019CVPR,Philbin10b,Balntas2017CVPR}. 
Structure-based approaches~\cite{Robertson04BMVC,Li2010ECCV,Li2012ECCV,Choudhary12ECCV,Sattler2017PAMI,Sattler2015ICCV,Irschara09CVPR,Zeisl2015ICCV,Svarm2017PAMI} represent a scene through a SfM point cloud, with each point being associated with at least one local image feature. 
2D-3D matches between features extracted from a new image and the 3D points in the SfM model can then be used for pose estimation~\cite{Fischler81CACM,HartleyZisserman,Haralick94IJCV,Kukelova13ICCV,Kukelova2016CVPR,Bujnak08CVPR}. 
State-of-the-art approaches for long-term localization, \ie, robust camera pose estimation under changing conditions, follow this approach, but use learned features~\cite{Sarlin2019CVPR,Taira2018CVPR,Dusmanu2019CVPR,Germain20193DV}. 
An alternative to using %storing and maintaining 
a 3D point cloud is to either compute the position and orientation of a test image from relative poses to database images~\cite{Zhang06TDPVT,Zhou2019ARXIV}, or to compute a 3D model on-the-fly~\cite{Sattler2017CVPR}. Still, these approaches rely on local image features and are thus susceptible to the image inversion approach from~\cite{Pittaluga2019CVPR}. 

Learned localization approaches either replace the complete localization pipeline \cite{Kendall2015ICCV,Kendall2017CVPR,Walch2017ICCV,Brahmbhatt2018CVPR} or the 2D-3D matching stage \cite{Donoser14CVPR,Shotton2013CVPR,Brachmann2017CVPR,Brachmann2018CVPR,Brachmann2019ICCVa,Cavallari2017CVPR,Cavallari20193DV,Massiceti2017ICRA,brachmann2020ARXIV} through machine learning. % techniques such as convolutional neural networks (CNNs) or random forests. 
The former, which directly regress a camera pose via a convolutional neural network (CNN), have been shown to perform similar to image retrieval techniques~\cite{Sattler2019CVPR}, \ie, nearest neighbor classifiers that only approximate the pose of the test image~\cite{Torii2015CVPR,Torii11,Arandjelovic2016CVPR}. 
The second family of learned localization methods, which regress a 3D scene coordinate for each pixel in a test image, has been shown to achieve high pose accuracy on small scenes~\cite{Shotton2013CVPR,Brachmann2018CVPR,Brachmann2019ICCVa,Cavallari2017CVPR,Cavallari20193DV,brachmann2020ARXIV}. 
However, they currently do not scale well to larger or more complex scenes~\cite{Cavallari20193DV,Brachmann2019ICCVa}. 
As such, such learned localization systems are currently not used in practice.

\PAR{Recovering image content from features.} 
The descriptors of local features represent an abstract representation of a patch centered around a keypoint. 
A long-standing result is that it is possible to recover the image content from gradient-based features such as SIFT~\cite{Weinzaepfel2011CVPR} and HOG~\cite{Vondrick2013CVPR}, even if the descriptors are quantized into visual words~\cite{Kato2014CVPR}. 
Naturally, training CNNs to recover images improves the reconstruction quality~\cite{Dosovitskiy2016NIPS,Dosovitskiy_2016_CVPR}. 
Further work shows that the deep representations learned by neural networks can be inverted to recover images, which can be used as a tool to visualize what such networks learn~\cite{yosinski-2015-ICML-DL,Zeiler2014ECCV,Mahendran2015CVPR}.
% can be inverted \cite{Weinzaepfel2011CVPR} HOG:\cite{Vondrick2013CVPR}
% \TS{Review methods that are able to recover image content from (sparse) image features (classical and CNN). Discuss recent work that showed that this is also possible from sparse SfM point clouds, where each 3D point is associated with one or more image descriptors.} \TODO{Torsten!}

\PAR{Privacy-preserving visual localization.} 
Pittaluga \etal extended these results on recovering images from 2D local features to 3D Structure-from-Motion point clouds~\cite{Pittaluga2019CVPR}. 
They showed that a CNN can be trained to recover an image from the projection of 3D points (together with their descriptors) into a synthetic view. 
They concluded that scene representations based on SfM point clouds can allow an attacker to recover private details. 
Recently, \cite{Song2020ECCV} showed that high-quality images can also be recovered from sparse colored point clouds without image descriptors. 
In order to enable privacy preserving localization, Speciale \etal proposed to replace each SfM point by a random line passing through that point~\cite{Speciale2019CVPR}, where the direction of the line is sampled uniformly at random from a unit sphere.  
They showed that the resulting line clouds still enable precise camera pose estimation. 
They argued that line clouds preserve privacy as they prevent the approaches from~\cite{Pittaluga2019CVPR,Song2020ECCV} from being applicable, which would ensure that user-recorded scenes can be safely stored in the cloud. 
Follow-up work to this seminal paper showed how to adapt SLAM systems to integrate this idea into a SLAM system~\cite{Shibuya2020ECCV} and how to enable 
% In a follow-up work, Speciale \etal extended their approach to 
privacy-preserving image queries for localization~\cite{Speciale2019ICCV} and privacy-preserving SfM~\cite{Geppert2020ECCV}. 
\cite{Speciale2019ICCV,Geppert2020ECCV} operate on 2D rather than 3D representations and replace each 2D image feature by a 2D line. %~\cite{Speciale2019ICCV}. 

This paper investigates the claim of preserving user privacy by lifting 3D point to line clouds made by~\cite{Speciale2019CVPR}. 
We show that it is possible to (approximately) recover the underlying point positions using only the provided line geometry. 
As a result, we show that image-level details can be obtained from line clouds through recovering the underlying 3D point cloud. 
Our approach is based on the observation that two random 3D lines often enough have their closest points nearby to the original 3D points. 
However, this does not hold in 2D, \ie, our approach is only applicable to recover 3D point clouds but not 2D feature clouds.

% As a result, 
% \TS{Discuss that the result from invSfM showed that classical, feature-based localization methods are not privacy preserving as it is possible to recover image content from their scene representations. Explain how this is relevant to cloud-based services. As a consequence, Speciale \etal proposed privacy preserving localization approaches: discuss them here, highlight that we are mostly interested in the scene representation)}

% \TODO{Make sure to explain how our paper is related to previous work, this is one of the main purposes of the related work section.}

\vspace{-2pt}
\section{From Point Clouds to Line Clouds}
\label{sec:privacy_representation}
Structure-based visual localization approaches~\cite{Li2010ECCV,Li2012ECCV,Sattler2017PAMI,Irschara09CVPR,Zeisl2015ICCV} use 2D-3D correspondences between pixels  and 3D points in a scene model to estimate the camera pose of a given query image. 
To this end, classical feature-based approaches represent the scene through a 3D point cloud $\mathcal{P} = \{(\mathbf{p}_i, \mathbf{d}_i)\}$, where each 3D point $\mathbf{p}_i \in \mathbb{R}^3$ is associated with one or more image feature descriptors $\mathbf{d}_i$, \eg, a 128-dimensional SIFT~\cite{Lowe04IJCV} descriptor.\footnote{$\mathcal{P} = \{(\mathbf{p}_i, \mathbf{d}_i)\}$ is the minimally required scene representation. Some methods such as~\cite{Sattler2017PAMI,Li2012ECCV} store additional details such as co-visibility information. We only use the minimal representation in this paper.} 
Pittaluga \etal showed that it is possible to ``invert" the point clouds, often constructed using SfM, used by feature-based localization systems~\cite{Pittaluga2019CVPR}. 
More precisely, they showed that it is possible to use a CNN to recover image details from a projection of a sparse set of 3D point and their descriptors into an image. 
They concluded that the commonly used point cloud scene representations do not preserve privacy. 

To avoid revealing details of a user-uploaded scene model through the inversion process, Speciale \etal \cite{Speciale2019CVPR} propose to lift the underlying point cloud $\mathcal{P} = \{(\mathbf{p}_i, \mathbf{d}_i)\}$ to a line cloud $\mathcal{L} = \{(\mathbf{l}_i, \mathbf{d}_i)\}$. 
Each point $\mathbf{p}_i$ is replaced by a random line $\mathbf{l}_i$ passing through it.\footnote{The chosen line representation, \eg, a Pl\"{u}cker vector as in~\cite{Speciale2019CVPR}, is not important in the context of this paper. } 
This introduces an additional degree of freedom, namely the true position of the point $\mathbf{p}_i$ along the line $\mathbf{l}_i$ is unknown. 
Since these 3D lines project to lines in 2D, the inversion approach from~\cite{Pittaluga2019CVPR} is not directly applicable anymore. 
Furthermore, the resulting scene representation is unintelligible to humans (\cf Fig.~\ref{fig:teaser}). 
Consequently, Speciale \etal claim that the ``3D line cloud representation hides the underlying scene geometry and prevents the extraction of sensitive information"~\cite{Speciale2019CVPR}. 
Yet, Sec.~\ref{sec:recovery} shows that lifting a point cloud to a line cloud does not completely hide the underlying geometric properties if the line directions are uniformly sampled from a unit sphere. 
Based on this insight, we develop an algorithm to recover point clouds from line clouds.

% \TODO{Review the formulation of Speciale \etal~\cite{Speciale2019CVPR} in more detail: highlight ease of use, highlight importance of randomness (quote from their paper?), highlight that the resulting representations are unintelligible to humans.}

% % \TODO{Summarize our main finding: the randomness preserves a lot of structure that can be used to recover rather detailed point clouds, briefly summarize our approach}

% \TODO{The purpose of this section should be to give an overview over the main parts of the paper to make it as easy as possible for a reader to follow. Especially, we should highlight the importance and impact of our findings.}

% % \TODO{This should be a rather short section.}

\vspace{-2pt}
\section{Recovering Point Clouds from Line Clouds}
\label{sec:recovery}
Considered in isolation, a single line $\mathbf{l}$ is perfectly privacy-preserving as all points on $\mathbf{l}$ are equally good candidates for the true 3D point $\mathbf{p}$ that gave rise to the line. 
However, not all points along the line will  be equally likely when taking other lines into account. 
This is due to the fact that the points in the original point cloud are not randomly distributed but lie on surfaces. 
Information about the distribution might be preserved in a line cloud. 
% , \eg, regions in space containing more points than others are likely to also contain more lines.
In this paper, we show that it can be possible to recover information about the local neighborhood of points from a line cloud. 
In turn, this information can be used to recover points from lines. 
% if information about the local neighborhood of a point can be extracted from the line cloud, this information can be used to recover points from lines.
% Rather, their distribution follows certain patterns. 

% Since 

% \textcolor{blue}{Lifting a point cloud to a line cloud obfuscates the underlying 3D structure to the human eye, as evident from Fig.~\ref{fig:teaser}. However, does it truly preserve privacy, \ie, is it impossible to even approximately recover a point cloud from its line representation? If not, then up to what level of detail and under which conditions is it possible to recover the positions of the underlying points. a In this section, we consider the above questions in some detail.}

% \textcolor{blue}{
A line cloud is characterized by the distribution of the directions of lines drawn through the underlying points. 
This design choice can be used to choose %gives a scope of creativity in choosing 
a distribution that conceals the most information. 
Speciale \etal propose to sample line directions independently and uniformly at random for each point, as this distribution helps to ensure good localization accuracy~\cite{Speciale2019CVPR}. 
Consequently, we assume that the line directions are drawn independently from a uniform distribution over a unit sphere and denote the resulting line clouds as \emph{uniform line clouds}. 
Sec.~\ref{sec:recovery:information} shows that this distribution implies that the two closest points between two lines $\mathbf{l}_i$, $\mathbf{l}_j$ are likely to be relatively close to the original 3D points $\mathbf{p_i}$, $\mathbf{p}_j$. 
Sec.~\ref{sec:method} shows how to leverage this information to recover 3D points from 3D lines. 
Sec.~\ref{sec:limitations} then discusses limitations of our approach, including listing conditions under which point recovery might not be possible. 
\subsection{Information in Uniform Line Clouds}
\label{sec:recovery:information}
If we consider the point and line clouds, $\mathcal{P}$ and $\mathcal{L}$, to be random variables, then the posterior distribution $P(\mathcal{P} | \mathcal{L})$ can be obtained via Bayes rule:
\begin{equation}
 P(\mathcal{P} | \mathcal{L}) = {P(\mathcal{L} | \mathcal{P})P(\mathcal{P})}/{P(\mathcal{L})} \propto P(\mathcal{L} | \mathcal{P})P(\mathcal{P}) \enspace .   
\end{equation}
Since all line directions are drawn independently from another, we have 
% The likelihood function $P(\mathcal{L} | \mathcal{P})$ for a \textit{uniform line cloud} is described in Section~\ref{sec:privacy_representation} and states that there is one line generated for each point with a uniformly random orientation. Explicitly, this means that
% \begin{equation}
$P(\mathcal{L} | \mathcal{P}) = \prod_{i=1}^N P(\mathbf{l}_i| \mathbf{p}_i)$. 
% \end{equation}
The probability $P(\mathbf{l}_i| \mathbf{p}_i)$ of a line $\mathbf{l}_i$ given its corresponding point $\mathbf{p}_i$ is zero if $\mathbf{p}_i$ does not lie on $\mathbf{l}_i$. 
Otherwise, the probability is constant as all line directions are equally likely. 
% $P(\mathbf{l}_i| \mathbf{p}_i) = 0$ if the line $\mathbf{l}_i$ does not go through $\mathbf{p}_i$ and $P(\mathbf{l}_i| \mathbf{p}_i)$ is constant otherwise. 
Thus, the likelihood function $P(\mathcal{L} | \mathcal{P})$ is piece-wise constant, \ie, two point clouds $\mathcal{P}$, $\mathcal{P}'$ will have the same likelihood $P(\mathcal{L} | \mathcal{P}) = P(\mathcal{L} | \mathcal{P}')$ as long as every point $\mathbf{p}_i$ lies on its line $\mathbf{l}_i$. 
Consequently, a maximum a posteriori estimate is obtained by maximizing 
% Using the principle of maximum likelihood for recovering the point cloud would be pointless as the likelihood function is piece-wise constant and all {\em valid} point clouds are equally likely. A maximum a posteriori estimator, that is, the one maximizing $P(\mathcal{L} | \mathcal{P})P(\mathcal{P})$ is  obtained by maximizing 
the prior $P(\mathcal{P})$, under the constraint that all points should lie on their lines. 
Unfortunately, defining or learning a general prior distribution $P(\mathcal{P})$ seems like a hard problem. 
We thus reason about local neighborhoods instead of the global point cloud. 
The neighboring points / lines are then used to recover a point position estimate $\mathbf{\hat{p}}_i$ from the line $\mathbf{l}_i$. 

% As it will seen later in this section, our method of recovering point positions relies heavily on the prior knowledge that the underlying point cloud consists of points lying on 3D scene surfaces and consequently, for a given point $\mathbf{p}$, there are typically several neighboring points $\mathbf{p}_i$ with small distances $d(\mathbf{p},\mathbf{p}_i)$.
% % }

\begin{figure}[t!]
    \centering
    \includegraphics[width=0.5\linewidth]{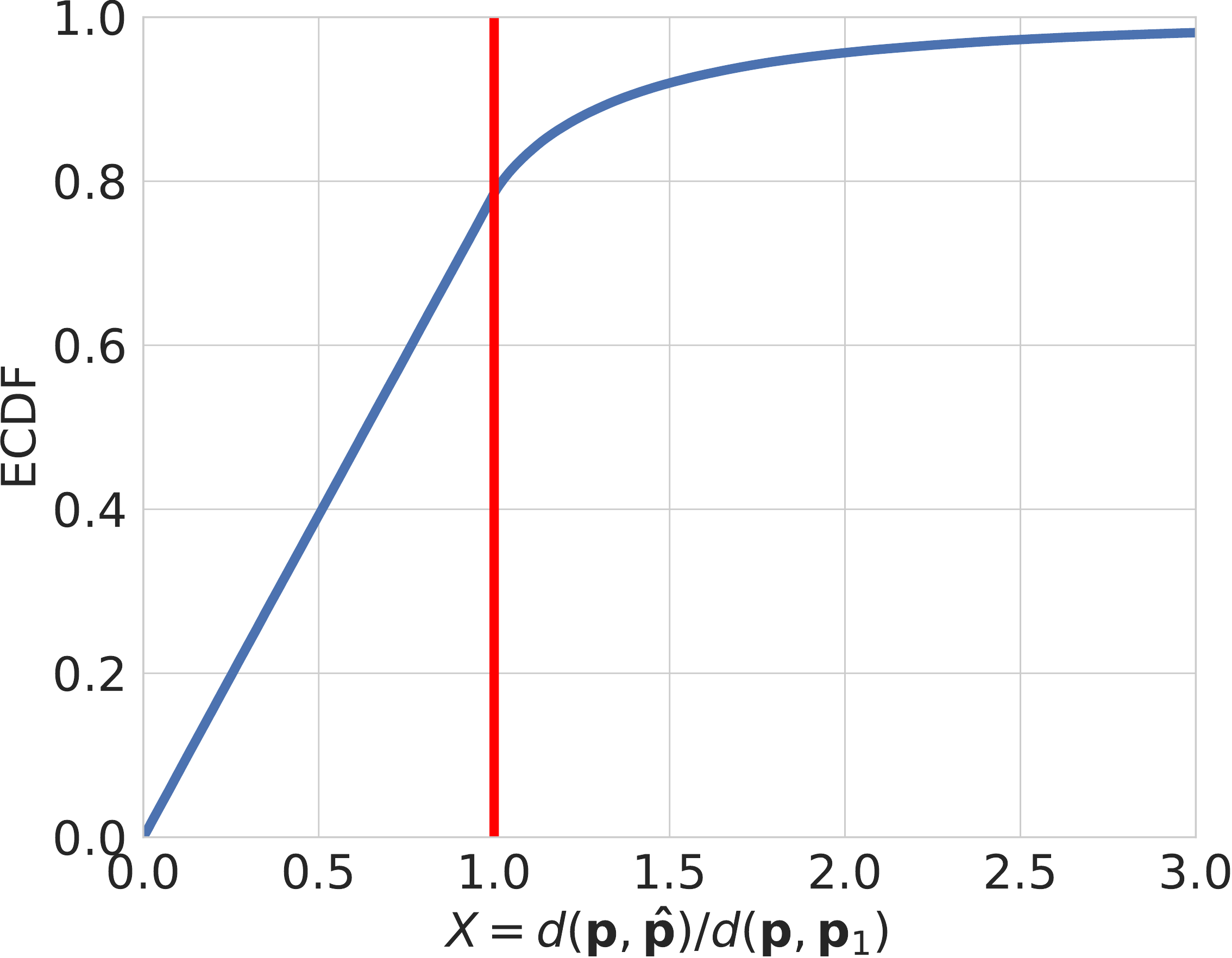}
    \caption{Empirical cumulative distribution of the relation between the Euclidean  distance $d(\mathbf{p}, \mathbf{p}_1)$ of two points $\mathbf{p}$, $\mathbf{p}_1$ and the distance $d(\mathbf{p}, \mathbf{\hat{p}})$ between $\mathbf{p}$ and an estimate $\mathbf{\hat{p}}$ for $\mathbf{p}$ obtained via the closest points of the two lines corresponding to $\mathbf{p}$ and $\mathbf{p}_1$. As can be seen, there is a high chance that $\mathbf{p}$ is closer to $\mathbf{\hat{p}}$ than to $\mathbf{p}_1$, where the red line denotes equal distance. 
    }% distance of the error on sampling random lines through 2 points. The red line e = 1 is used to provide an idea of the accuracy of estimation using closest points on lines.}
    \label{fig:err_cumulative_dist}
\end{figure}

Consider two 3D points, $\mathbf{p}$ and a nearby point $\mathbf{p}_1$, as well as their corresponding lines $\mathbf{l}$, $\mathbf{l}_1$. 
A simple approach to obtain an estimate for $\mathbf{p}$ is to find the point $\mathbf{\hat{p}}$ on the line $\mathbf{l}$ with minimum Euclidean distance to $\mathbf{l}_1$. 
In the following, we show that for uniform line clouds, $\mathbf{\hat{p}}$ is likely to be a relatively good estimate for $\mathbf{p}$. 
To this end, we study the distribution of the random variable $\mathrm{X} = d(\mathbf{p}, \mathbf{\hat{p}}) / d(\mathbf{p}, \mathbf{p}_1)$, where $d(\cdot, \cdot)$ is the Euclidean distance between two 3D points. 
We empirically measure this distribution by fixing the distance $d(\mathbf{p}, \mathbf{p}_1)=1$ and randomly sampling the directions of the two lines $\mathbf{l}$, $\mathbf{l}_1$ for one million iterations. 
%we use an experimental result obtained using 1 million runs of randomly sampling the directions $\mathbf{v}$ and $\mathbf{v}_1$, to motivate the possibility of point recovery. 
Fig.~\ref{fig:err_cumulative_dist} shows the resulting cumulative distribution of $\mathrm{X}$. A key observation that can be drawn from this is that in nearly $80\%$ of the cases, $d(\mathbf{p}, \mathbf{\hat{p}})$ is smaller than $d(\mathbf{p}, \mathbf{p_1})$. 
This implies that if %It can also further be concluded that knowing 
$d(\mathbf{p}, \mathbf{p_1})$ is small, there is a good chance that $\mathbf{\hat{p}}$ will be close to the true point position $\mathbf{p}$. %we know that there is a good chance that the estimate along $\mathbf{l}$ obtained from $\mathbf{l}_1$ is quite accurate.  

The above analysis can be further extended to a neighborhood of size $k$. Let $\mathcal{N}^k(\mathbf{p}) = \{1, 2, \dots , k \}$ be the indices of the neighboring points of $\mathbf{p}$. Let $\{\mathbf{l}_1, \mathbf{l}_2, \dots ,\mathbf{l}_k \}$ be the lines through these neighboring points and let
\begin{equation}
    \mathbf{d}_{max} = \max\nolimits_{j \in \mathcal{N}^k(\mathbf{p})} d(\mathbf{p}, \mathbf{p}_j)  
\end{equation}
be the maximum distance between $\mathbf{p}$ and any of its neighbors. 
Since all directions are drawn independently, uniformly at random from a unit sphere, the result from the 2-point analysis from above holds pairwise for each pair $(\mathbf{p}, \mathbf{p}_j)$, $j \in \mathcal{N}^k(\mathbf{p})$. 
Thus, from the $k$ estimates obtained on $\mathbf{l}$ using lines $\{\mathbf{l}_1, \mathbf{l}_2, \dots ,\mathbf{l}_k \}$, $0.8k$ can be expected to lie within a distance $\mathbf{d}_\text{max}$ from $\mathbf{p}$. If $\mathbf{d}_\text{max}$ is small, this leads to a clustering of estimates close to the true point $\mathbf{p}$.   
% }

\begin{figure*}
    \centering
    \subfloat[Ground Truth]{\includegraphics[width = 0.195\linewidth]{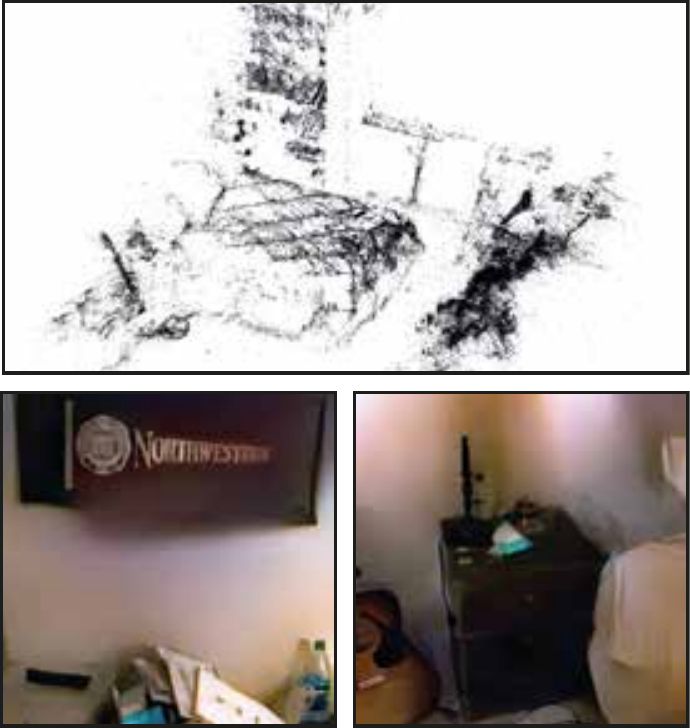}} \hspace{0.5pt}
    \subfloat[{No outliers}]{\includegraphics[width = 0.195\linewidth]{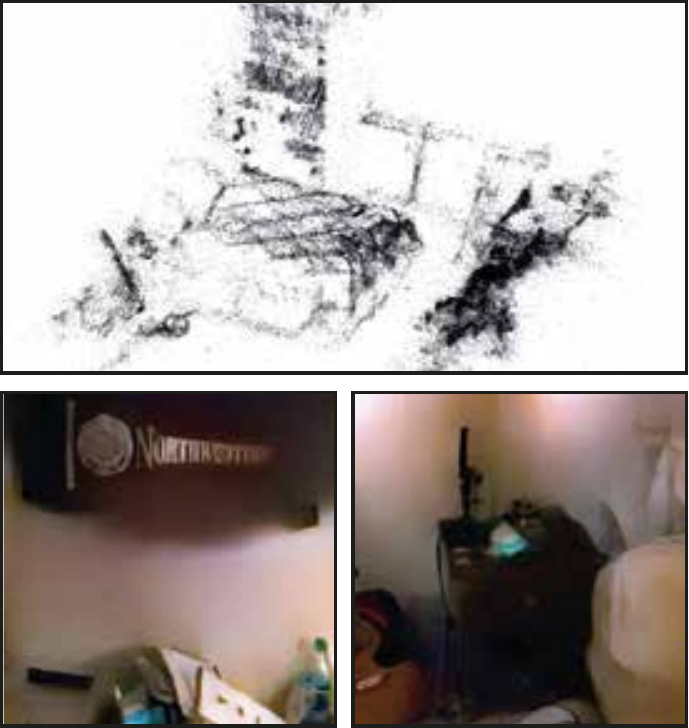}} \hspace{0.5pt}
    \subfloat[{50\% outliers}]{\includegraphics[width = 0.195\linewidth]{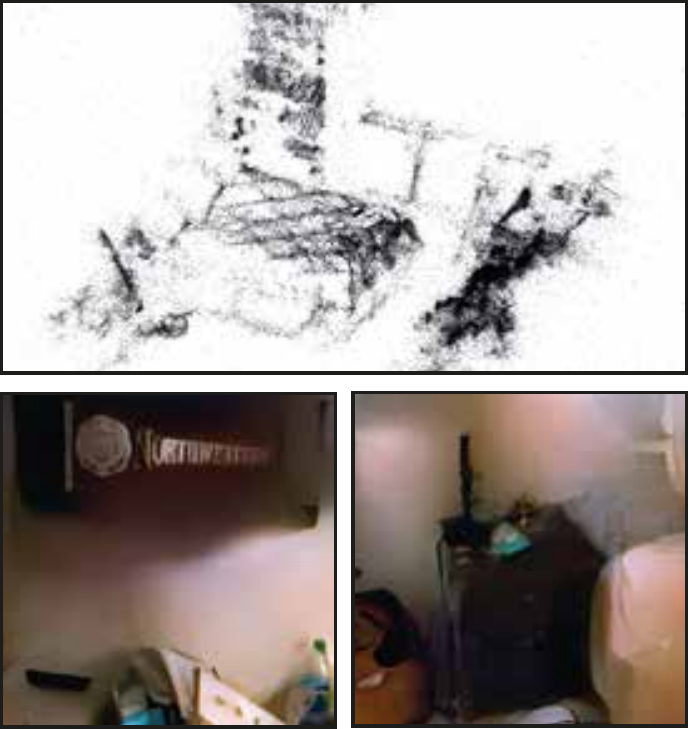}} \hspace{0.5pt}
    \subfloat[{90\% outliers}]{\includegraphics[width = 0.195\linewidth]{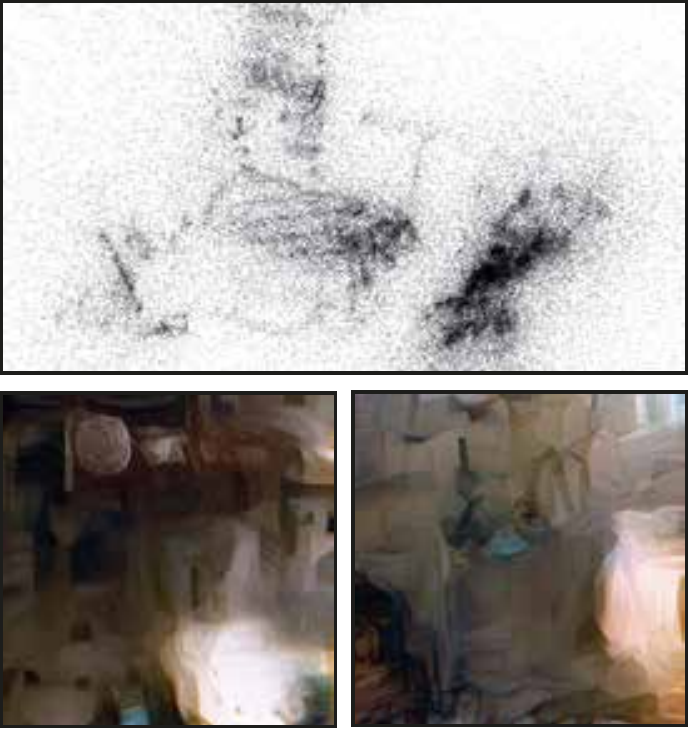}} \hspace{0.5pt}
    \subfloat[{Ours}]{\includegraphics[width = 0.195\linewidth]{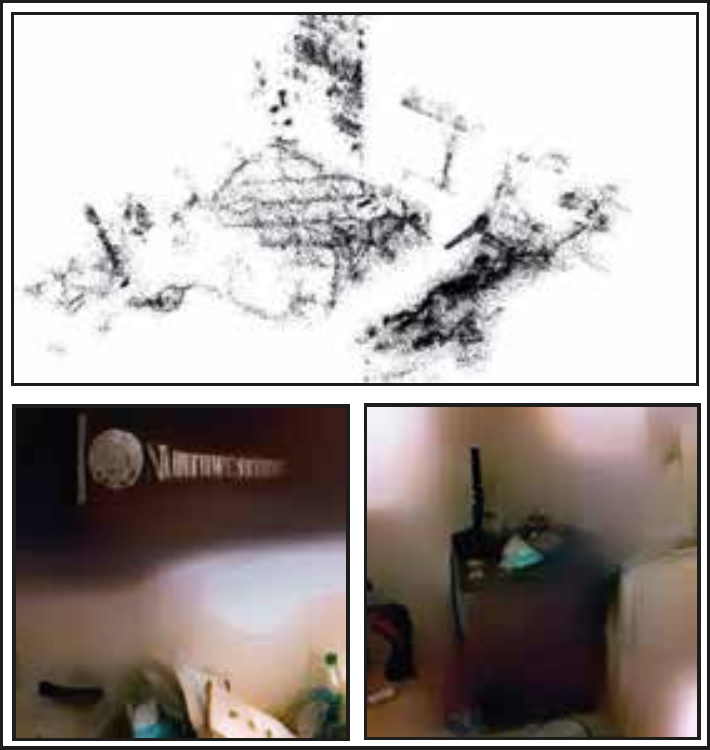}}
    % \begin{subfigure}{.195\linewidth}
    %\centering
    % \includegraphics[width = 1\linewidth]{images/nns_exp_gt.pdf}
    % \caption{Ground Truth}
    % \end{subfigure}
    % \begin{subfigure}{.195\linewidth}
    %\centering
    % \includegraphics[width = 1\linewidth]{images/nns_exp_true50.pdf}
    % \caption{No outliers}
    % \end{subfigure}
    %  \begin{subfigure}{.195\linewidth}
    %  %\centering
    % \includegraphics[width = 1\linewidth]{images/nns_exp_true25.pdf}
    % \caption{50\% outliers}
    % \end{subfigure}
    % \begin{subfigure}{.195\linewidth}
    %  %\centering
    % \includegraphics[width = 1\linewidth]{images/nns_exp_true5.pdf}
    % \caption{90\% outliers}
    % \end{subfigure}
    %  \begin{subfigure}{.195\linewidth}
    %  %\centering
    % \includegraphics[width = 1\linewidth]{images/nns_true_exp_ours.pdf}
    % \caption{Ours}
    % \end{subfigure}
    \caption{Top row: (a) ground truth point cloud, (b) - (d) point clouds recovered using true neighborhoods with different outlier contamination levels. (e) the result from our method. Bottom row: Images reconstructed from the point clouds using~\cite{Pittaluga2019CVPR}. }
    \label{fig:nns_exp}
\end{figure*}

The results from above suggest that given information about the $k$ nearest neighbors of each point, it should be possible to obtain accurate point estimates from a line cloud. % points along their lines. 
We verify this intuition % that this analysis translates nicely into practical usage, 
though a simple experiment on an indoor scene (\cf  Fig.~\ref{fig:nns_exp}(a)) for $k=50$. 
% Using the $k=50$ nearest neighbors of each point in the underlying point cloud, we recover the point positions from the produced uniform line cloud. 
For each point, we obtain 50 estimates on its line using the lines through its nearest neighboring points. 
Simply taking the median of these estimates produces the result shown in Fig.~\ref{fig:nns_exp}(b). 
To measure the impact of imperfect neighborhoods, we randomly replace 50\% / 90\% of the neighbors with random lines from the line cloud. 
% We then analyse what happens if the correct neighborhood is not known precisely and consider cases of 50\% and 90\% of the neighbors being randomly taken from the sample of all points. 
As can be seen in Fig.~\ref{fig:nns_exp}(c), even with 50\% outliers, it is still possible to recover the underlying point cloud. 
However, 90\% outliers lead to a very noisy point cloud and the images obtained via \cite{Pittaluga2019CVPR} from this point cloud become unintelligible (\cf Fig.~\ref{fig:nns_exp}(d)). 
For comparison, Fig.~\ref{fig:nns_exp}(e) shows results obtained with our approach, introduced in Sec.~\ref{sec:method}, that aims to recover the neighborhood from all lines in the line cloud. 

% \textcolor{blue}{
\subsection{Recovering Points from Uniform Line Clouds}%Method - Estimating Point Positions}
\label{sec:method}
As motivated above, estimating a 3D point position $\mathbf{p}_i$ from a given line $\mathbf{l}_i$ can be modelled as a two-stage process: % as a two-step process:
% }
% \textcolor{blue}{
the first stage, \textbf{neighborhood estimation}, identifies the neighboring 3D points of $\mathbf{p}_i$ through their corresponding lines. 
For each such line $\mathbf{l}_j$, the point on $\mathbf{l}_i$ closest to $\mathbf{l}_j$ provides an estimate for the true point position $\mathbf{p}_i$. % estimates a set of candidate positions for $\mathbf{p}$ as the nearest 
Given these candidate positions along $\mathbf{l}_i$, the second stage then selects a single candidate. 
This is achieved by finding high-density regions of the candidates along $\mathbf{l}_i$ via \textbf{peak finding}. 
% \begin{enumerate}
    % \item Neighborhood Estimation: Find lines passing through the neighboring points of $\mathbf{p}$ and obtain the nearest points from these lines on $\mathbf{l}$ as a set of estimates.
    % \item Peak Finding: Find the point of peak density of the estimates obtained above along the line to be used as the final estimate. 
% \end{enumerate}

Multiple iterations of these two steps are performed to improve the estimates. 
In the first iteration, only line-to-line distances can be computed and used to select a neighborhood of lines for $\mathbf{l}_i$. 
Starting from the second iteration, point estimates are available for each line. 
These point estimates can be used to obtain better neighborhood estimates, which in turn lead to better 3D point predictions (\cf Fig.~\ref{fig:nns_exp}). 
% While the algorithm for the peak finding step remains same across all iterations, the Neighborhood Estimation step is changed after the first iteration to use the additional information available about initial point estimates.
% Multiple iterations are helpful because improved point estimates help in better approximation of the neighborhood and as can be seen from Fig.~\ref{fig:nn_impact}, neighborhood estimation is the key for accurate recovery of point positions.
% }
% \textcolor{blue}{
% We discuss Neighborhood Estimation in detail in Sec.~\ref{sec:nn_estimation} and explain the Peak Finding algorithm in Sec.~\ref{sec:peak_finding} 
% }

% \textcolor{blue}{
\PAR{Neighborhood Estimation.} \label{sec:nn_estimation}
In the first iteration, we can only measure the minimum distance $d_{ll}(\mathbf{l}_i, \mathbf{l}_j)$ between lines $\mathbf{l}_i$ and $\mathbf{l}_j$, defined as the Euclidean distance between the closest points on the lines. 
The minimum distance between two lines cannot be larger than the distance between the corresponding points. 
Thus, the set $\mathcal{N}_{ll}^K(\mathbf{l}_i) = \{i_1, \dots, i_K\}$ containing the indices of the $K$ nearest lines to $\mathbf{l}_i$ should contain part of the true neighborhood $\mathcal{N}^k(\mathbf{p}_i)$ if $K>k$ is large enough. %, but also outliers. % with ouliers, \ie, lines corresponding to points that are not part of the true neighborhood. 
 
% With only a set of lines as input, one simple and intuitive way to estimate the neighborhood of the underlying point of a line would be to use the minimum line-line distance as a metric for finding the line's nearest neighbors. This is based on the fact that the minimum distance between lines through two neighboring points can not be more than distance between those points.
% }
% \textcolor{blue}{
Once point position estimates $\{\mathbf{\hat{p}}_i | i \in 1, \dots, N \}$ for the input lines $\{\mathbf{l}_i | i \in 1, \dots, N \}$ are obtained, we compute two additional neighborhoods: 
$\mathcal{N}_{lp}^K(\mathbf{\hat{p}}_i) = \{i_1, \dots, i_K\}$ is the set of indices belonging to the $K$ lines that have the smallest Euclidean distances to the point estimate $\mathbf{\hat{p}}_i$. 
Similarly, $\mathcal{N}_{pl}^K(\mathbf{l}_i)$ is the set of indices belonging to the $K$ point estimates that have the smallest Euclidean distances to the line $\mathbf{l}_i$. 
%
% instead of just a line $\mathbf{l}_i$ at index $i$, a tuple $(\mathbf{\hat{p}}_i, \mathbf{l}_i)$ is available for computing distances. Therefore, while computing the nearest neighbors of tuple at index $i$, the following distance metrics to a tuple at index $j$ become possible:
% \begin{enumerate}
%     \item $d(\mathbf{\hat{p}}_i, \mathbf{\hat{p}}_j)$ - point to point distance
%     \item $d(\mathbf{\hat{p}}_i, \mathbf{l_j})$ - point to line distance 
%     \item $d(\mathbf{l}_i, \mathbf{\hat{p}}_j)$ - line to point distance
%     \item $d(\mathbf{l}_i, \mathbf{l}_j)$ - line to line distance.
% \end{enumerate}
% Let the corresponding neighborhoods obtained using these distance metrics be called $\mathcal{N}_{pp}(i)$, $\mathcal{N}_{pl}(i)$, $\mathcal{N}_{lp}(i)$, $\mathcal{N}_{ll}(i)$ respectively. Note that these neighborhoods are sets of indices pointing to tuples instead of just sets of lines.
Intuitively, we expect $\mathcal{N}_{pl}^K(\mathbf{l}_i)$ to contain those true neighbors that have been 
% it makes sense to consider nearest estimated points to a line as neighbors since it is expected that some of the true neighbors would have been 
estimated close to their true 3D position in the previous iteration. 
In practice, we observe that $\mathcal{N}_{pl}^K$ often has a comparatively high overlap with the true neighborhood of points in $\mathcal{N}^k(\mathbf{p}_i)$. % in most cases. 
However, it still contains outliers corresponding to 3D points estimated in regions through which $\mathbf{l}_i$ passes. %in the sense that many far away estimated points can also lie close to the line. Using just the intersection of $\mathcal{N}_{lp}$ with either $\mathcal{N}_{pl}$ or $\mathcal{N}_{pp}$, when it contains enough neighbors, helps counter this problem to a great effect.
% }
Similarly, we can expect $\mathcal{N}_{lp}^K(\mathbf{\hat{p}}_i)$ to contain lines corresponding to the true neighboring points from $\mathcal{N}^k(\mathbf{p}_i)$, but also outliers from lines from unrelated points that pass through the region containing $\mathbf{\hat{p}}_i$. 
We thus estimate the neighborhood of $\mathbf{l}_i$ as the intersection  $\mathcal{N}_{pl}^K(\mathbf{l}_i) \cap \mathcal{N}_{lp}^K(\mathbf{\hat{p}}_i)$.

In our experience, an initial neighborhood size of $K$ between 300 and 500 for $\mathcal{N}_{ll}^K(\mathbf{l}_i)$, followed by a smaller neighborhood size of 100 to 200 for $\mathcal{N}_{pl}^K(\mathbf{l}_i)$ and $\mathcal{N}_{lp}^K(\mathbf{\hat{p}}_i)$ %and $\mathcal{N}_{pl}$ of size 100-200 
works well for a line cloud of about 100k lines.

% \textcolor{blue}{
% To summarize, initially the line-line distance based neighborhood, $\mathcal{N}_{ll}$ is used to obtain initial coarse estimates. Then, either $\mathcal{N}_{lp}$ or its intersection with $\mathcal{N}_{pl}$ is used when possible to iteratively refine the obtained estimates. In practice, an initial neighborhood of $\mathcal{N}_ll$ size 300-500, followed by a smaller $\mathcal{N}_{lp}$ and $\mathcal{N}_{pl}$ of size 100-200 work well for a line cloud of 100k lines.
% }

% \textcolor{blue}{The neighborhood once obtained, provides several estimates for the point position along its line and the set of scalars defining the positions of these estimates are the input to the peak finding algorithm. More specifically, consider a line $\mathbf{l}_i = [ \mathbf{o}_{i}^{\top} \mathbf{v}_{i}^{\top}]$ with $\mathcal{N}(i)$ as its appropriately computed neighborhood. Let $\{\mathbf{\hat{p}}_{ik} | k \in \mathcal{N}(i)\}$ be the set of closest points on $\mathbf{l}_i$ from the lines belonging to the neighborhood $\mathcal{N}(i)$. The input to the next step is the set
% \begin{equation}\label{eq:estimate_set}
% \mathcal{E}_i = \{ \beta_{ik} | \mathbf{\hat{p}}_{ik} = \mathbf{o}_i + \beta_{ik} \mathbf{v}_i , \forall k \in \mathcal{N}(i)\}    
% \end{equation}
% }

% \TODO{Describe how we select $K$.}

\begin{figure*}[!t]
    \centering
    \includegraphics[width=0.9\linewidth]{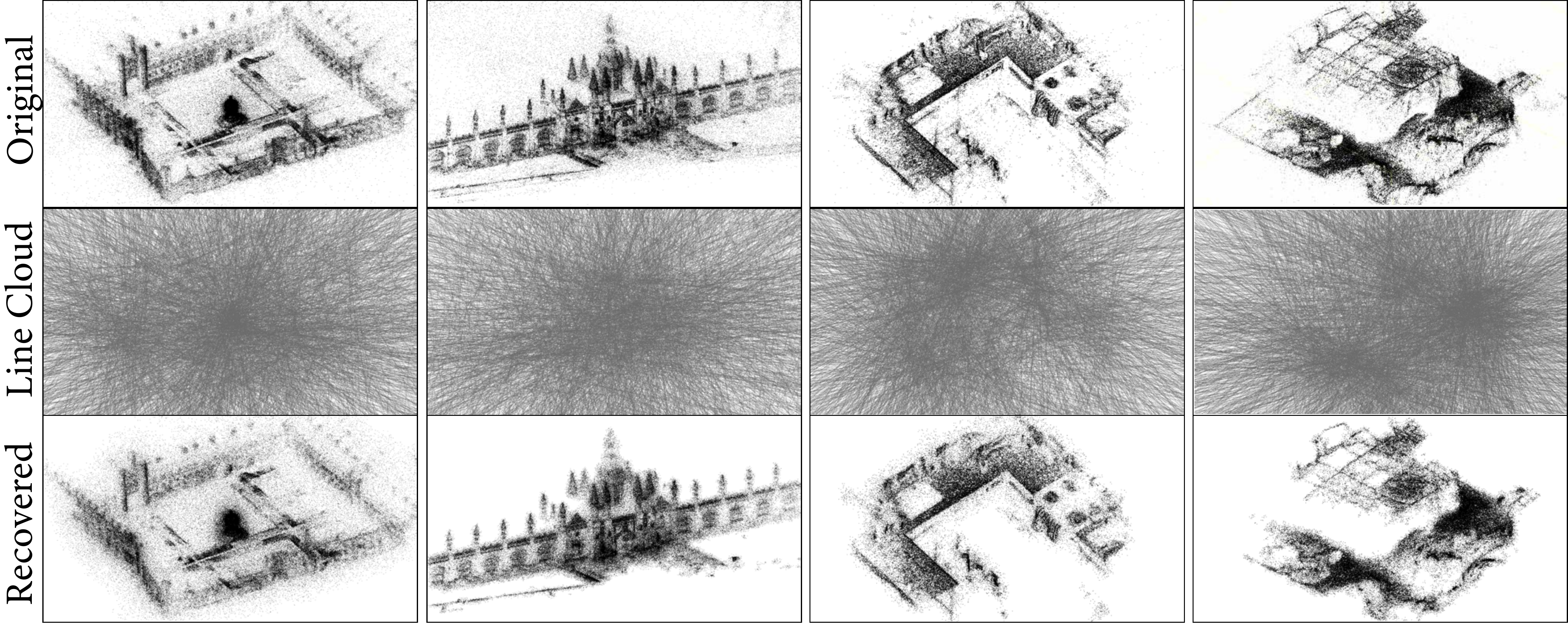}
    \caption{Qualitative results showing point clouds recovered for two scenes each from the outdoor Cambridge Landmark~\cite{Kendall2015ICCV} and the indoor 12 Scenes~\cite{Valentin20163DV} datasets. From left to right, we have the 'Great Court', 'Kings College', 'Apt2-Kitchen'  and 'Apt2-Bedroom' scenes. For the Cambridge scenes, 30\% of all available lines were used for the shown recovered point cloud, while for the indoor scenes from 12 Scenes, all the lines were used. A statistical outlier removal step is performed on the recovered point clouds before visualization.}
    \label{fig:qualitative_points}
\end{figure*}

% \textcolor{blue}{
\PAR{Peak Finding.} \label{sec:peak_finding}
Let $\mathcal{N}^K(\mathbf{l}_i)$ be the neighborhood for line $\mathbf{l}_i$ from the neighborhood estimation stage. 
Each line $\mathbf{l}_j$, $j \in \mathcal{N}^K(\mathbf{l}_i)$, provides a 3D point candidate estimate $\mathbf{\hat{p}}_{ij}$ for $\mathbf{p}_i$, defined as the point on $\mathbf{l}_i$ with the minimum distance to $\mathbf{l}_j$. 
Let $\mathbf{o}_i$ be an arbitrary 3D point on $\mathbf{l}_i$ and $\mathbf{v}_i$ be the 3D direction of the line. We can then parameterize the candidates as a distribution of scalar values 
\begin{equation}\label{eq:estimate_set}
\mathcal{E}_i = \{ \beta_{ij} | \mathbf{\hat{p}}_{ij} = \mathbf{o}_i + \beta_{ij} \mathbf{v}_i , \forall j \in \mathcal{N}^K(i)\}    
\end{equation}
along the line. 
We select the candidate 3D position $\mathbf{\hat{p}}_i$ for $\mathbf{p}_i$ by finding high-density regions in this distribution. %\TODO{Probably want to have more details in the supp. mat.}
Following~\cite{Lynen2017IJCV,Lynen20143DV}, we use the Kuiper's statistic \cite{Kuiper1960KNAW}, a non-parametric test statistic that can be used to measure where a given cumulative distribution function (CDF) $F_{X}(x)$ differs the most from a reference CDF $F_{T}(x)$, to identify high density regions. 
We measure the unweighted empirical CDF 
\begin{equation}
    F_{i}(x) = \frac{1}{K} \sum\nolimits_{j = 1}^{K}{I_{\beta_{ij} < x}}     
\end{equation}
describing the distribution of the points from $\mathcal{E}_i$ on $\mathbf{l}_i$. 
Here, $I_{\beta_{ij} < x}$ is an indicator variable taking value 1 if ${\beta_{ij} < x}$ and 0 otherwise. 
We compare $F_{i}(x)$ with the CDF $F_U(x)$ of a uniform distribution between the minimum and maximum value from $\mathcal{E}_i$. 
We compute the two points $\bar{x}^{-} = \argmax_{x} \left(F_{U}(x) - F_{i}(x)\right)$ and $\bar{x}^{+} = \argmax_{x} \left(F_{i}(x) - F_{U}(x)\right)$ corresponding to the positions along the line where the two distributions differ most. 
The differences between the distributions at these points are given as $D^{-} = \left(F_{U}(\bar{x}^{-}) - F_{i}(\bar{x}^{-})\right)$ and $D^{+} = \left(F_{i}(\bar{x}^{+}) - F_{U}(\bar{x}^{+})\right)$ and the Kuiper's statistic is then defined as $\text{KS} = D^{-} + D^{+}$. 

Intuitively, $\bar{x}^{-}$ and $\bar{x}^{+}$ define the start and end points of a high-density region of the points along the line and $\text{KS}$ provides a measure for how much this density differs from a uniform distribution. 
% can be interpreted as a measure of the the prominence of the peak in some sense. We can then recursively move close to the actual peak value by considering the CDF of estimates within the obtained peak and repeat the process of finding start and end points of the peak. 
Recursion within the range $\bar{x}^{-}$ and $\bar{x}^{+}$ can be used to refine the range. 
The value of $\text{KS}$ can be used to decide when we are sufficiently close to the peak and we stop once $\text{KS}$ drops below 0.4. 
The median value within the range is then used as the estimate $\mathbf{\hat{p}}_i$ for $\mathbf{p}_i$. 
% and hence, can stop re-cursing. 
% In practice, if $K$ drops below a threshold of 0.4 it signifies the flat top of the peak and hence a median value of the estimates within it are used as a robust final estimate.
Similarly, recursion outside the range $\bar{x}^{-}$ and $\bar{x}^{+}$ can be used to identify multiple candidates and we select the one with the largest $\text{KS}$ value. 
% In case of multiple peaks, we break the distribution at each end of peak, individually analyse each peak and pick the one with the highest $K$. The same procedure as above is then repeated for that peak to obtain the final estimate. Figure \ref{fig:peaks} shows a typical example of finding the peak for estimates on a line. The figures on the left depict finding start and end points of peak ($\bar{x}^{-}$ and $\bar{x}^{-}$) recursively, while the figure on the right shows the corresponding region selected in the distribution of estimates.
Please see the supp. material for a more detailed description of the peak finding procedure.

\vspace{-2pt}
\subsection{Limitations}
\label{sec:limitations}
Our approach fails if it either is not able to recover enough true neighbors in the first stage (\cf Fig.~\ref{fig:nns_exp}) or if good position estimates cannot be found in the second stage. 

Naturally, changing the distribution of line directions such that the closest points between pairs of lines are far away from the original points will cause both stages to fail. 
However, the localization approach presented in~\cite{Speciale2019CVPR} is based on modelling pose estimation as a relative pose problem for generalized cameras. This allows~\cite{Speciale2019CVPR} to estimate the absolute scale of the translation and thus the absolute pose of the test image. 
In the case that all lines used for pose estimation intersect in the same point, the relative pose problem degenerates to the classical perspective relative problem~\cite{Nister06CVPR}. 
In this case, the translation can only be recovered up to a unknown scaling factor and localization fails. 
As such, there is a trade-off between pose accuracy and  preventing the use of closest points on lines. 
While out of scope for this work, we believe that this is  an interesting direction for future research (\cf Sec.~\ref{sec:conclusion}). %.\footnote{Potentially, our approach could be used with a human or classifier in the loop to determine whether a recovered point cloud is meaningful. In case our approach fails, one could then exclude certain lines from the neighborhoods used in the next iterations.}

An alternative strategy to prevent our method from recovering accurate 3D point positions is to represent the scene as a sparse line cloud. % the input line clouds. 
Using only a subset of the original 3D points~%, as commonly done when compressing representations for memory-efficient localiation~\cite{Li2010ECCV,Cao2014CVPR,Camposeco2019CVPR}, increases the distance between the points. 
Consequently, it becomes harder to identify true neighbors based on line-line and point-line distances in the first stage of our algorithm. 
Even if we can recover the true neighborhood, increasing distances between the true neighbors decreases the chance that the closest points on two lines are close to the true point positions (\cf Sec.~\ref{sec:recovery:information}). 
Thus, peak finding might fail if the line clouds are too sparse. 
Since~\cite{Speciale2019CVPR} showed that accurate localization is still possible when using 5-10\% of the original lines, we consider this case in our experiments. % evaluation. 

Linked to the challenges induced by using sparse line clouds is that our approach struggles to recover points in areas with low point density. 
Consider the line of a point from a sparse region in the original point cloud that passes through a region containing more points. 
In this case, there is a good chance that most of the neighbors identified by our method will come from this denser region rather than the original neighborhood. 
This results in predicting the point position in the wrong part of the scene. 

% , shown in Fig.~\ref{fig:nns_exp}, 
% \TODO{Torsten: Discuss limitations: How valid is the assumption of uniform line directions? $\rightarrow$ mention that other distributions can get our approach to fail, but also that not every distribution is suitable for localization $\rightarrow$ needs more experiments, which is outside of the scope of this work\\
% Discuss problems with more dense regions $\rightarrow$ if line goes through dense regions in space, many neighbors will be outliers and it gets harder to recover the point, say that we observe this in practice $\rightarrow$ motivate combining line with sparsity.}

% \TODO{Mention that we are only using line geometry}

% Central to our approach is that the random lines are created by uniformly at random sampling directions on the unit sphere as this leads to the two closest points between lines providing good approximations to the original 3D points. 
% A natural direction for future work is thus to investigate different probability distributions to avoid this behavior. A naive solution would be to have all lines intersect in one or a few points. 

\begin{figure*}[!t]
    \centering
    \subfloat{\includegraphics[width = 0.25\linewidth]{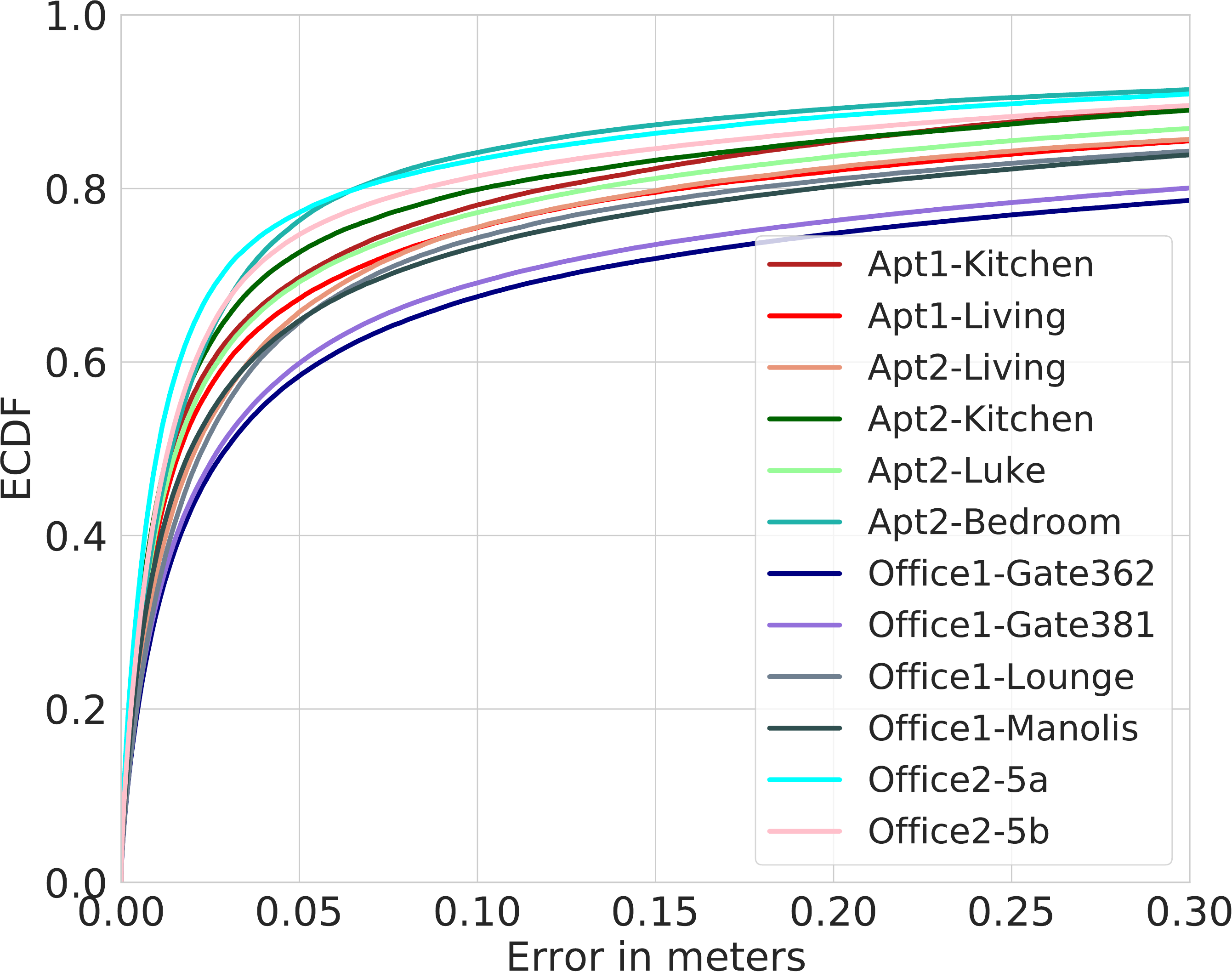}} \hspace{0.5pt}
    \subfloat{\includegraphics[width = 0.25\linewidth]{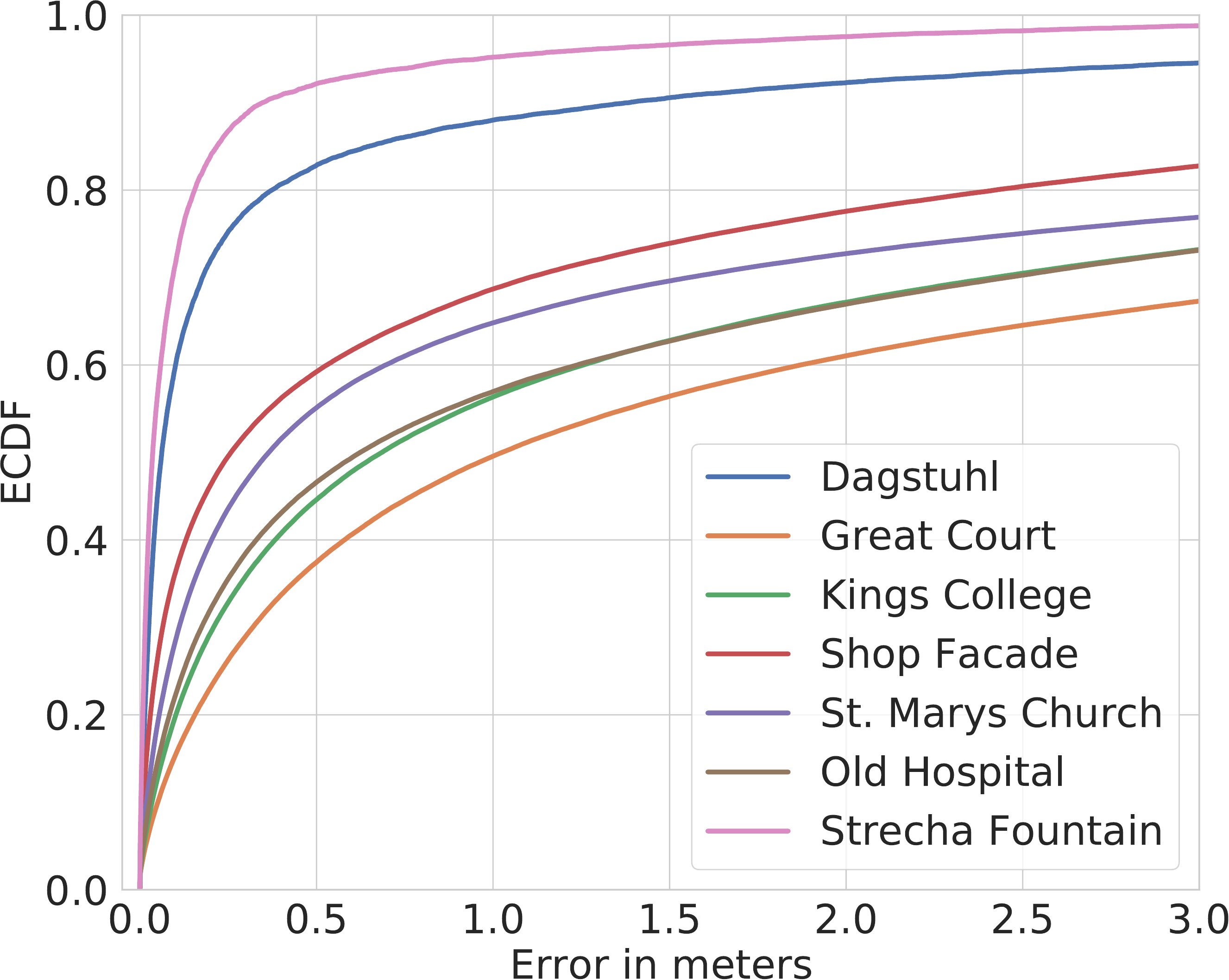}} \hspace{0.5pt}
    \subfloat{\includegraphics[width = 0.25\linewidth]{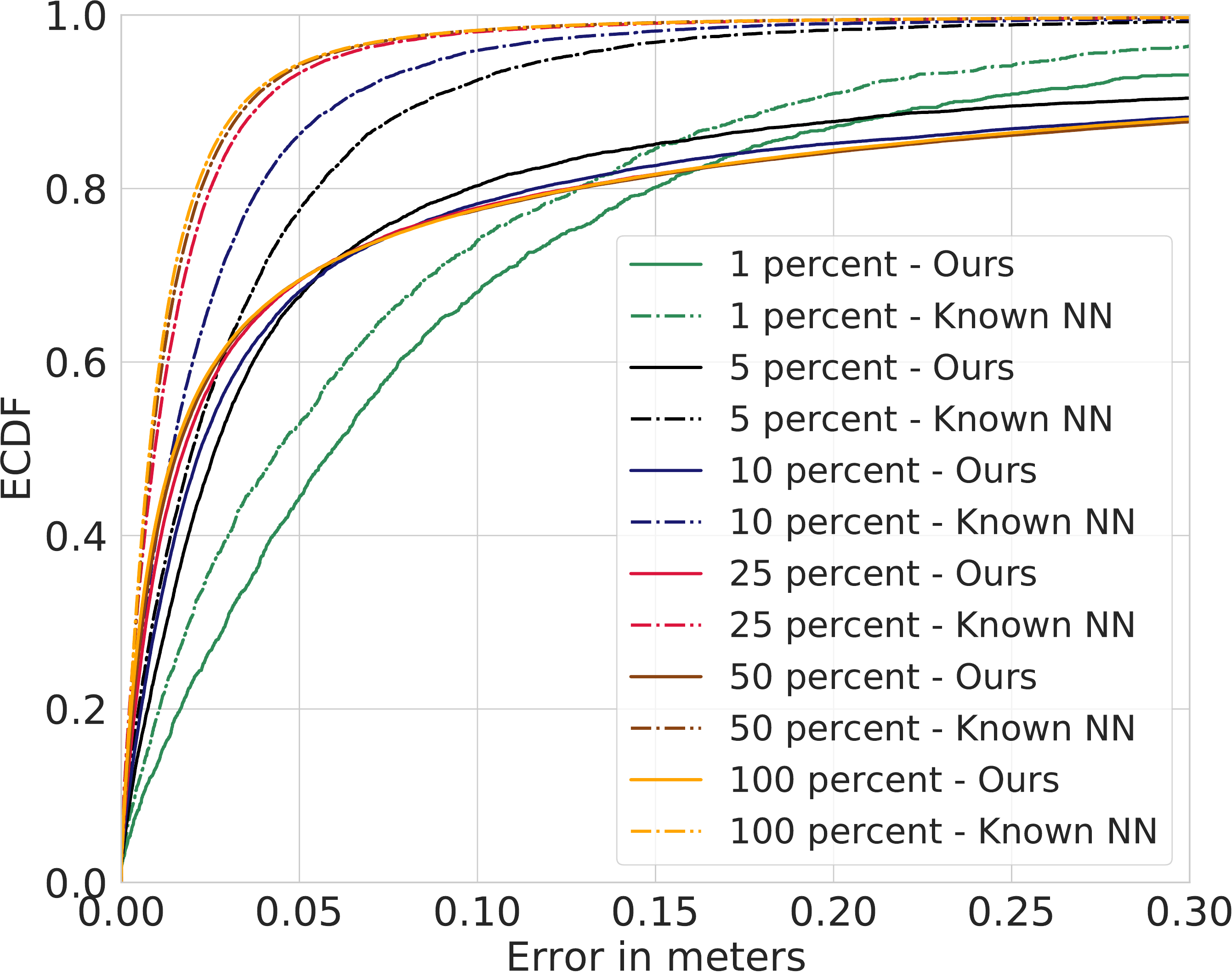}} %\hspace{0.5pt}
    % \begin{subfigure}{0.33\linewidth}
    % \includegraphics[width=1\linewidth]{images/Indoor_Errs-crop.pdf}%
    % \caption{}
    % \end{subfigure}
    % \begin{subfigure}{0.33\linewidth}
    % \includegraphics[width=1\linewidth]{images/Outdoor_Errs-crop.pdf}%
    % \caption{}
    % \end{subfigure}
    % \begin{subfigure}{0.33\linewidth}
    % \includegraphics[width=1\linewidth]{images/Sparsity_Errs-crop.pdf}%
    % \caption{}
    % \end{subfigure}
    \caption{Quantitative results showing the cumulative distribution of errors in the recovering point positions. (Left) Results for each scene from the 12 scenes datasets.  (Middle) Results for the outdoor datasets. (Right) Results averaged over four  indoor scenes for different sparsity levels. ``Known NN" denotes a variant of our method where the true neighborhood is given by an oracle.}
    \label{fig:quantitative}
\end{figure*}

\vspace{-2pt}
\section{Experimental Evaluation}
In this section, we evaluate our approach for recovering point clouds from line clouds on a set of indoor and outdoor scenes. 
We first show qualitative and quantitative results for the recovered 3D point clouds. 
Next, we analyze to what degree image content can be recovered by applying the SfM inversion procedure from~\cite{Pittaluga2019CVPR} on our recovered point clouds. 
Finally, we show that very sparse line clouds can be used to prevent~\cite{Pittaluga2019CVPR} from obtaining human-interpretable images from our recovered point clouds. 
Overall, our experiments show that lifting point clouds to line clouds by itself is not sufficient to obfuscate the underlying scene geometry. 

% \textcolor{blue}{
% In the context of camera localization, a privacy breach would refer to leakage of details about the scene being considered. These can be the structural details that are visible in the point cloud itself, for \eg the high level geometry of the scene, presence of common objects such as tables, beds, chairs etc. At the same time, if the point cloud has been recovered to reasonable accuracy, an inversion attack can lead to a possible leak of finer details such as textual information present on papers, screens, brands and colors of objects etc.
% }
% \textcolor{blue}{
% In this section, we present exhaustive results, both qualitative and quantitative, to measure the extent to which these violations can occur.  As noted before, our method of recovering a point's position heavily relies upon finding lines through points that are very close to it in the underlying point cloud. This naturally becomes tough in the case of sparse point clouds. We therefore also analyse the impact of sparsity on our method's performance and discuss the overall effectiveness of line clouds.
% }

%To prove the effectiveness of our technique in revealing the structure of underlying point clouds from the lifted line clouds, we conduct qualitative and quantitative tests over real world datasets of varying complexity, scale and density. After detailing the datasets and evaluation protocol in the following, we present qualitative results in Sec~\ref{sec:results:qualitative}. 
%Sec.~\ref{sec:results:quantitative} then provides quantitative results on the accuracy of the recovered point clouds. 

\PAR{Datasets.} For evaluation, we use the Cambridge Landmarks~\cite{Kendall2015ICCV} and 12 Scenes~\cite{Valentin20163DV} datasets. % commonly used in the visual localization literature. 
% The Cambridge Landmarks dataset 
The former depicts individual outdoor landmarks %. %, ranging in size from a corner shop to a few hundred meters of a single street. % in Cambridge, UK. 
while the latter depicts 
% The 12 Scenes dataset depicts 
12 smaller rooms in indoor scenes. 
% In addition, w
We also use the Strecha Fountain~\cite{Strecha2008CVPR} and a dataset of castle Dagstuhl in Germany. 

%and an indoor dataset kindly provided by the authors of~\cite{Speciale2019CVPR} for qualitative evaluation. 

%Due to the need to find neighboring lines and the lack of suitable spatial acceleration structures, our approach has a computational complexity that grows quadratically in the number of lines. 
%For the datasets containing more than 500k sparse points, \ie, mostly the outdoor datasets, we thus typically used only around 30\% of all available lines by randomly selecting a subset of them. 

% \PAR{Evaluation protocol.} 
% \TODO{Explain our evaluation measures (accuracy of recovered point cloud and inversion of SfM point clouds).}

\begin{figure*}[!t]
    \centering
    \includegraphics[width=0.88\linewidth]{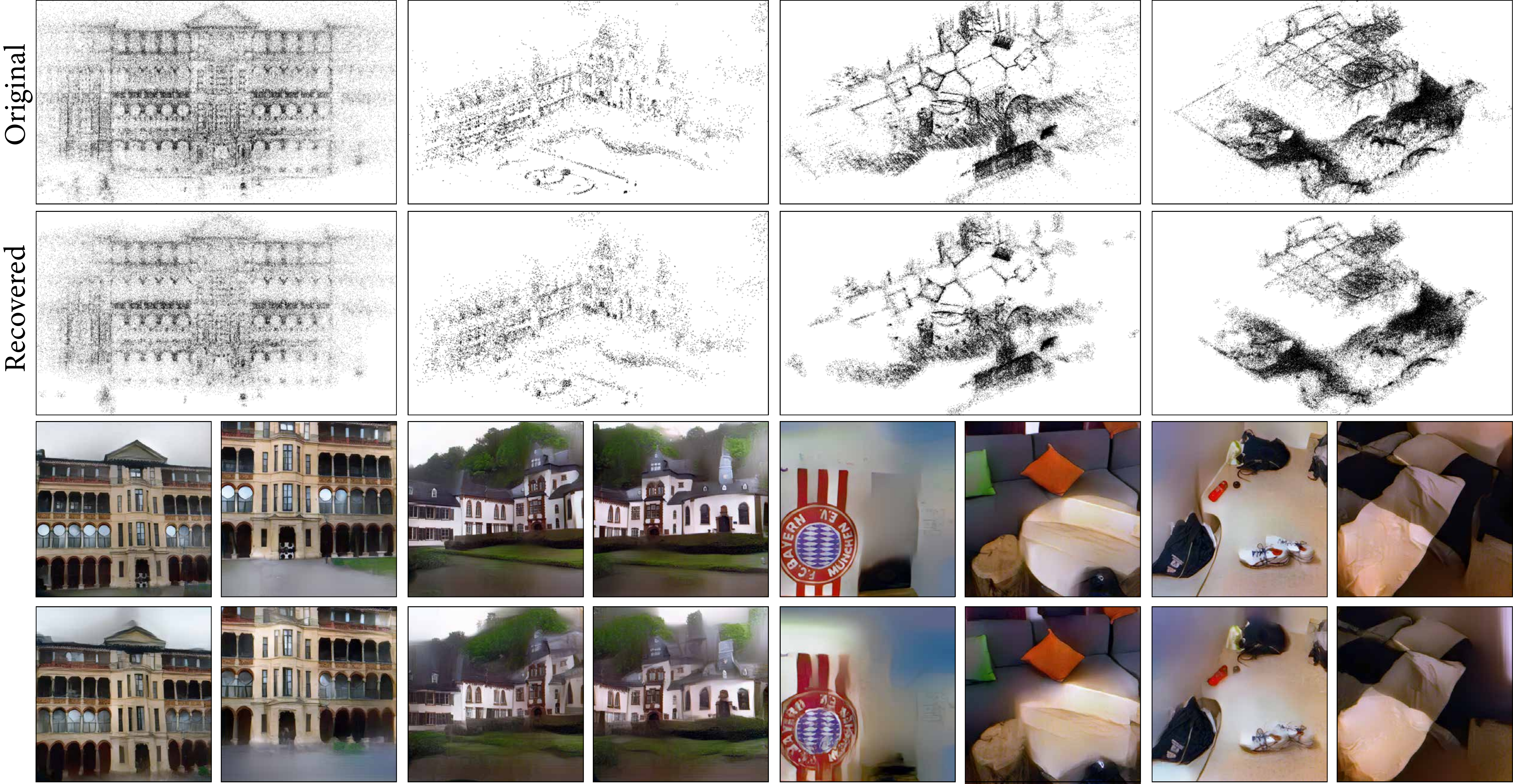}
    \caption{Qualitative results for the recovered point clouds and the images obtained by applying the inversion technique from \cite{Pittaluga2019CVPR} on the original (above) and recovered (below) point clouds. Left to right: 'Old Hospital', 'Dagstuhl', 'Office1-Lounge', 'Apt2-Bed'.}%
    \label{fig:qualitative_inverse}
\end{figure*}

\PAR{Recovering point clouds.} 
Fig.~\ref{fig:qualitative_points} shows the point clouds recovered by our method for two outdoor and two indoor scenes. 
As can be seen, our approach is able to faithfully reproduce the overall structure of the scene. 
At the same time, our method is also able to reveal details such as the presence of the kitchen sink and the circular burner plates in the kitchen scene, or the chequered design of the bed linen in the bedroom scene. 
We observe that regions of higher point density are recovered in more detail. 
This is consistent with the fact that shorter distances between neighboring points increase the chance of accurately recovering the point positions. % to the motivation our algorithm is based on. 
As %theoretically 
predicted in Sec.~\ref{sec:limitations}, 
% In contrast, 
sparser regions are often not well-recovered. 
Rather, the corresponding point estimates fall into regions with a higher point density. 
%One reason for this is that our approach falsely positions points belonging to sparse regions, into regions of high density if their lines happen to pass through the denser region.
% Additional
More qualitative results are shown in Fig.~\ref{fig:qualitative_inverse} and in the supp. material.

% \textcolor{blue}{
% \subsection{Recovered Point Clouds}
% % While what amounts to a breach of privacy in the context of camera localization remains a topic of discussion, in this section we try to show, both qualitatively and quantitatively that in general our approach leads to accurate recoveries of the original scene details.
% Figure \ref{fig:qualitative_points} shows a few examples of the point clouds that our method recovers for both outdoor and indoor scenes. It can be seen that our approach is able to faithfully reproduce the overall structure of the scene, while also revealing details like the presence of kitchen-sink and circular conduction heaters in the kitchen scene and the chequered design of bed-linen in the bedroom scene. It can also be noted that regions of higher point density are recovered in more detail, conforming to the motivation our algorithm is based on. In contrast, sparser regions are often not recovered well. One reason for this is that our approach falsely positions points belonging to sparse regions, into regions of high density if their lines happen to pass through the denser region.}

To quantify the accuracy with which our method recovers point clouds, we measure the Euclidean distance between the original and recovered point positions. 
Fig.~\ref{fig:quantitative}(left) and (middle) show the cumulative distributions of these errors for each of the indoor and outdoor scenes. 
As can be seen, our approach is able to recover a large fraction of the points within 5cm of their original position for the indoor scenes, with median errors in the range of 1-3cm. 
In contrast, the errors are considerably larger for the outdoor scenes. 
This can be explained by the fact that the accuracy with which a point can be recovered depends % goes back to the fact the error depends 
on the underlying distances to its estimated neighbors. 
% In outdoor scenes
Outdoors, where the structure is farther away from the camera than in indoor scenes, these distances are larger. 
Furthermore, we only use 30\% of all lines, selected uniformly at random, for the Cambridge %Landmarks 
scenes for computational efficiency. 
Still, %the results from 
Fig.~\ref{fig:qualitative_points} shows that the accuracy of the recovered points is sufficient to produce human-interpretable point clouds. 

% which is expected to be higher for large-scale scenes. Overall, these results certainly depict the effectiveness of our approach in revealing structural details of the scene obfuscated by the line cloud.

% It should be noted that we used only 30 percent of the lines for recovering scenes from the Cambridge Landmarks dataset, whereas we used all lines for the 12 scenes dataset. While the median error for indoor scenes lies in the range of 1-3 cm, it is much higher for outdoor scenes. 

% \textcolor{blue}{
% To further gauge the performance of our approach, we also analyse the error in terms of the Eucledian distance between the estimated position of each point and its actual position. Fig. \ref{fig:quantitative} (a) and (b) show the cumulative distribution of errors for each of the indoor and outdoor scenes, respectively. It should be noted that we used only 30 percent of the lines for recovering scenes from the Cambridge Landmarks dataset, whereas we used all lines for the 12 scenes dataset. While the median error for indoor scenes lies in the range of 1-3 cm, it is much higher for outdoor scenes. This again goes back to the fact the error depends on the underlying distance to the estimated neighbors, which is expected to be higher for large-scale scenes. Overall, these results certainly depict the effectiveness of our approach in revealing structural details of the scene obfuscated by the line cloud.
% }

\begin{figure*}[!t]
    \centering
    \includegraphics[width=0.9\linewidth]{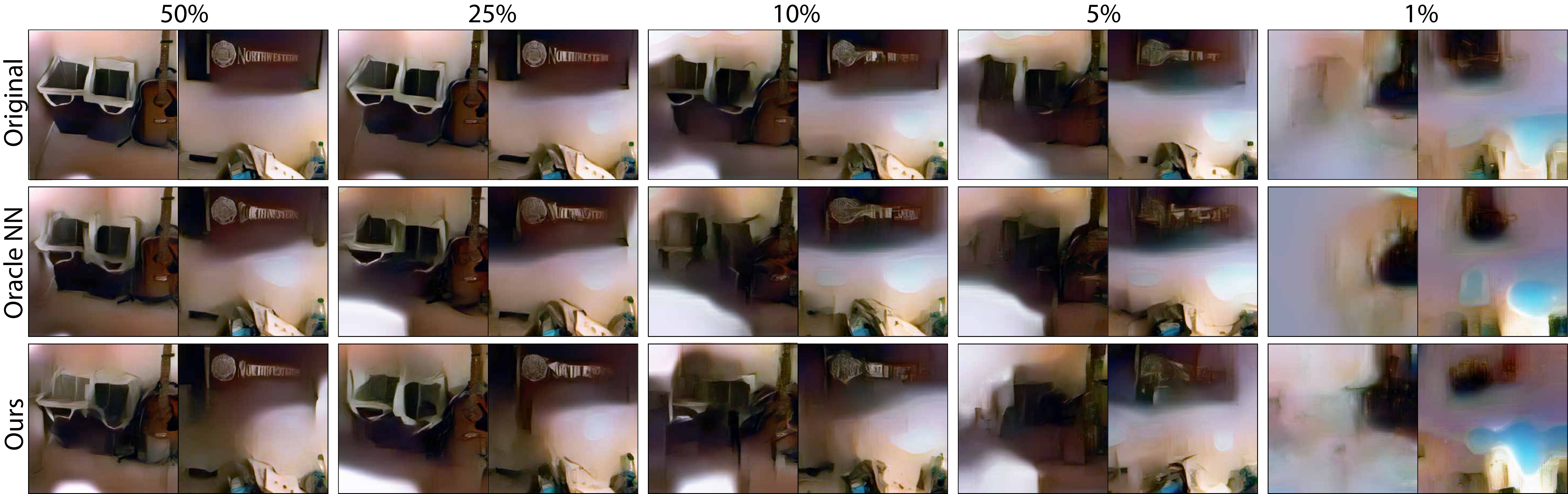}
    \caption{Qualitative results (on scene 'Apt2-Luke' from 12 scenes) for studying the impact of using sparser line clouds to represent the scene. We compare the images obtained by inversion of the original point cloud (top), from a point cloud recovered by our approach using the true neighborhood given by an oracle (middle), and from the point cloud recovered by our method without the oracle (bottom).}%our method of estimating neighborhood. (bottom)}%
    \label{fig:sparsity:qualitative}
\end{figure*}

\PAR{Recovering image details.} 
The main motivation behind using line instead of point clouds was to prevent recovering image content from SfM point clouds~\cite{Speciale2019CVPR}. 
In the next experiment, we thus apply the SfM inversion approach from~\cite{Pittaluga2019CVPR} on our recovered point clouds to see to which degree image content can be recovered from them. 
% Note that w
We apply the pre-trained model from~\cite{Pittaluga2019CVPR}, trained on SfM datasets, without fine-tuning it on our more noisy point clouds. 

Fig.~\ref{fig:qualitative_inverse} shows qualitative results for two outdoor and two indoor scenes, comparing the inversion results of the original and our recovered point clouds. %results of applying the inversion procedure from~\cite{Pittaluga2019CVPR} on our recovered point clouds and compares them with the level of quality that can be achieved when using the original point clouds. 
As can be expected, our more noisy point clouds lead to 
% the images from recovered point clouds produce certain
artifacts such as blurry regions and wavy lines instead of straight ones. 
% 
% This is because we are using the inversion network out of the box, which has been trained over much less noisy data. In spite of this
Still, the obtained images clearly %observe that the images from recovered point clouds in Fig. ~\ref{fig:qualitative_inverse} also 
reveal scene information like the overall structure of the  buildings, shapes of doors and windows, as well as the presence of small objects such as cushions, shoes, a Bayern Munich flag, \etc 
Based on these results, we conclude that lifting point clouds to line clouds does not guarantee that image details cannot be recovered. 
% Textual content, brand names and other such fine details are also possible to be recovered in certain cases.     
Additional qualitative results can be found in the supp. mat.

% \textcolor{blue}{
% \subsection{Reconstructed Images}
% Applying the inversion procedure of \cite{Pittaluga2019CVPR} on the recovered point clouds often produces meaningful images further revealing contents of the scene.
% Fig.~\ref{fig:qualitative_inverse} show results of applying the inversion procedure from~\cite{Pittaluga2019CVPR} on our recovered point clouds and compares them with the level of quality that can be achieved when using the original point clouds. As can be expected, the images from recovered point clouds produce certain artifacts like blurry regions and wavy lines instead of straight ones. This is because we are using the inversion network out of the box, which has been trained over much less noisy data. In spite of this, one can clearly observe that the images from recovered point clouds in Fig. ~\ref{fig:qualitative_inverse} also reveal scene information like the overall structure of building, shapes of doors and windows, presence of small objects such as cushions, shoes, a Bayern Munich flag etc. Textual content, brand names and other such fine details are also possible to be recovered in certain cases.     
% }

\PAR{(Very) sparse line clouds prevent image recovery.} 
As discussed in Sec.~\ref{sec:limitations}, using sparser line clouds still allows accurate localization while potentially preventing our approach from recovering accurate point clouds. 
In this experiment, following~\cite{Speciale2019CVPR}, we thus consider sparse line clouds obtained by randomly selecting a subset of the lines. 

Fig.~\ref{fig:quantitative}(right) shows quantitative results averaged over four indoor scenes for different levels of sparsity. 
We show results for our method and a variant that receives the true neighbors from an oracle. 
As can be seen, our approach can recover the point positions rather accurately when using down to 5\% of the original line cloud. 
However, there is a significant drop in accuracy when using only 1\% of all lines. 
The comparison with the oracle shows that there is considerable room for improvement in terms of better recovering the true neighborhood of each point / line. 
However, even using the oracle still results in a considerable drop in performance when using 1\% of all lines. 

Fig.~\ref{fig:sparsity:qualitative} shows qualitative results for the images obtained via~\cite{Pittaluga2019CVPR} from the recovered point clouds. 
While Fig.~\ref{fig:quantitative}(right) shows that our method provides rather accurate position estimates even at densities as low as 10\% and 5\%, the resulting images show very little details as compared to images from the original sparse point cloud. 
Part of the reason is that points from sparser regions are wrongly recovered in denser regions, causing problems for the inversion process. 
At the same time, sparser point clouds have been shown to be harder to recover in the first place~\cite{Pittaluga2019CVPR}. 
% The knowledge of true neighborhood does not help much in case of low densities because of the increased distance to neighbors leading to increased uncertainty in estimating position.
As can be seen, meaningful image details cannot be recovered when using 5\% or less of the lines, even when given the true neighborhood by some oracle. 
Based on the results, we conclude that sparse line clouds can effectively prevent the recovery of image content using existing methods. 
Additional qualitative results can be found in the supp. mat.

\vspace{-4pt}
\section{Conclusion}
\label{sec:conclusion}
In this paper, we have shown that lifting point clouds to line clouds does not necessarily obfuscate the underlying 3D scene structure. 
% 
% have considered the problem of privacy preserving scene representations in the context of visual localization. 
% In particular, we considered a recent claim~\cite{Speciale2019CVPR} that lifting SfM point clouds to line clouds preserves privacy as it prevents point cloud-to-image translation attacks such as~\cite{Pittaluga2019CVPR}. 
We have shown that it is possible to (approximately) recover the underlying 3D point cloud by identifying local neighborhoods. %, which in turn are used to provide point position estimates for each line. 
In turn, these neighborhoods are used to obtain point position estimates using closest points between lines, which we have shown to often provide a good approximation to the original 3D points. % that gave rise to these lines.
Quantitative and qualitative results show that our approach enables us to recover image details from line clouds. 
However, our results also show that sparsification can effectively prevent recovering image details. % (at least using existing recovery methods). 
In the context of privacy-preserving visual localization, we thus conclude that using lines alone does not guarantee privacy-preservation, but that using sparse representations is similarly important. 
This conclusion is based on the ability of existing SfM inversion methods to recover images from sparse and noisy point clouds. 
Yet, better recovery algorithms might require us to revisit this problem in the future. 

As shown in Fig.~\ref{fig:sparsity:qualitative}, using very sparse point clouds is also effective in preventing~\cite{Pittaluga2019CVPR} from recovering image details. 
Since point-based localization methods are more accurate than line-based ones~\cite{Speciale2019CVPR}, %determining whether there is a 
finding sparsity levels at which point clouds are privacy-preserving while enabling more accurate pose estimates %the trade-off between image quality and pose accuracy for sparse point-/line-clouds 
is an interesting direction for future work. 
Further, our results show room for improvement for our method. 
Including descriptor information into the recovery process might help unlock this potential. 

As detailed in Sec.~\ref{sec:limitations}, exploring different distributions for line directions is another interesting research direction. Potential approaches to define the distribution include:
1) ensuring that there are no closest points between lines that are within a given threshold of the original points. 
2) creating line neighborhoods that do not contain closest points, \eg, by ensuring that lines from many far away points come very close to each other %the line of a given point and that this happens 
far away from the original point position.  
3) ensuring that most lines go through dense neighborhoods, thus allowing to recover these parts while preventing recovery in all the sparser parts. 
Computing such distribution likely cannot be done individually per point anymore but rather requires complex iterative schemes. % and cannot be done individ
% We believe computing these distributions is not easy. Line directions would depend on other lines, leading to an iterative scheme. 
% At the same time, line directions should still allow for accurate pose estimation.
% 
% Note that the distributions above are not necessarily privacy-preserving. 
At the same time, iteratively applying our approach with a human in-the-loop might allow us to still handle such distributions: 
after visual inspection, the human  removes lines from the neighborhoods of falsely classified points. %, \eg, by excluding lines passing through certain regions in 3D space. 
Similarly, co-occurrence statistics over feature descriptors could be used to filter out irrelevant neighbors. %This suggests that defining privacy-preserving line distributions is not trivial and requires further research.
{
\small{
\PAR{Acknowledgements.} This work has been funded by a Google Faculty Research Award, the Chalmers AI Research Centre (VisLocLearn), the EU Horizon 2020 project RICAIP (grant agreeement No 857306), and the European Regional Development Fund under IMPACT No.~CZ.02.1.01/0.0/0.0/15 003/0000468.
}
}

\appendix
\appendixpage

This provides the following information:
Sec.~\ref{sec:peak_finding} provides additional details about the \textbf{peak finding} step of our algorithm (\cf Sec. 4.2 in the main paper). %In particular, we add a figure that aids in understanding our peak-finding algorithm.
Sec.~\ref{sec:qualitative} provides additional details for the experiments conducted in the main paper, including more qualitative results (\cf Sec.~5 in the main paper). %shows results from further experiments conducted over the datasets considered in the main paper. More specifically, 
% \begin{enumerate}
%     \item In the case when a point's true neighborhood is provided by an oracle, the impact of the size of this neighborhood on the recovery algorithm is analysed. 
%     \item The impact of outliers in the neighborhood set as presented for a single scene in Fig. ~\ref{fig:nns_exp} of the main paper, is studied for more scenes.  
% \end{enumerate}

\section{Peak Finding Algorithm}\label{sec:peak_finding}
In the following, we provide additional details on the peak finding algorithm used in our approach. 

The input to the peak finding algorithm is a set of candidates for the 3D point position $\mathbf{p}_i$. These candidates are obtained as the closest points on the line $\mathbf{l}_i$ to the lines contained in the neighborhood $\mathcal{N}^K(\mathbf{l}_i)$ provided by the previous stage of our algorithm (Neighborhood Estimation, \cf Sec. 4.2 in the main paper). 
Each candidate is parameterized as a scalar value $\beta_{ij}$ that provides a 3D point position on the line $\mathbf{l}_i = \mathbf{o}_i + \beta \mathbf{v}_i$. 
Here, $\mathbf{o}_i$ is any point on the line and $\mathbf{v}_i$ is a unit vector describing the direction of the line. The set can thus be written as:
\begin{equation}\label{eq:estimate_set}
\mathcal{E}_i = \{ \beta_{ij} | \mathbf{\hat{p}}_{ij} = \mathbf{o}_i + \beta_{ij} \mathbf{v}_i , \forall j \in \mathcal{N}^K(i)\} \enspace .   
\end{equation}
We use these candidates to compute the unweighted empirical cumulative distribution function (CDF) of the candidates along the line as  
\begin{equation}
    F_{i}(x) = \frac{1}{K} \sum\nolimits_{j = 1}^{K}{I_{\beta_{ij} < x}} \enspace .      
\end{equation}
As described in the main paper, $I_{\beta_{ij} < x}$ is an indicator variable taking value 1 if ${\beta_{ij} < x}$ and 0 otherwise. 
This CDF is the compared against the CDF $F_U(x)$ of a uniform distribution of points along the line.\footnote{For practical reasons, we only consider the interval between the minimum and maximum values from $\mathcal{E}_i$ when computing $F_U(x)$.}

As described in the main paper, the Kuiper's statistic (KS) is used to compare the two CDFs. 
More precisely, the KS is used to identify regions where both CDFs differ the most. 
Intuitively, these regions correspond to intervals along the line where there is a higher density of candidates than can be accounted for by a uniform distribution.

\begin{figure}[t!]
    \centering
    \includegraphics[width=0.9\linewidth]{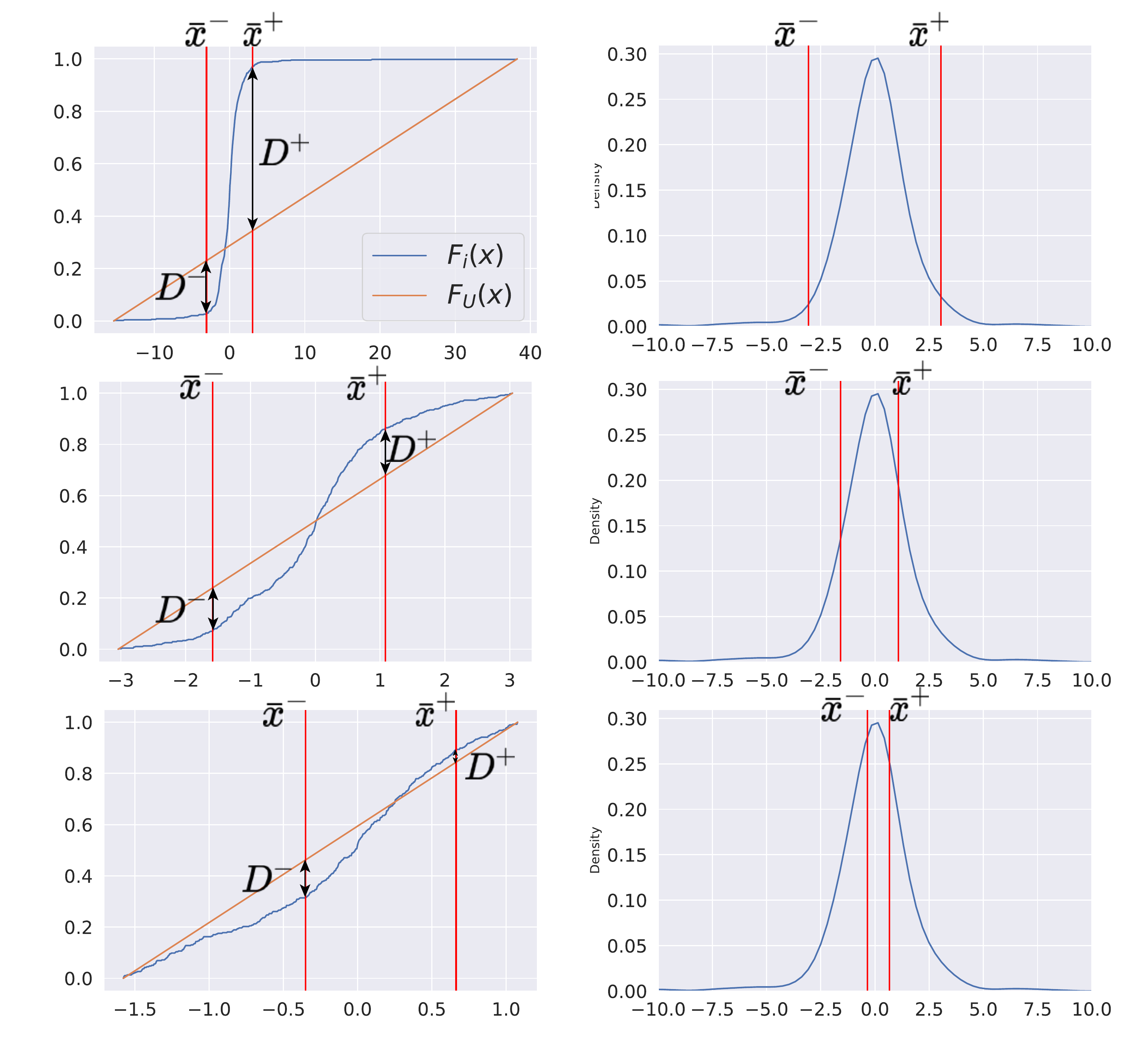}
    \caption{Illustrative example of our peak finding algorithm for a KS threshold of 0.3. The rows from top to bottom indicate three subsequent iterations of our approach, starting from the top. Left column: empirical cumulative distribution function (CDF) $\textbf{F}_{i}(x)$ of the point position candidates and the CDF of a Uniform distribution $\textbf{F}_{U}(x)$. We show the two points $\bar{x}^{-}$ and $\bar{x}^{+}$ defining the interval via red lines and also illustrate the meaning of $D^{-}$ and $D^{+}$. Right column: the approximate density distribution of the candidates along the line, together with the intervals defined by $\bar{x}^{-}$ and $\bar{x}^{+}$. As can be seen, the peak finding approach iteratively narrows down the interval towards the peak of the density distribution.}
    \label{fig:peak_fig}
\end{figure}

\begin{table*}
\begin{center}
\begin{tabular}{|l|c|c|c|}
\hline
Scene & Point Density & Iteration-1 (Coarse Estimation) &  Iteration 2 and later (Refinement)\\
\hline\hline
Outdoor Scenes from \cite{Kendall2015ICCV} & more than5\%  & 500 & 200 \\
Outdoor Scenes from \cite{Kendall2015ICCV} & 5\% or less  & 100 & 50\\
Indoor Scenes from \cite{Shotton2013CVPR} &  more than 5\% & 250 & 100\\
Indoor Scenes from \cite{Shotton2013CVPR} &  5\% or less & 50 & 25\\
\hline
\end{tabular}
\end{center}
\caption{Information about the number of nearest neighbors used for estimating point positions in different scenes, under varying densities of input line cloud and at different stages of estimation.}
\label{tab:num_nns}
\end{table*}

As detailed in the main paper, we compute the two points $\bar{x}^{-} = \argmax_{x} \left(F_{U}(x) - F_{i}(x)\right)$ and $\bar{x}^{+} = \argmax_{x} \left(F_{i}(x) - F_{U}(x)\right)$ corresponding to the positions along the line where the two distributions differ most. 
The differences between the distributions at these points are given as $D^{-} = \left(F_{U}(\bar{x}^{-}) - F_{i}(\bar{x}^{-})\right)$ and $D^{+} = \left(F_{i}(\bar{x}^{+}) - F_{U}(\bar{x}^{+})\right)$ and the Kuiper's statistic is then defined as $\text{KS} = D^{-} + D^{+}$. 
As illustrated in Fig.~\ref{fig:peak_fig}, we recursively use this process to find regions of high density: 
applying the process to the interval defined by $\bar{x}^{-}$ and $\bar{x}^{+}$ (shown via the red lines in the figure), the points $\bar{x}^{-}$ and $\bar{x}^{+}$ as well as the Kuiper's Statistic are re-estimated within this interval. %are re-computed for  interval defined by $\bar{x}^{-}$ and $\bar{x}^{+}$ 
Once the KS falls below a given threshold, set to 0.3 for the example shown in Fig.~\ref{fig:peak_fig}, the recursion is aborted. In practice, we observe that a KS threshold in the range of 0.3 - 0.4 performs well.

The right column of Fig.~\ref{fig:peak_fig} shows the density distribution along the line for the example considered in the figure. 
As can be seen, our peak finding approach iteratively shrinks the interval towards the peak of the density distribution.  
% of 0.3, it can be seen that the algorithm is able to find the region very close to the actual peak. 

% we search for the 
% In particular, we find the region of peak density in the distribution of estimates along the line. Fig. ~\ref{fig:peak_fig} shows an example of the performance of the peak finding algorithm for a KS threshold of 0.3. The red lines on the cumulative distribution plot correspond to $\bar{x}^{-}$ and $\bar{x}^{+}$, respectively. 
% The estimates lying in the range $(\bar{x}^{-},\bar{x}^{+})$ are then used in the next iteration to further move closer to the actual peak point. The figures on the left also show $\textbf{D}^{-}$ and $\textbf{D}^{+}$ where the Kuiper's statistic value,

% \begin{equation}
% \textbf{KS} = \textbf{D}^{-} + \textbf{D}^{+}.     
% \end{equation}
% As KS falls below the threshold value of 0.3, it can be seen that the algorithm is able to find the region very close to the actual peak. In practice, we observe that a KS threshold in the range of 0.3 - 0.4 performs well.    

There are also instances where multiple peaks are obtained in the distribution of estimates. 
Such cases correspond to, \eg, situations where a line passes through more than one region in the scene that contains many of the original 3D points. % the actual scene. 
To handle such cases, we break at points where $F_U(x)$ intersects $F_i(x)$ from below, \ie $F_U(x) = F_i(x)$ and $F_i(x_{-}) - F_U(x_{-}) > 0$ where $\epsilon = x - x_{-} > 0$ is very small.
Let $\{x_{1}, x_{2} \dots x_{k}\}$ be such points of intersection. Then each of the ranges $(x_{min}, x_{1}), (x_{1}, x_{2}), \dots (x_{k}, x_{max})$ corresponds to a particular peak. 
Among all these detected peaks, the one with the highest $\text{KS}$ value is selected.

% To accommodate such cases, multiple pairs of $(\bar{x}^{-},\bar{x}^{+})$ are obtained and the one with a higher $\textbf{KS}$ is taken to repeat the above process of iterative refining.   
    
\section{Detailed Results}\label{sec:qualitative}
This section provides additional qualitative results (\cf Sec.~5 in the main paper). 
It also contains a more detailed analysis of the impact of the neighborhood size on the quality of the recovered point clouds and images (\cf Sec.~4.1 and Fig.~3 in the main paper). 

\PAR{Impact of the neighborhood size.} 
In the main paper, we showed results for the case where an oracle provides the true neighborhood for each 3D point / line based on the original point cloud. 
Following the example provided in Sec.~4.1, we used the 50 closest neighbors for these experiments. 
In an ablation study, we thus study the impact of varying the number of true neighbors. We also list the number of nearest neighboring lines/points used in our recovery algorithm for different scenes in Table \ref{tab:num_nns}.% number of neighboring lines used for estimation of a point position. we consider the case where the underlying true neighborhood is provided by an oracle. 

\begin{figure*}
    \centering
    \subfloat{\includegraphics[width=0.3\linewidth]{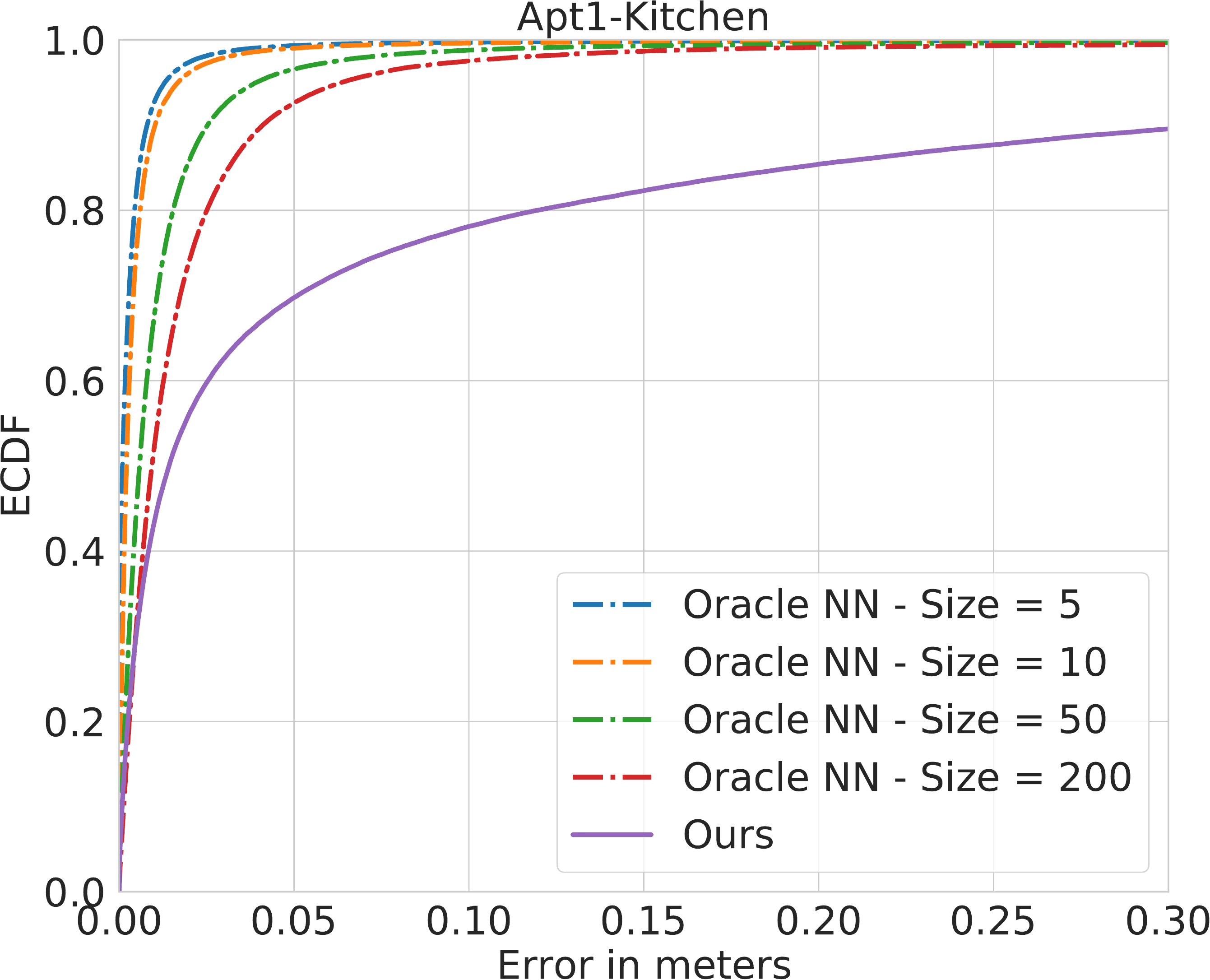}} 
    \hspace{0.5pt}
    \subfloat{\includegraphics[width=0.3\linewidth]{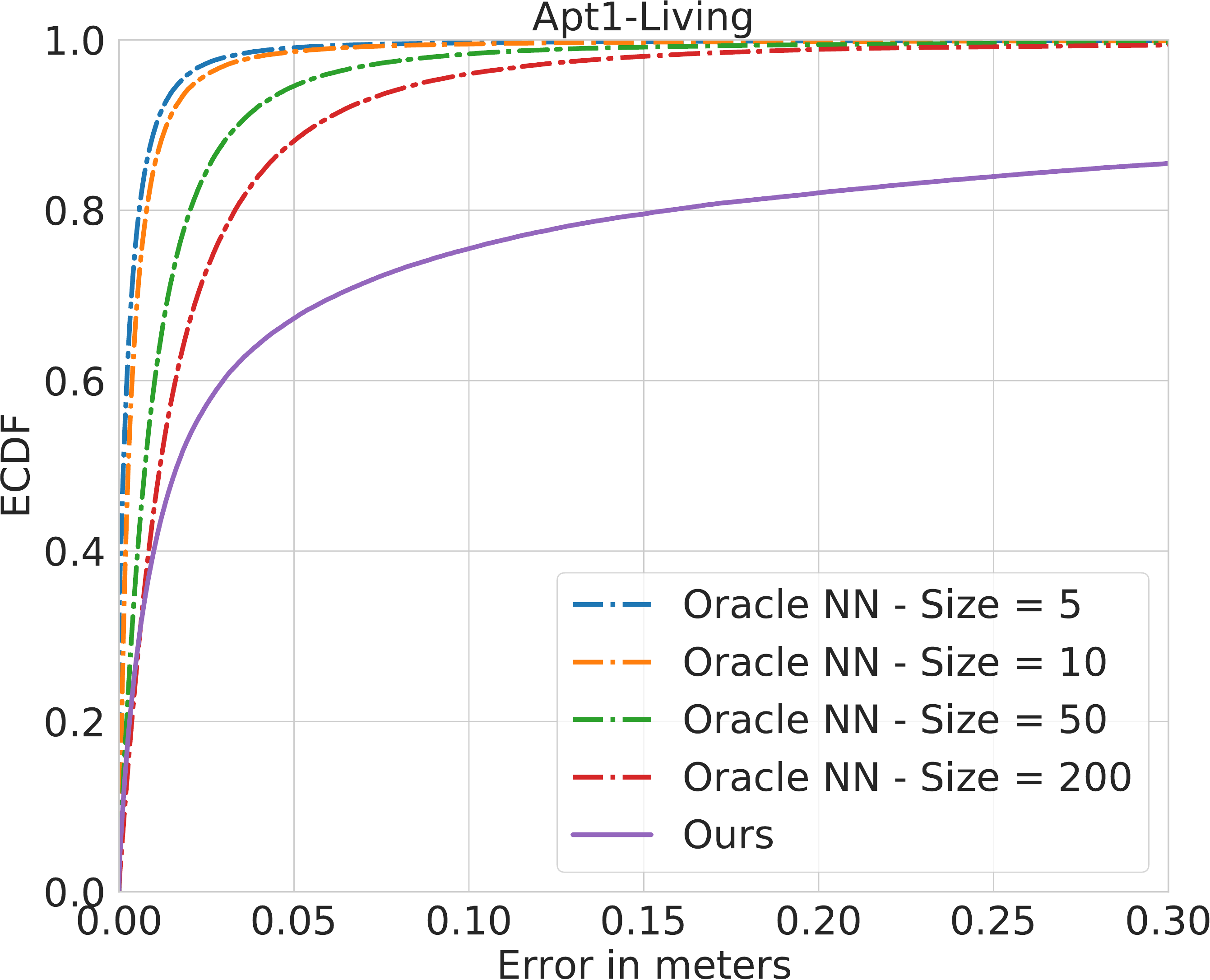}} \hspace{0.5pt}
    \subfloat{\includegraphics[width = 0.3\linewidth]{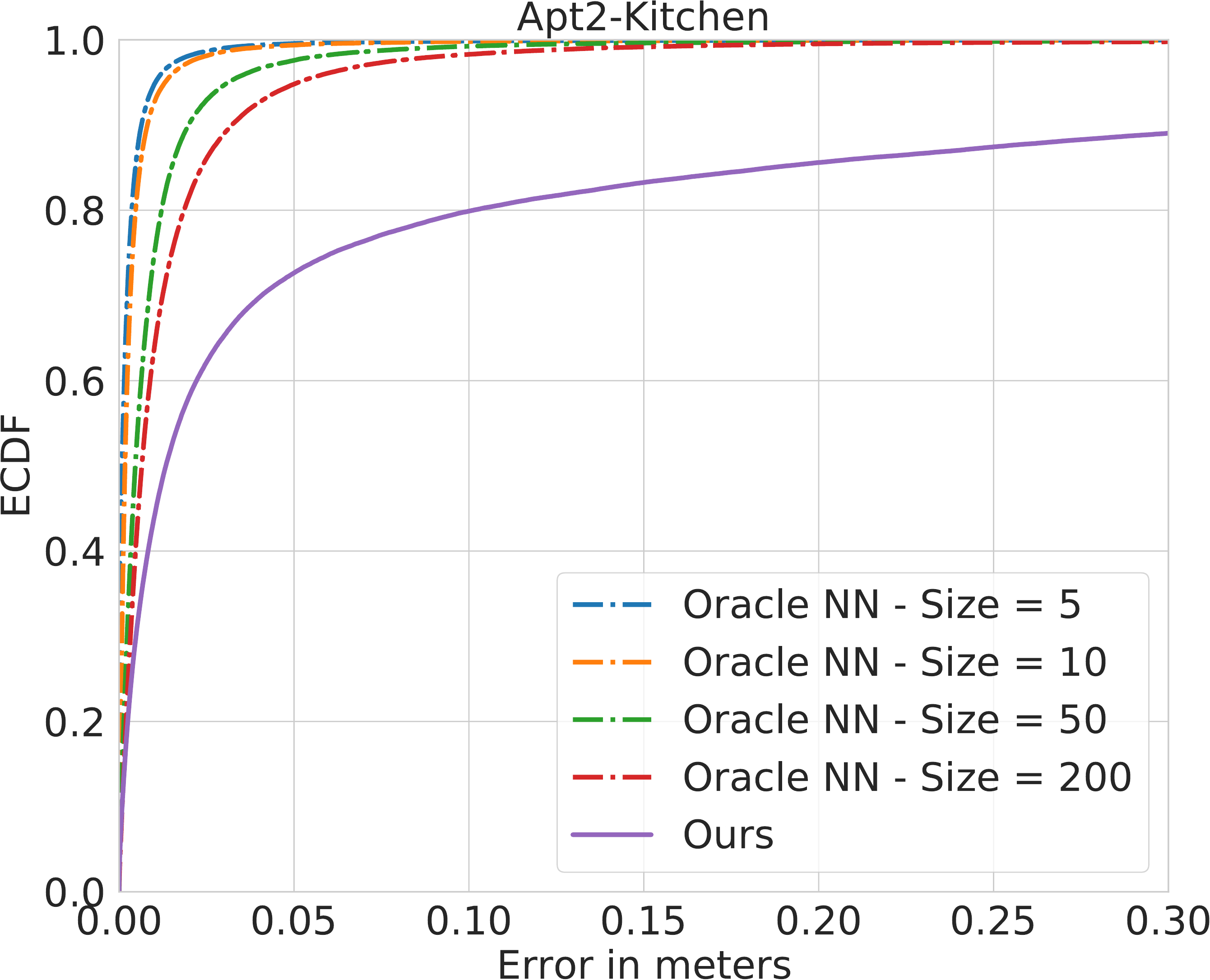}}
    \hspace{0.5pt}
    \centering
    \subfloat{\includegraphics[width = 0.3\linewidth]{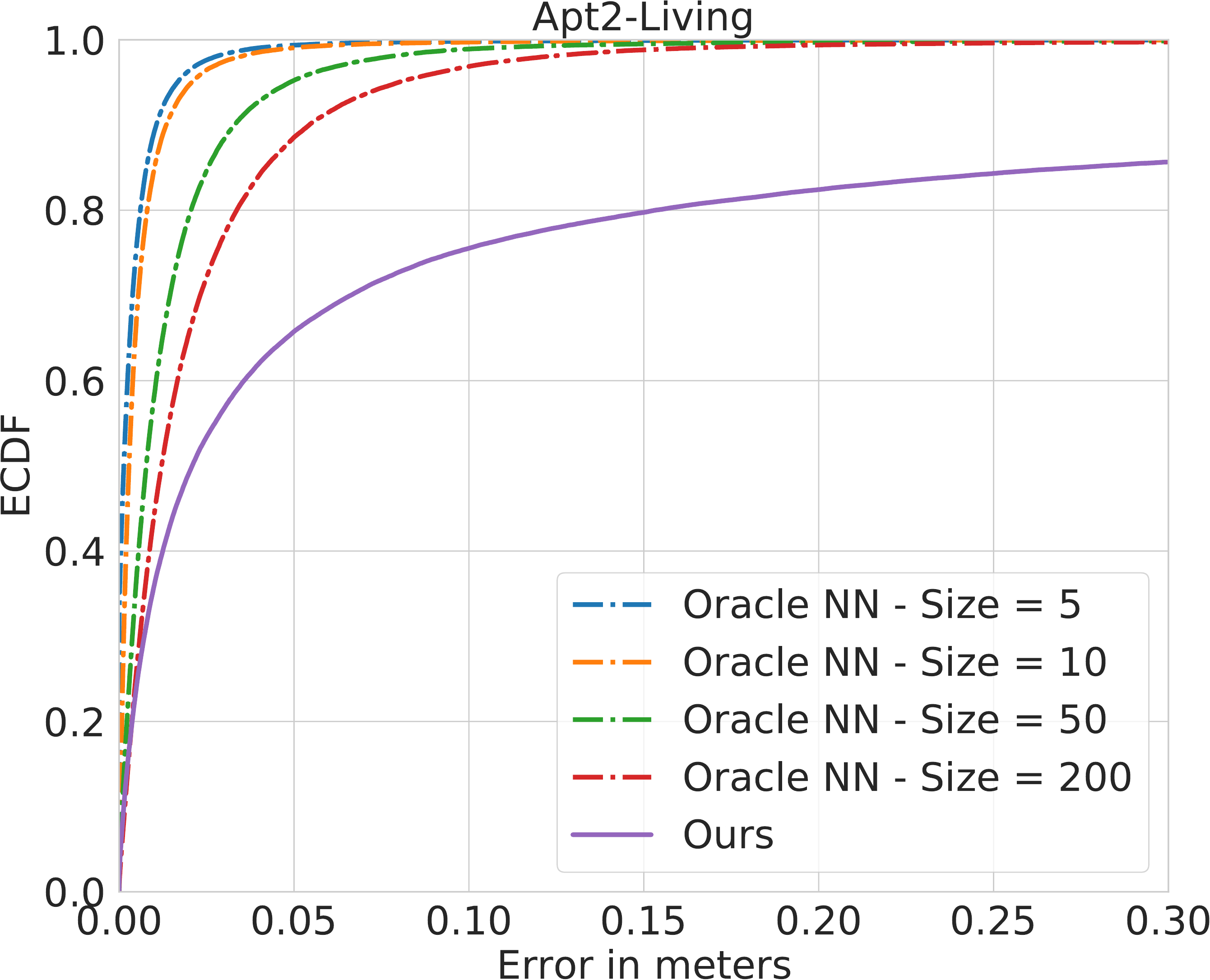}}
    \hspace{0.5pt}
    \centering
    \subfloat{\includegraphics[width = 0.3\linewidth]{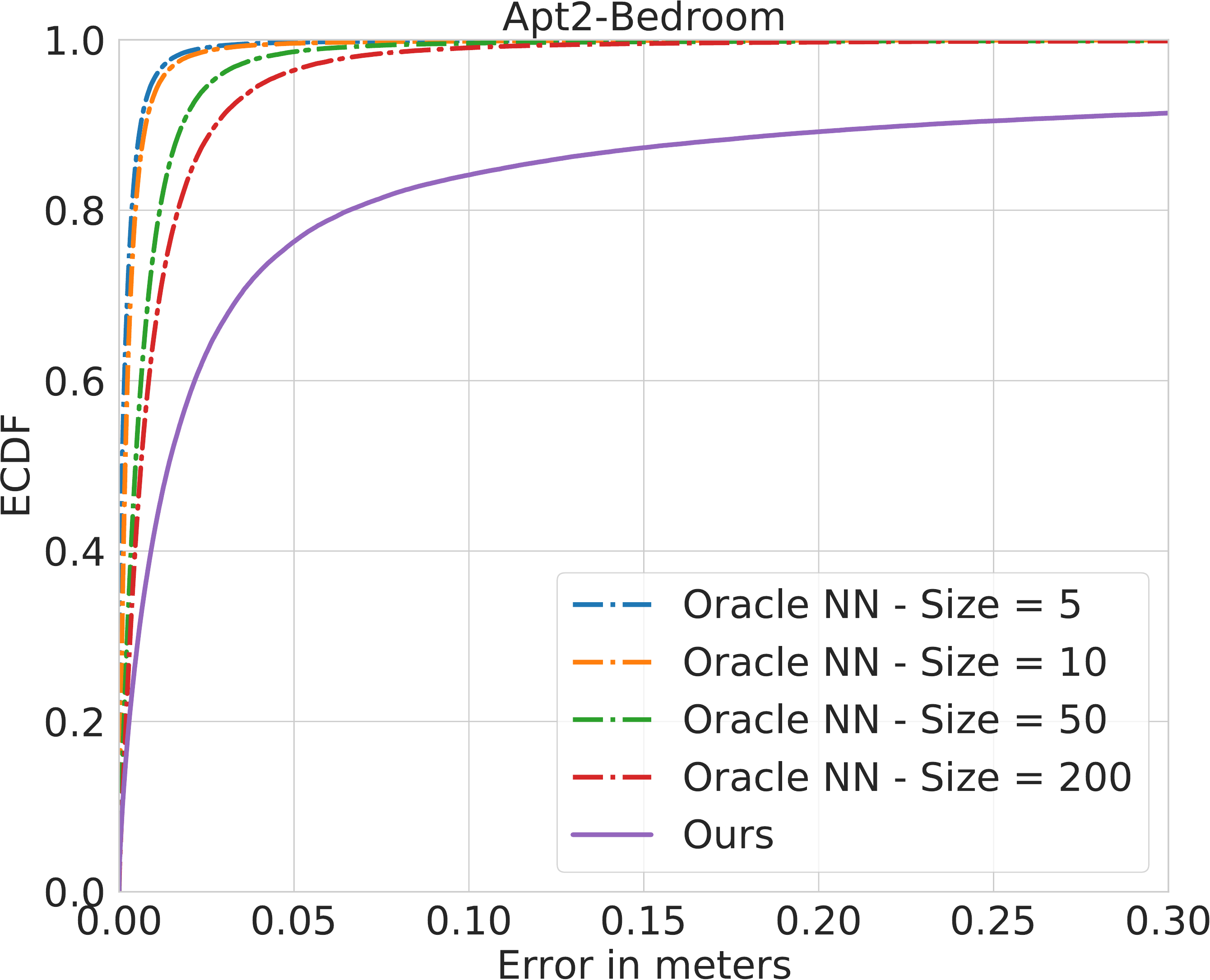}}
    \hspace{0.5pt}
    \centering
    \subfloat{\includegraphics[width = 0.3\linewidth]{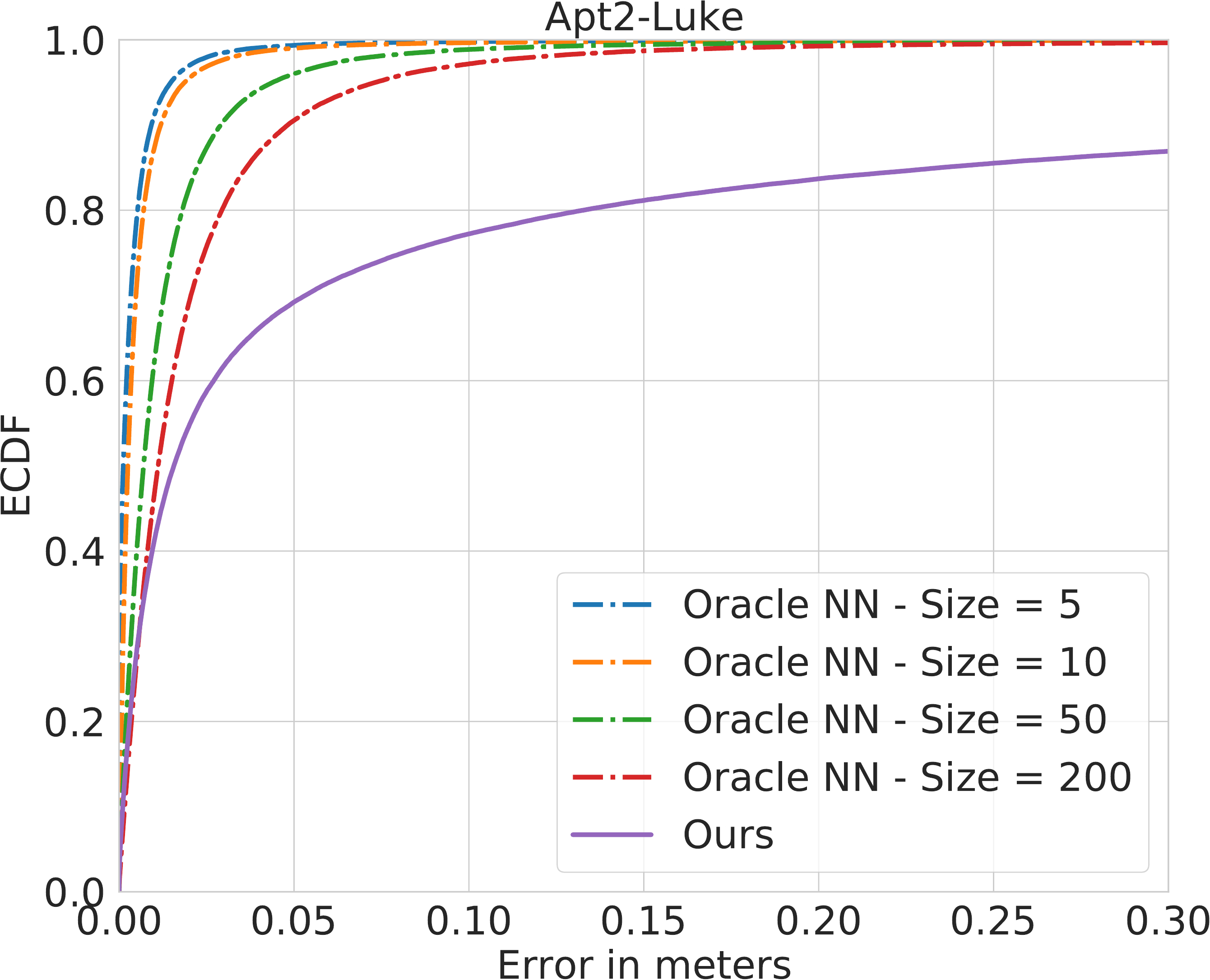}}
    \hspace{0.5pt}
    \centering
    \subfloat{\includegraphics[width = 0.3\linewidth]{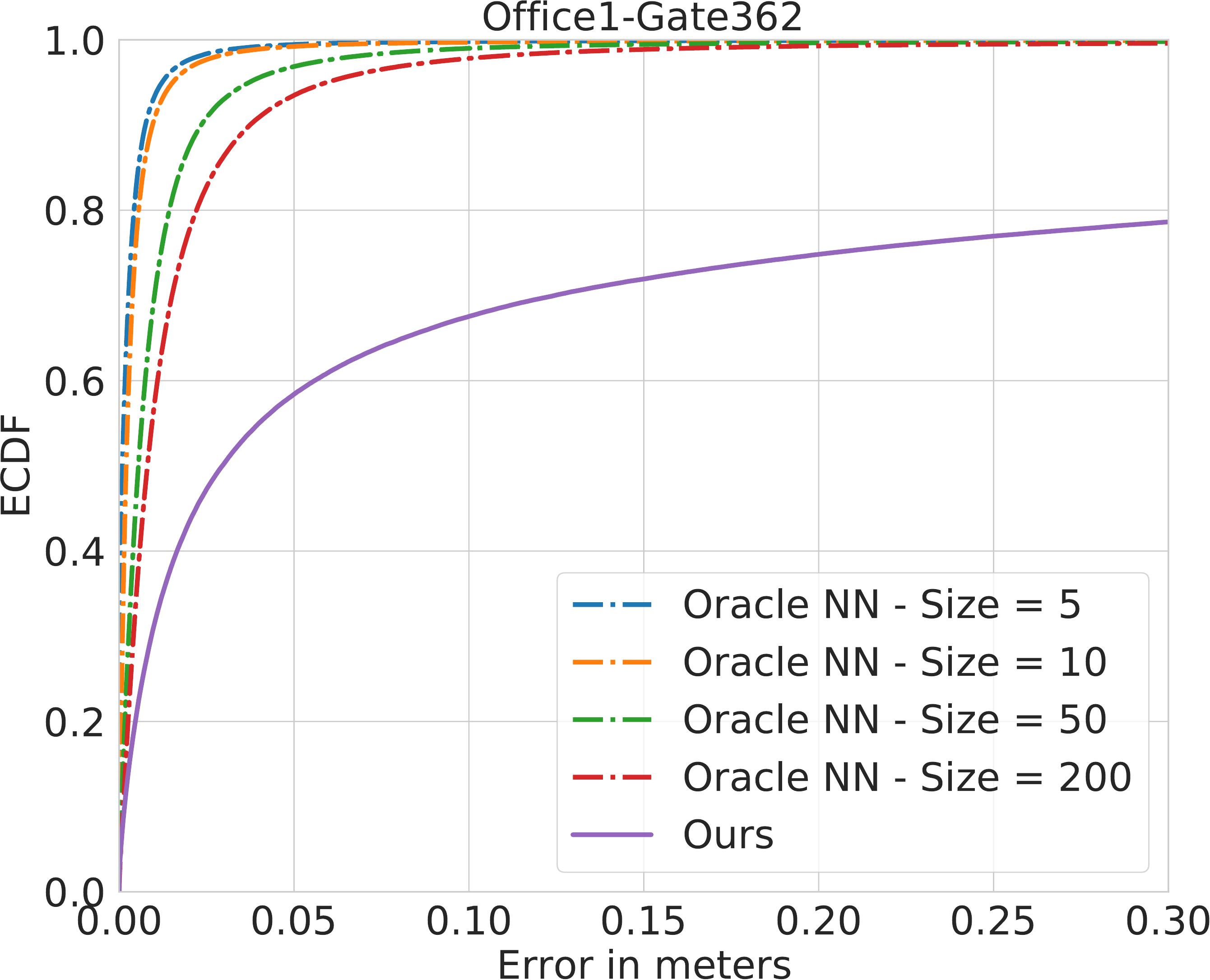}}
    \hspace{0.5pt}
    \centering
    \subfloat{\includegraphics[width = 0.3\linewidth]{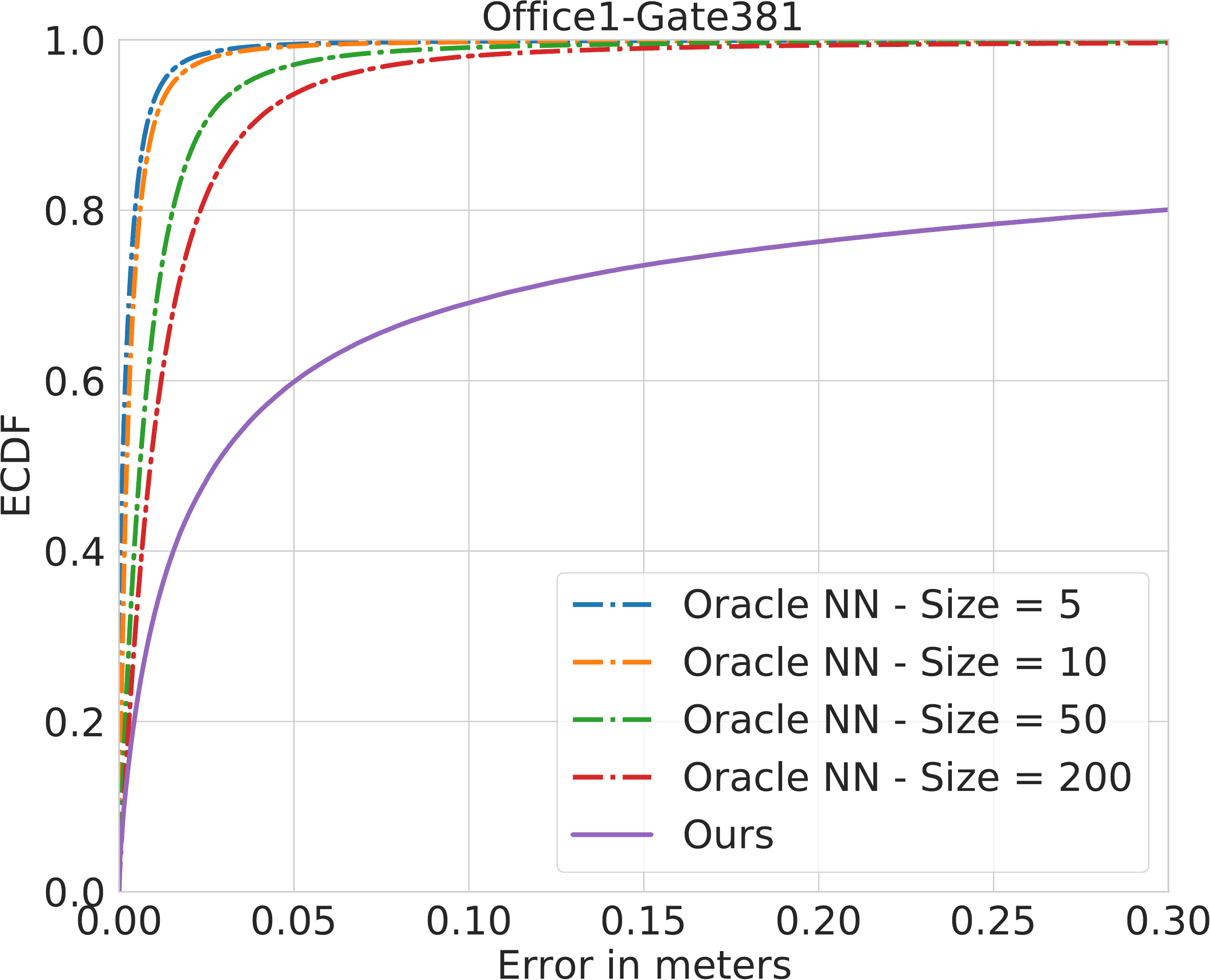}}
    \hspace{0.5pt}
    \centering
    \subfloat{\includegraphics[width = 0.3\linewidth]{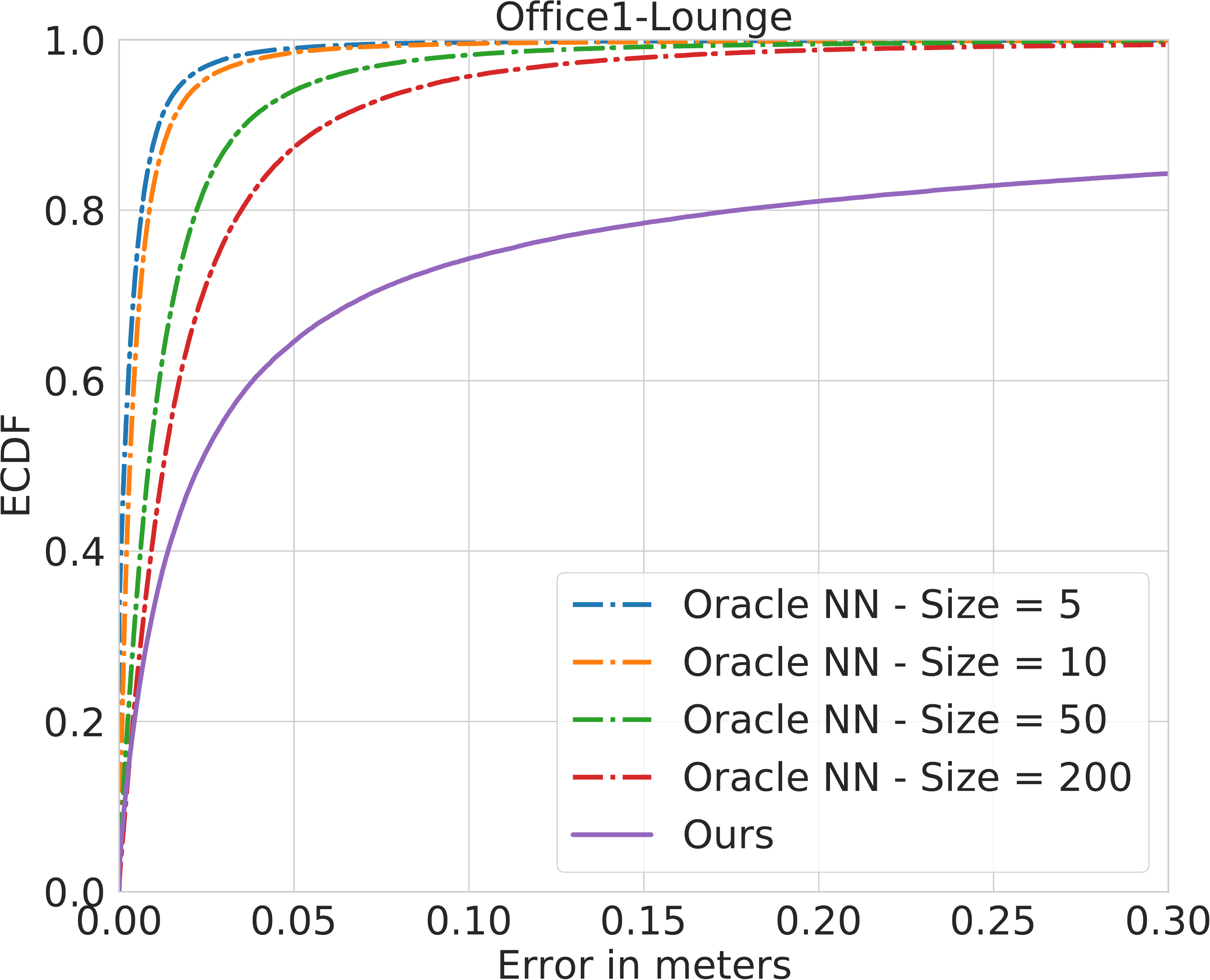}}
    \hspace{0.5pt}
    \centering
    \subfloat{\includegraphics[width = 0.3\linewidth]{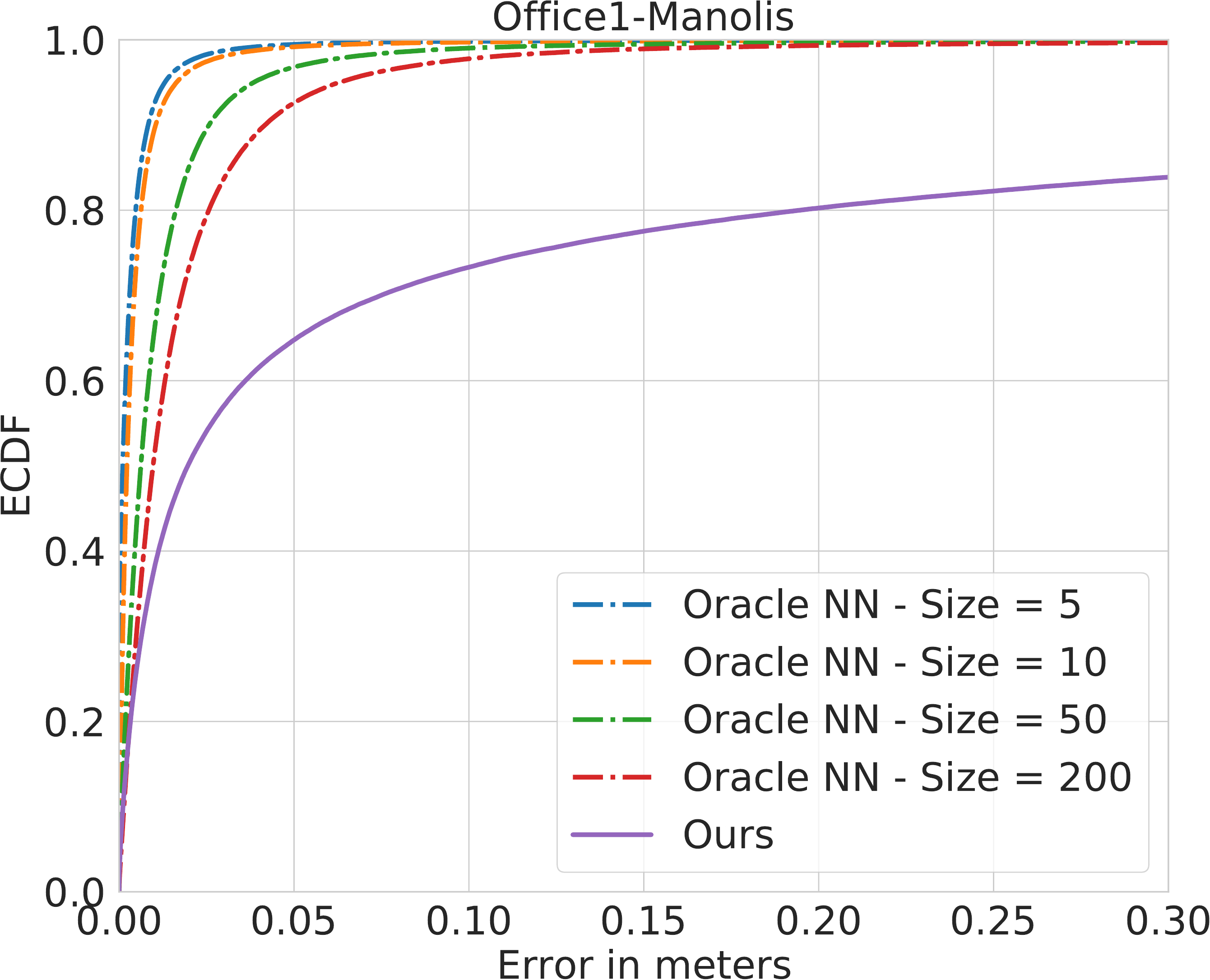}}
    \hspace{0.5pt}
    \centering
    \subfloat{\includegraphics[width = 0.3\linewidth]{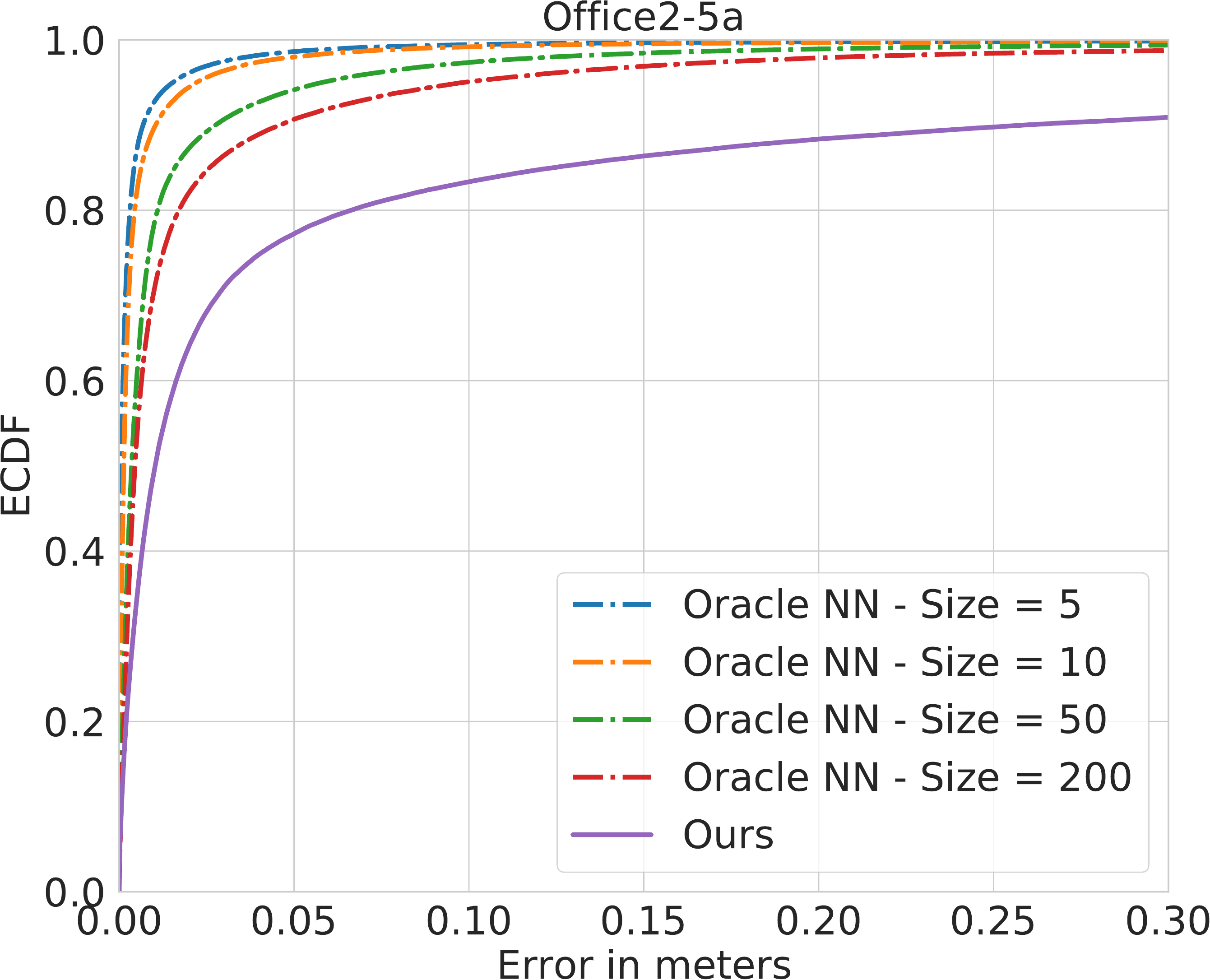}}
    \hspace{0.5pt}
    \centering
    \subfloat{\includegraphics[width = 0.3\linewidth]{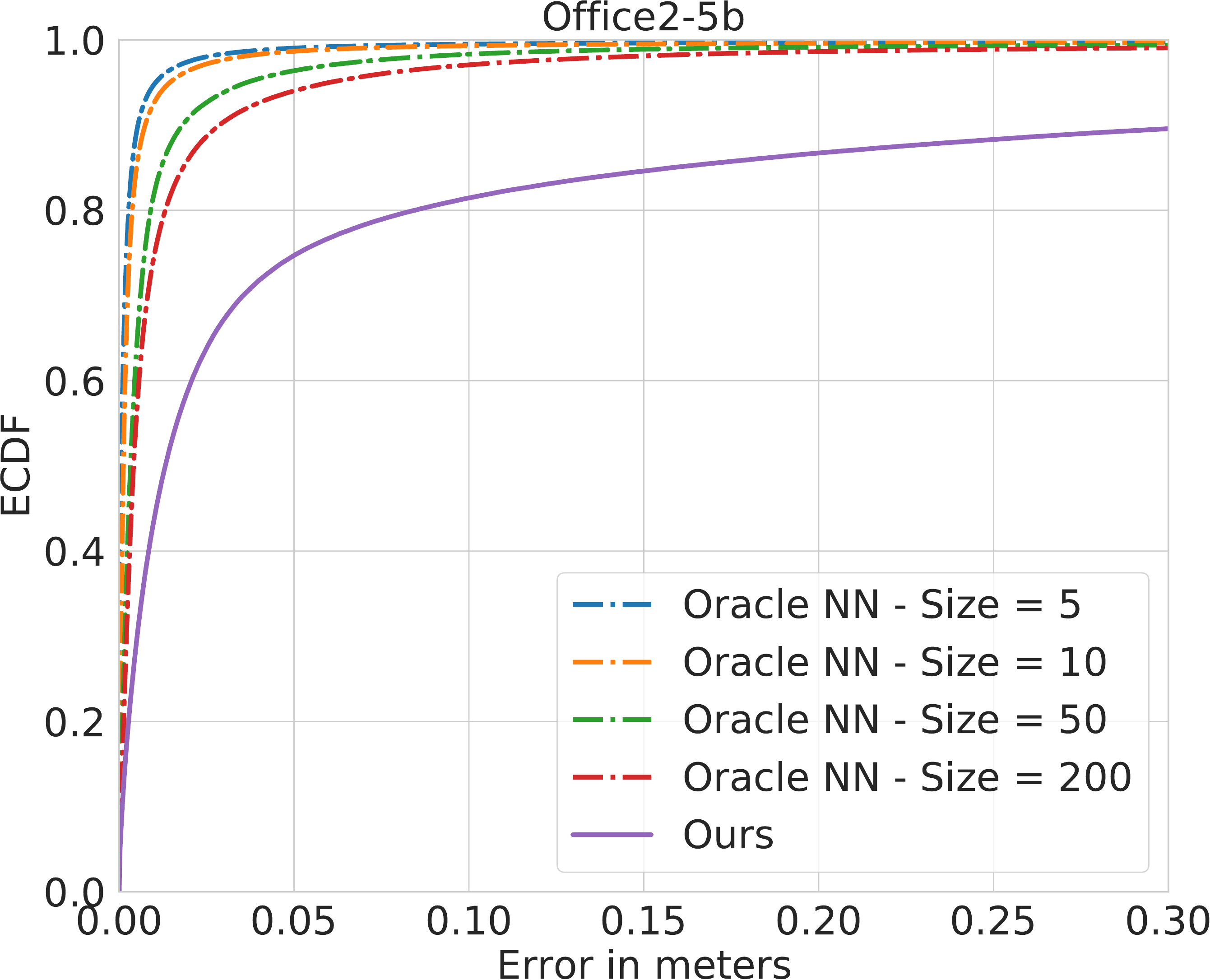}}
    \caption{Quantitative results showing the cumulative distribution of errors in the recovering point positions. We show results obtained on all scenes of the 12 Scenes dataset~\cite{Shotton2013CVPR}, for our approach and when the true neighborhood of each point / line is provided by an oracle. For the latter, we vary the number of neighbors.}%Error distribution of our results and for oracle-provided neighborhoods of different sizes over all scenes from the indoor 12 Scenes dataset~\cite{Shotton2013CVPR}.}
    \label{fig:num_nn_indoors_quantitative}
    
\end{figure*}

Fig.~\ref{fig:num_nn_indoors_quantitative} shows cumulative distributions of the errors in recovering point positions when varying the number of true nearest neighbors (NN) for all scenes from the 12 Scenes dataset~\cite{Shotton2013CVPR}. 
As can be expected from the analysis presented in the main paper, using a smaller neighborhood leads to more accurate point estimates.
% As shown in Fig. ~\ref{fig:num_nn_indoors_quantitative}, as the size of the neighborhood decreases, a lower error is obtained. 
This can be explained by the relation between the error in estimation of a point's position and distance to neighbors used for estimation (\cf Sec.~4.1 in the main paper). 

\begin{figure*}
    \centering
    \includegraphics[width=0.99\linewidth]{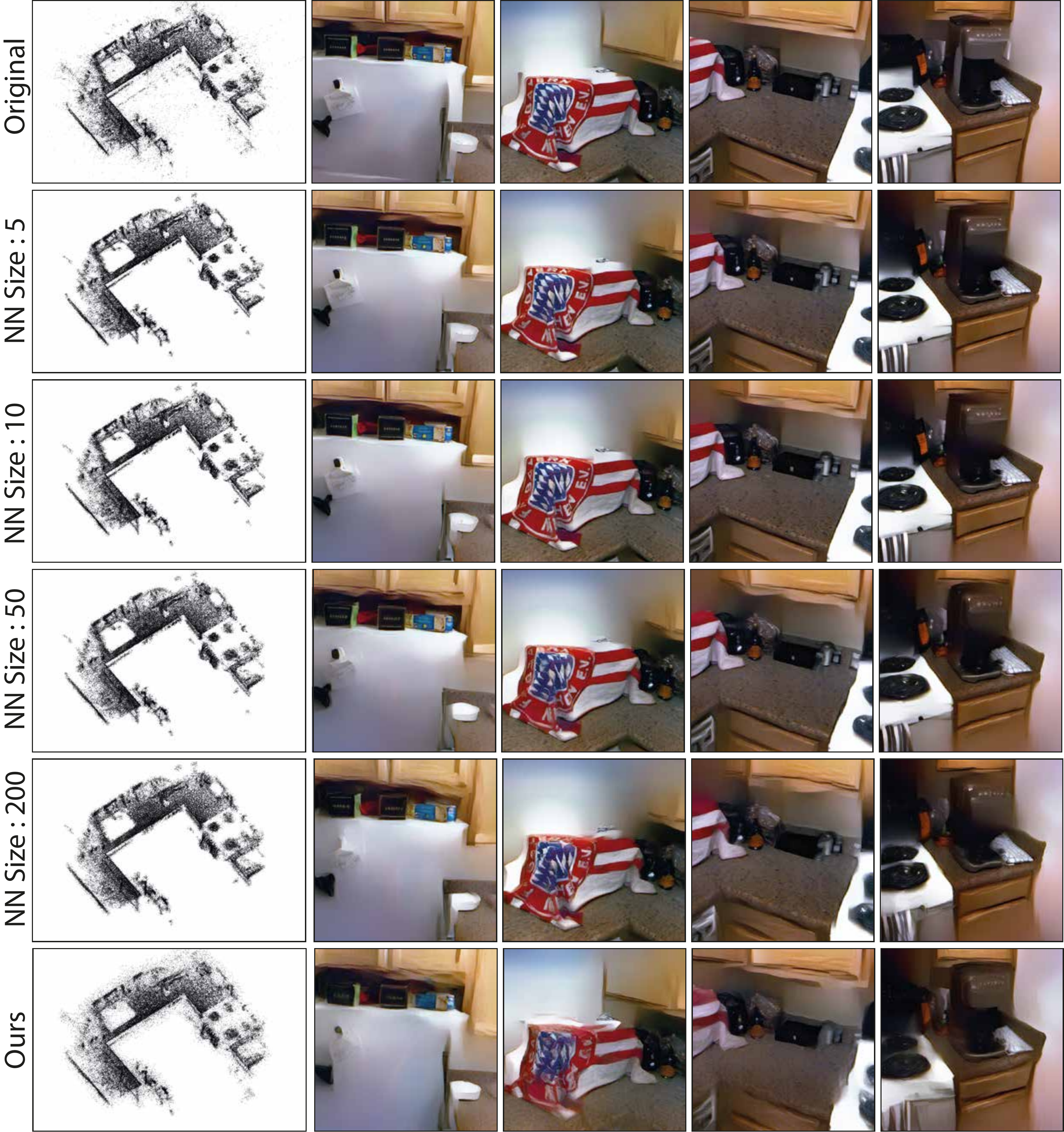}
    \caption{Qualitative results showing the impact of the size of neighborhoods used for estimating the point position. For rows 2 through 5, the true neighborhood is provided by an oracle and we vary the number of used neighbors. The results are shown on the "Apt2-Kitchen" scene from the 12 Scenes dataset~\cite{Shotton2013CVPR}. In addition, we also show the original point cloud and the point cloud recovered by our full approach (which also estimates the neighborhoods). In each case, we show images obtained from the corresponding point clouds via the SfM inversion process from~\cite{Pittaluga2019CVPR}.}
    \label{fig:num_nn_indoors_apt2kitchen}
\end{figure*}

Fig.~\ref{fig:num_nn_indoors_apt2kitchen} shows the point clouds and images recovered using different numbers of nearest neighbors for the \textit{Apt2-Kitchen} scene of the 12 Scenes dataset~\cite{Shotton2013CVPR}. 
As can be seen, the point clouds and images (obtained via the SfM inversion process from~\cite{Pittaluga2019CVPR} applied on our recovered point clouds) are visually similar for all numbers of nearest neighbors. 
This shows that the recovered point clouds do not need to be extremely accurate in order to be able to obtain good quality images. 

% However, the recovered point clouds and the images obtained by the inversion process of ~\cite{} visually appear alike as shown for example over the \textit{Apt2-Kitchen} scene in Fig. \ref{fig:num_nn_indoors_qualitative}.  

\begin{figure*}
    \centering
    \subfloat{\includegraphics[width=0.3\linewidth]{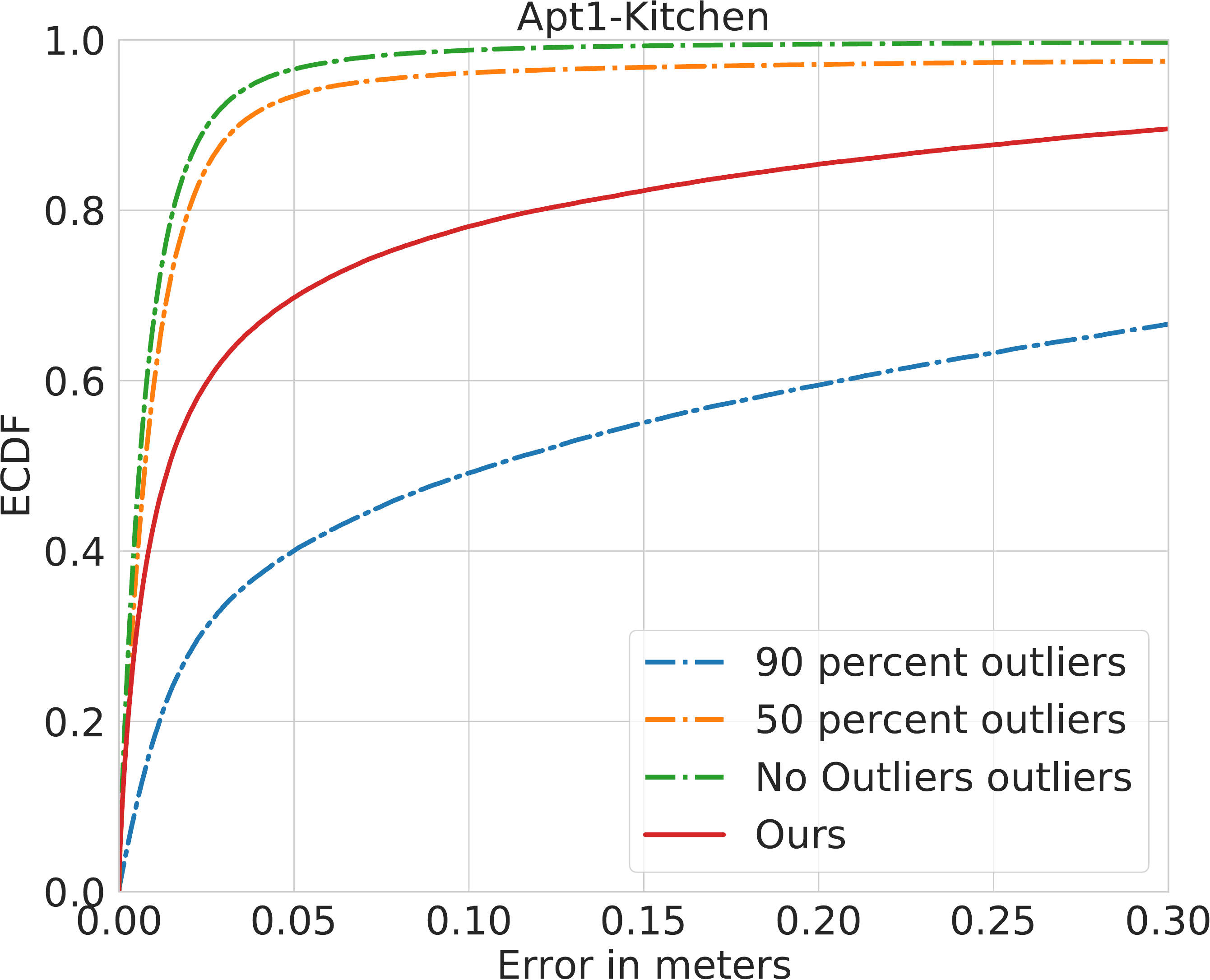}} 
    \hspace{0.5pt}
    \subfloat{\includegraphics[width=0.3\linewidth]{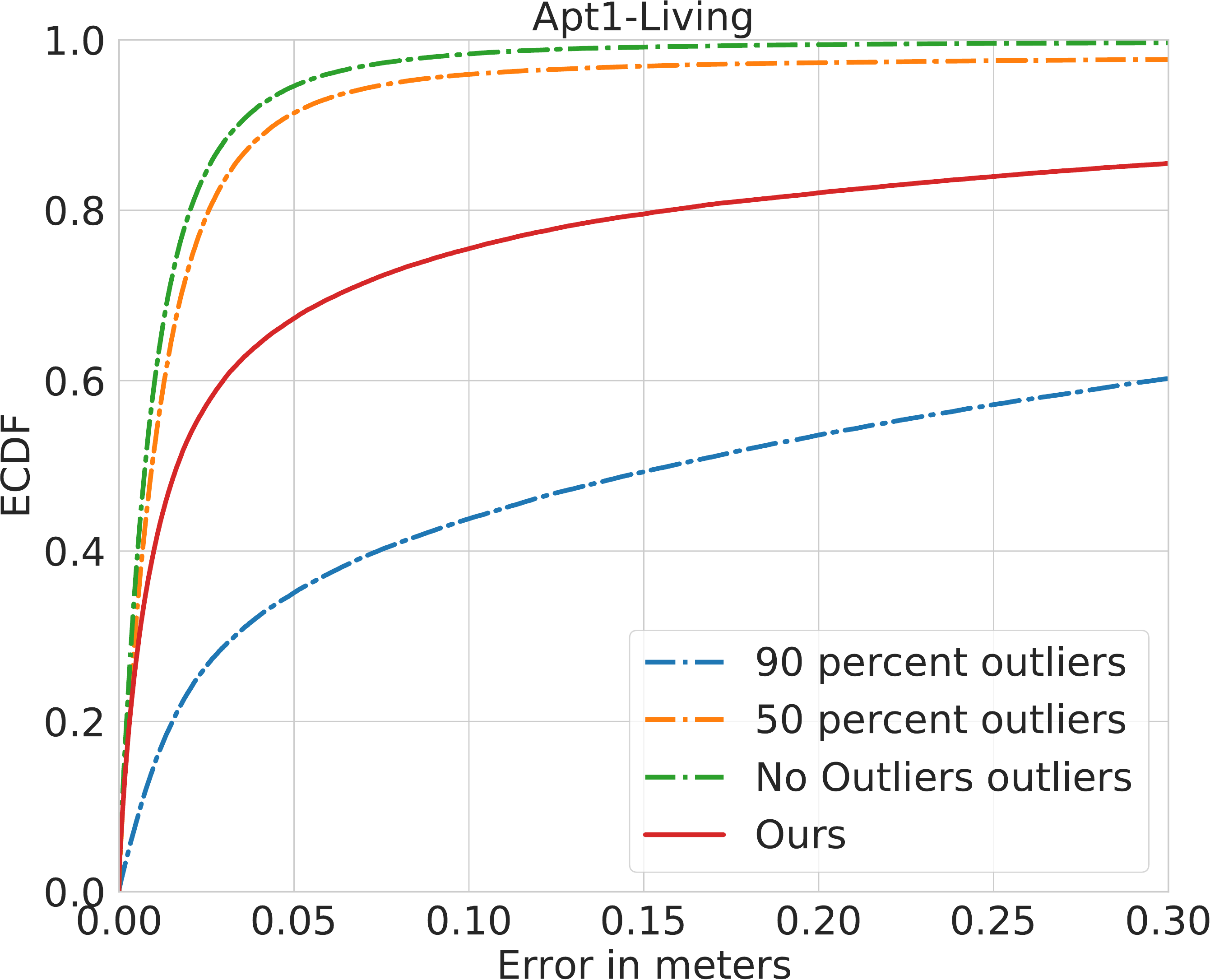}} \hspace{0.5pt}
    \subfloat{\includegraphics[width = 0.3\linewidth]{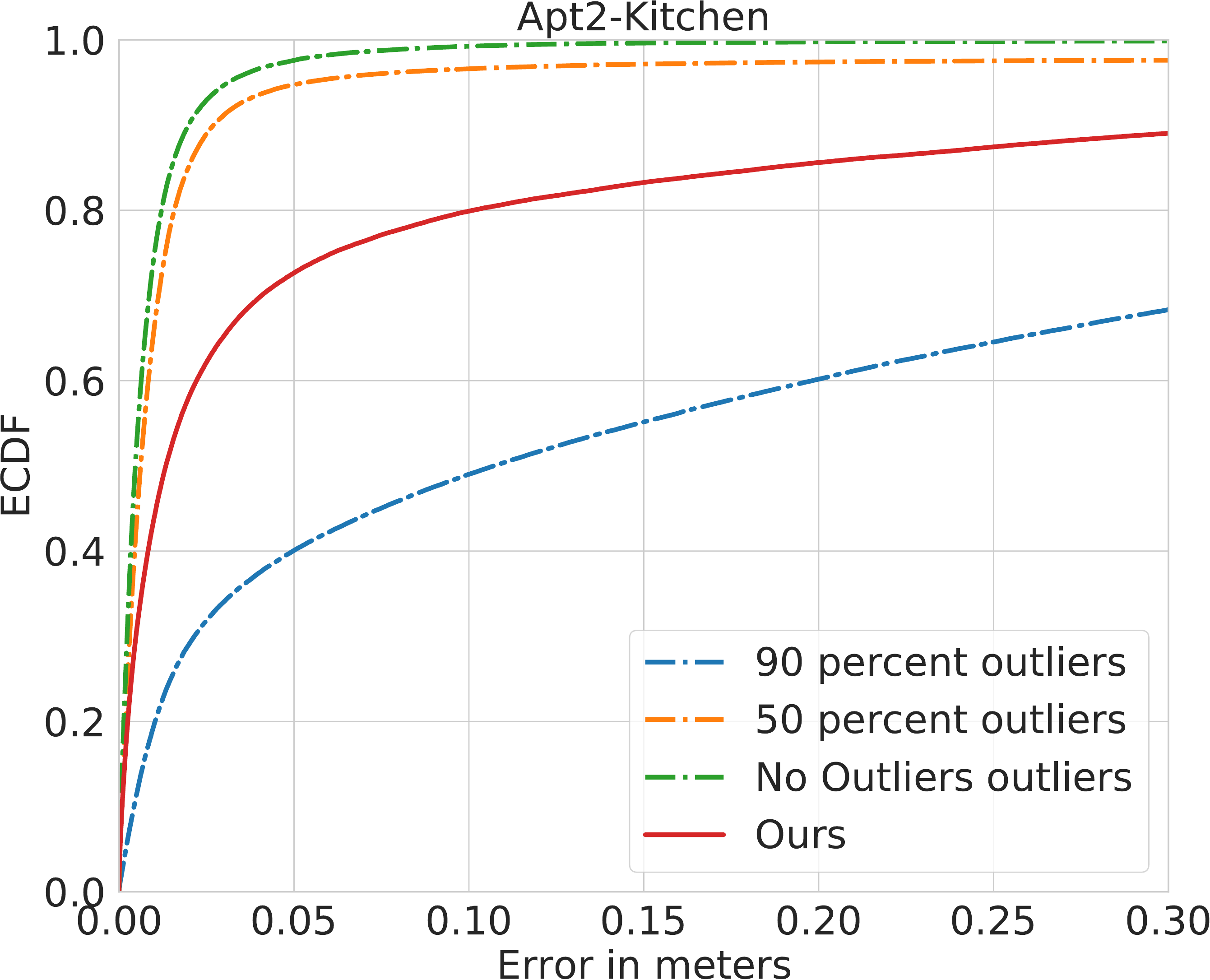}}
    \hspace{0.5pt}
    \centering
    \subfloat{\includegraphics[width = 0.3\linewidth]{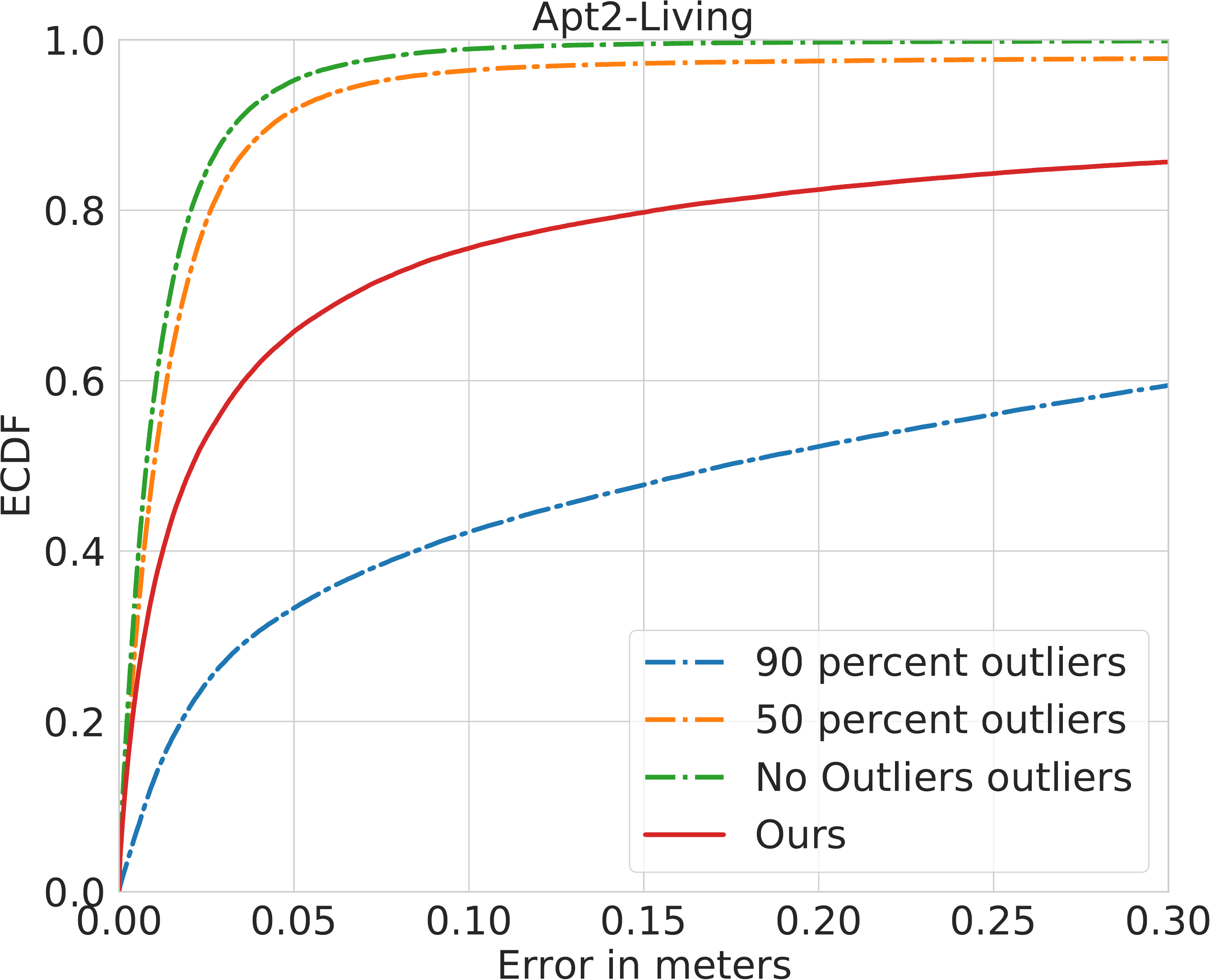}}
    \hspace{0.5pt}
    \centering
    \subfloat{\includegraphics[width = 0.3\linewidth]{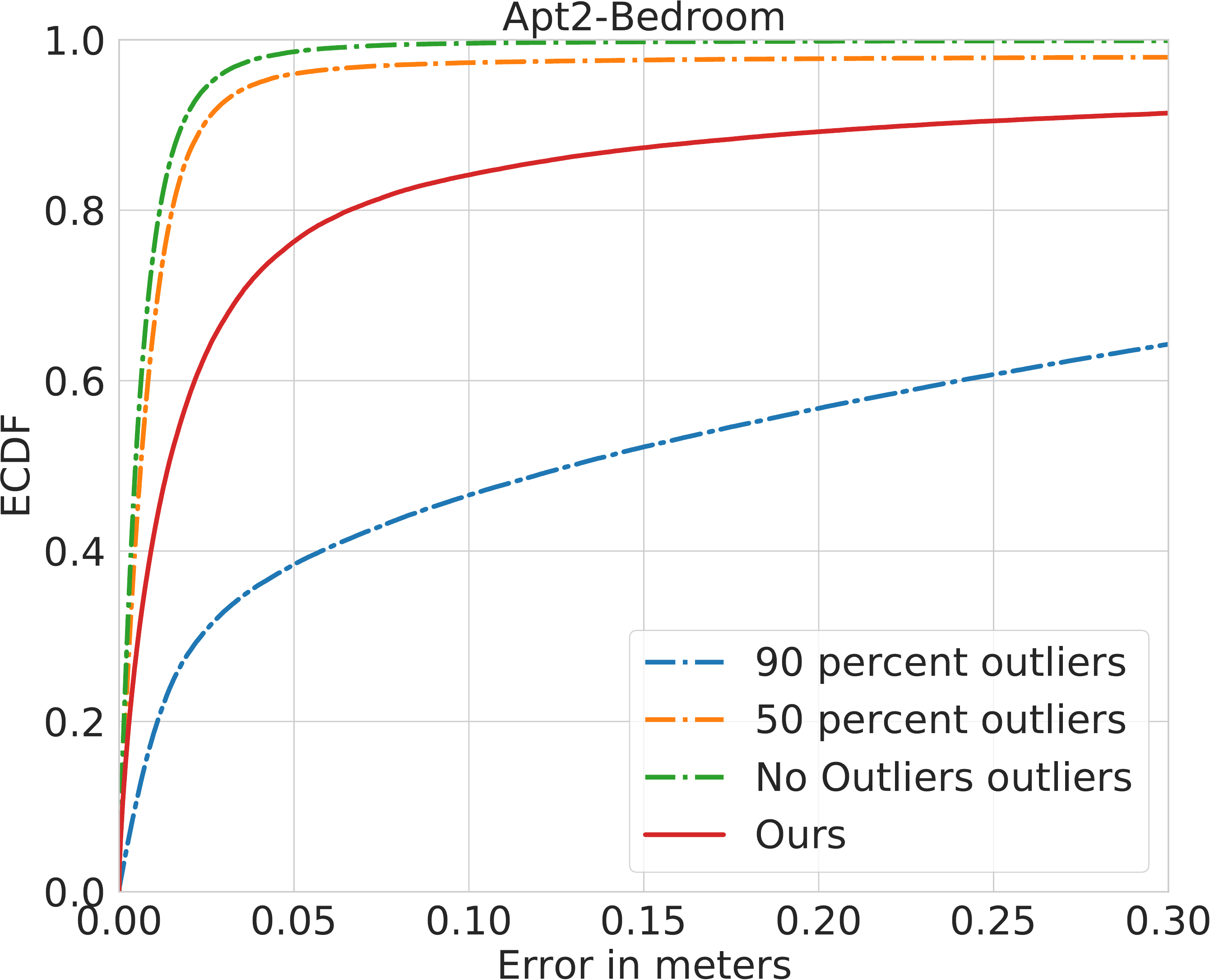}}
    \hspace{0.5pt}
    \centering
    \subfloat{\includegraphics[width = 0.3\linewidth]{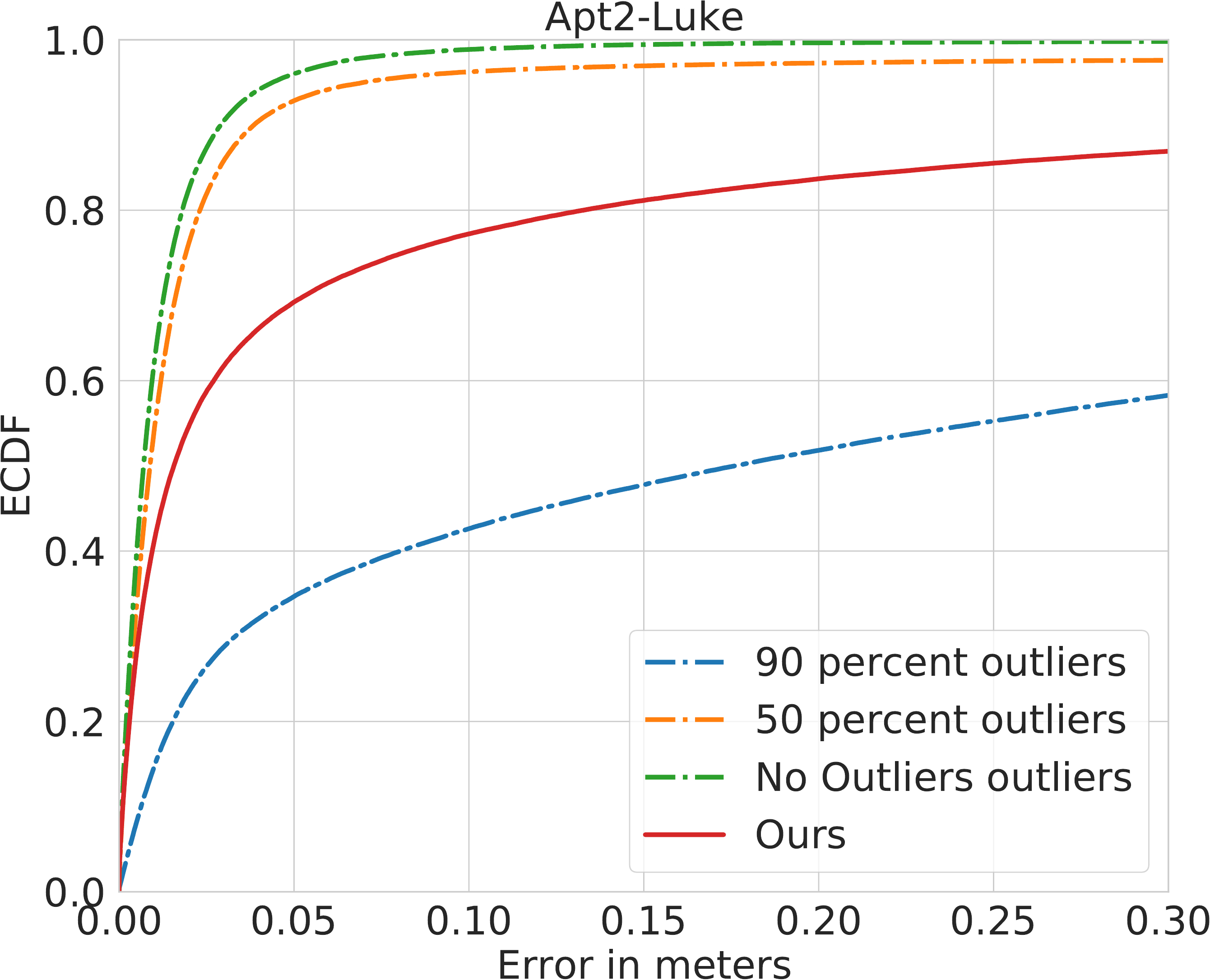}}
    \hspace{0.5pt}
    \centering
    \subfloat{\includegraphics[width = 0.3\linewidth]{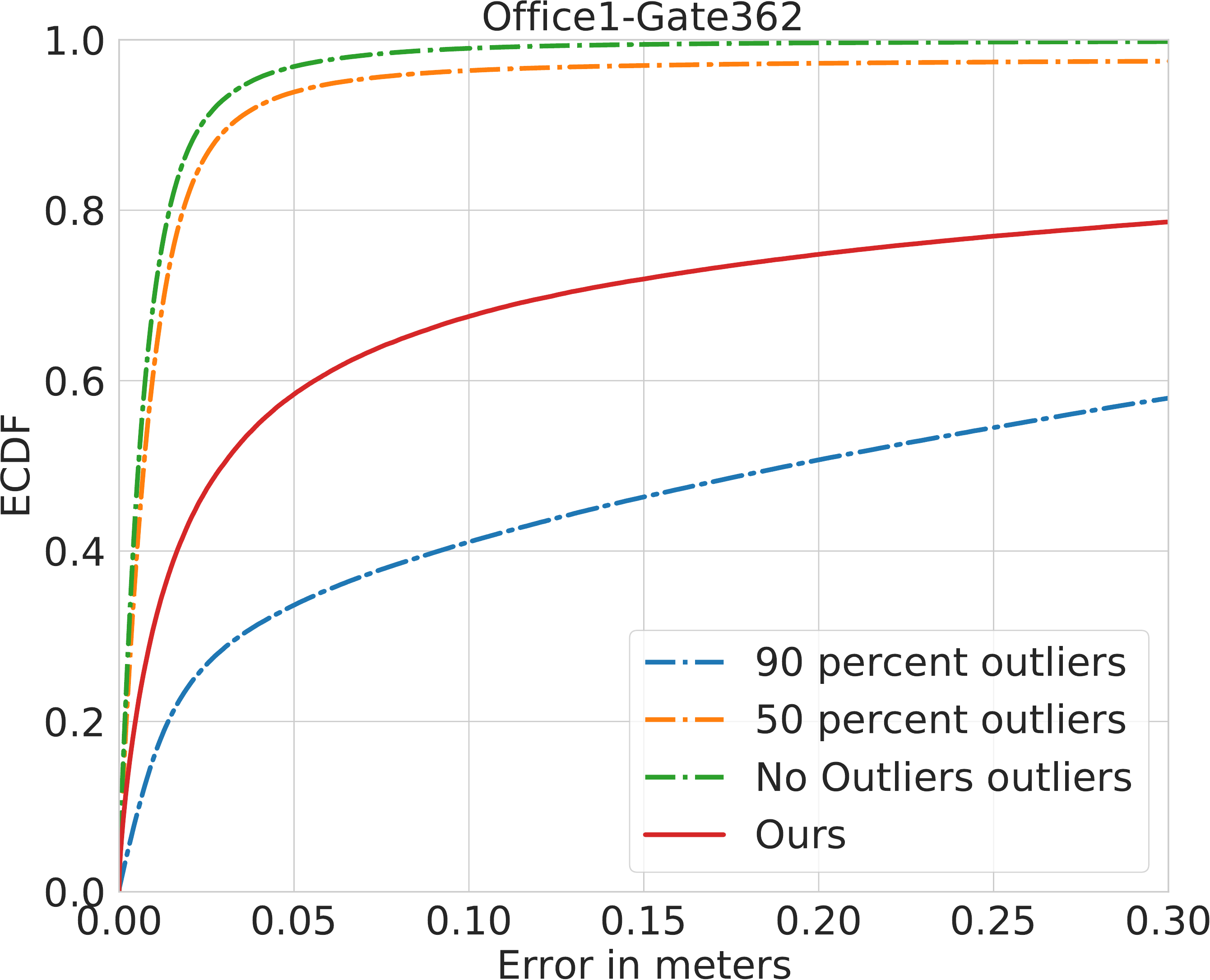}}
    \hspace{0.5pt}
    \centering
    \subfloat{\includegraphics[width = 0.3\linewidth]{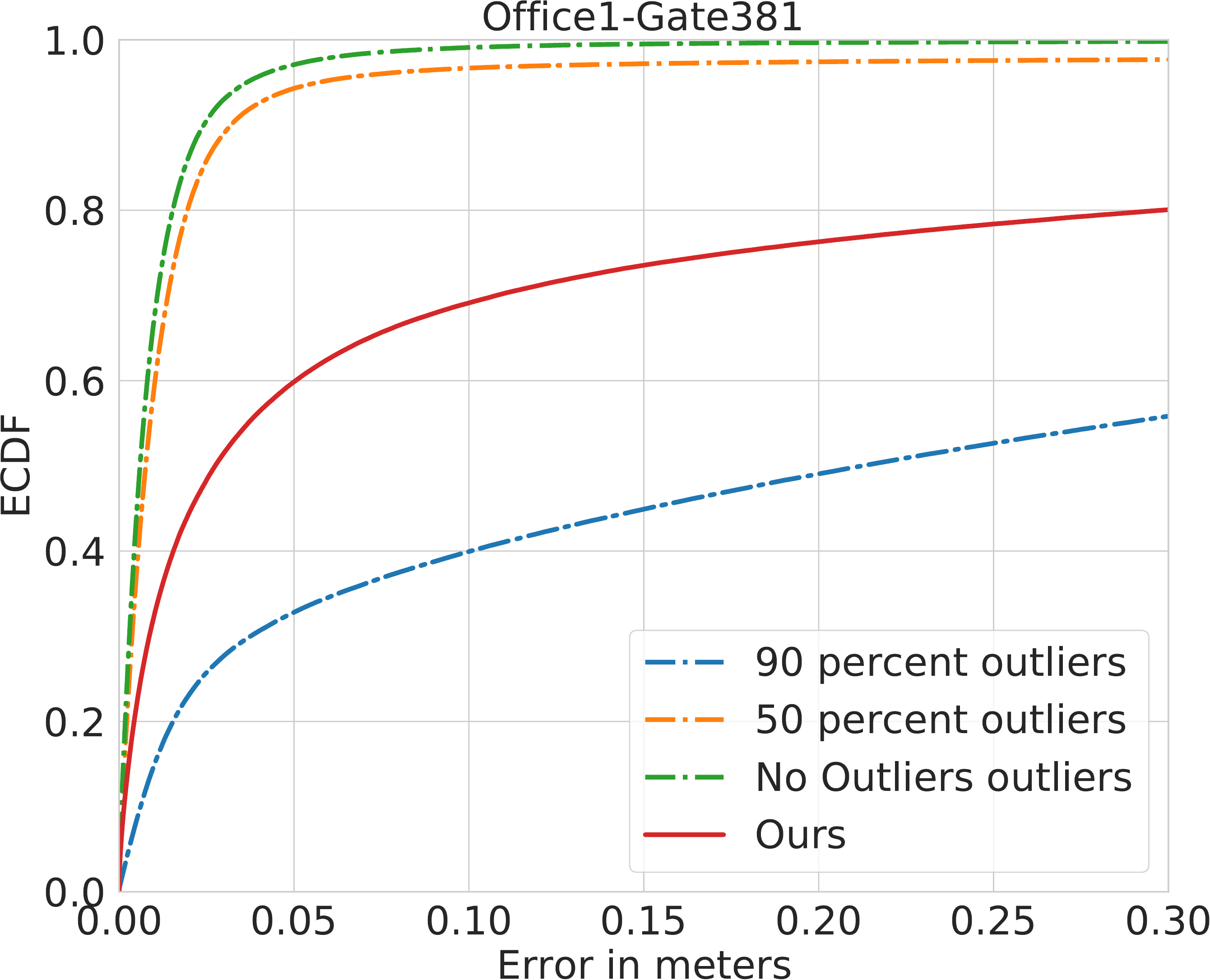}}
    \hspace{0.5pt}
    \centering
    \subfloat{\includegraphics[width = 0.3\linewidth]{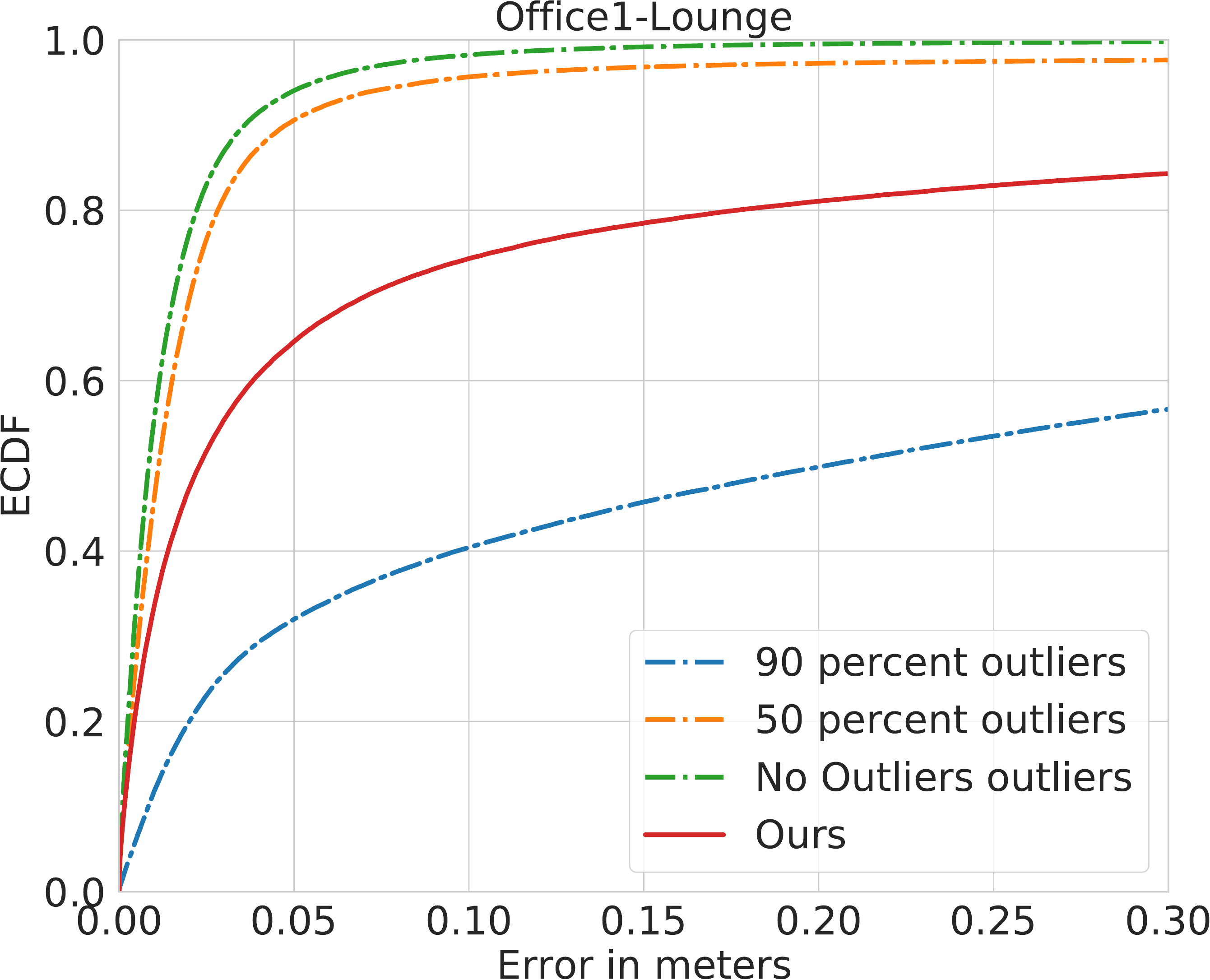}}
    \hspace{0.5pt}
    \centering
    \subfloat{\includegraphics[width = 0.3\linewidth]{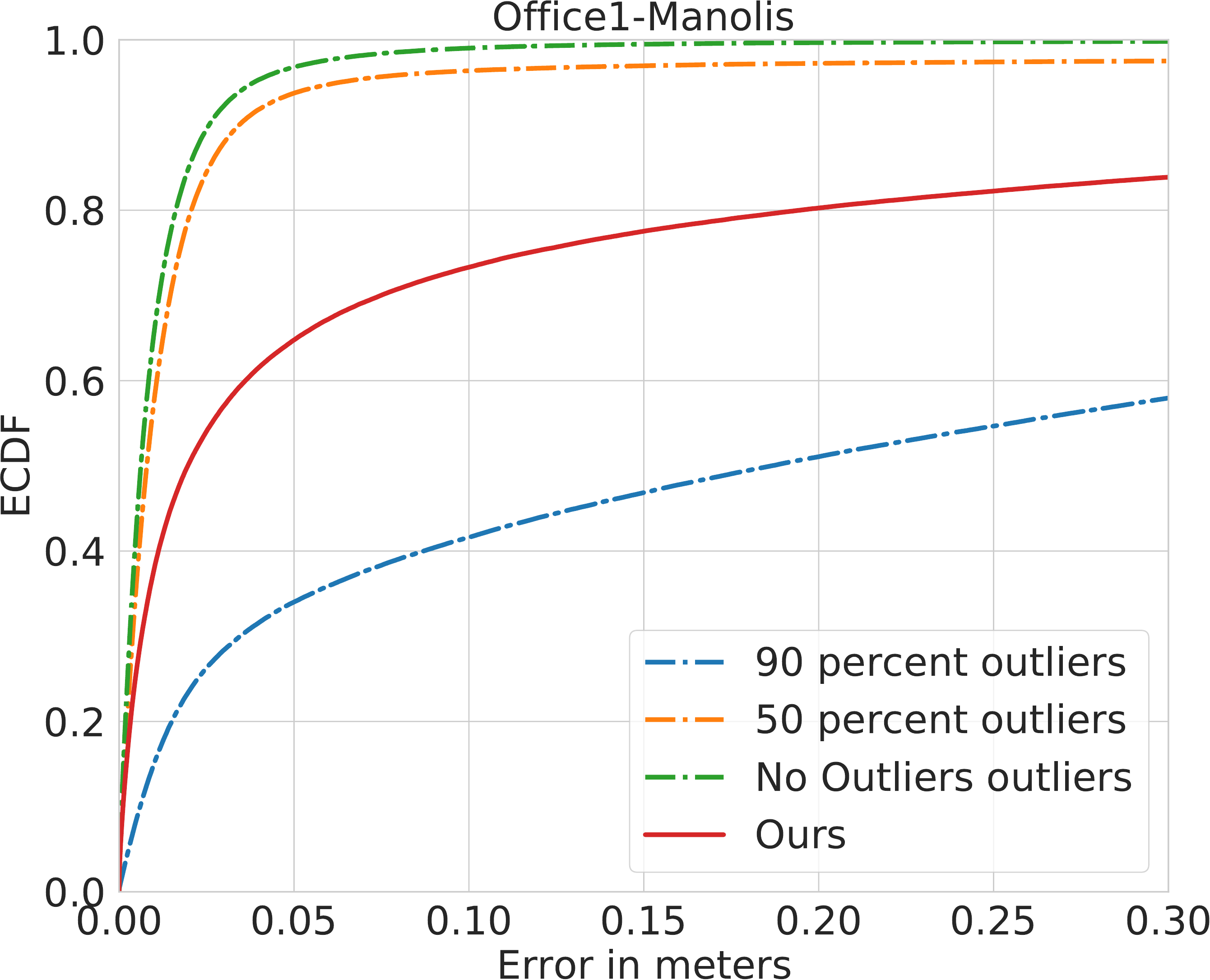}}
    \hspace{0.5pt}
    \centering
    \subfloat{\includegraphics[width = 0.3\linewidth]{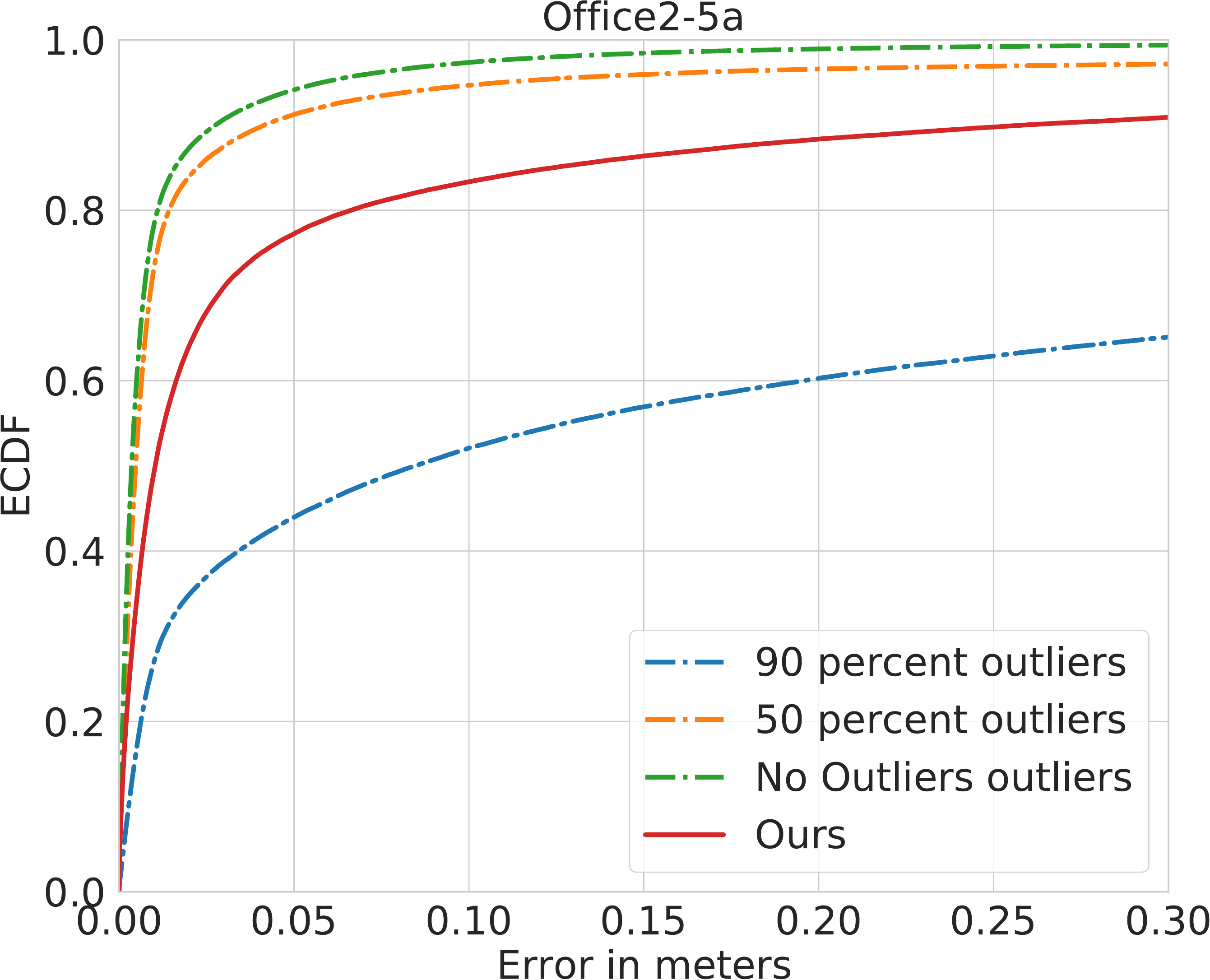}}
    \hspace{0.5pt}
    \centering
    \subfloat{\includegraphics[width = 0.3\linewidth]{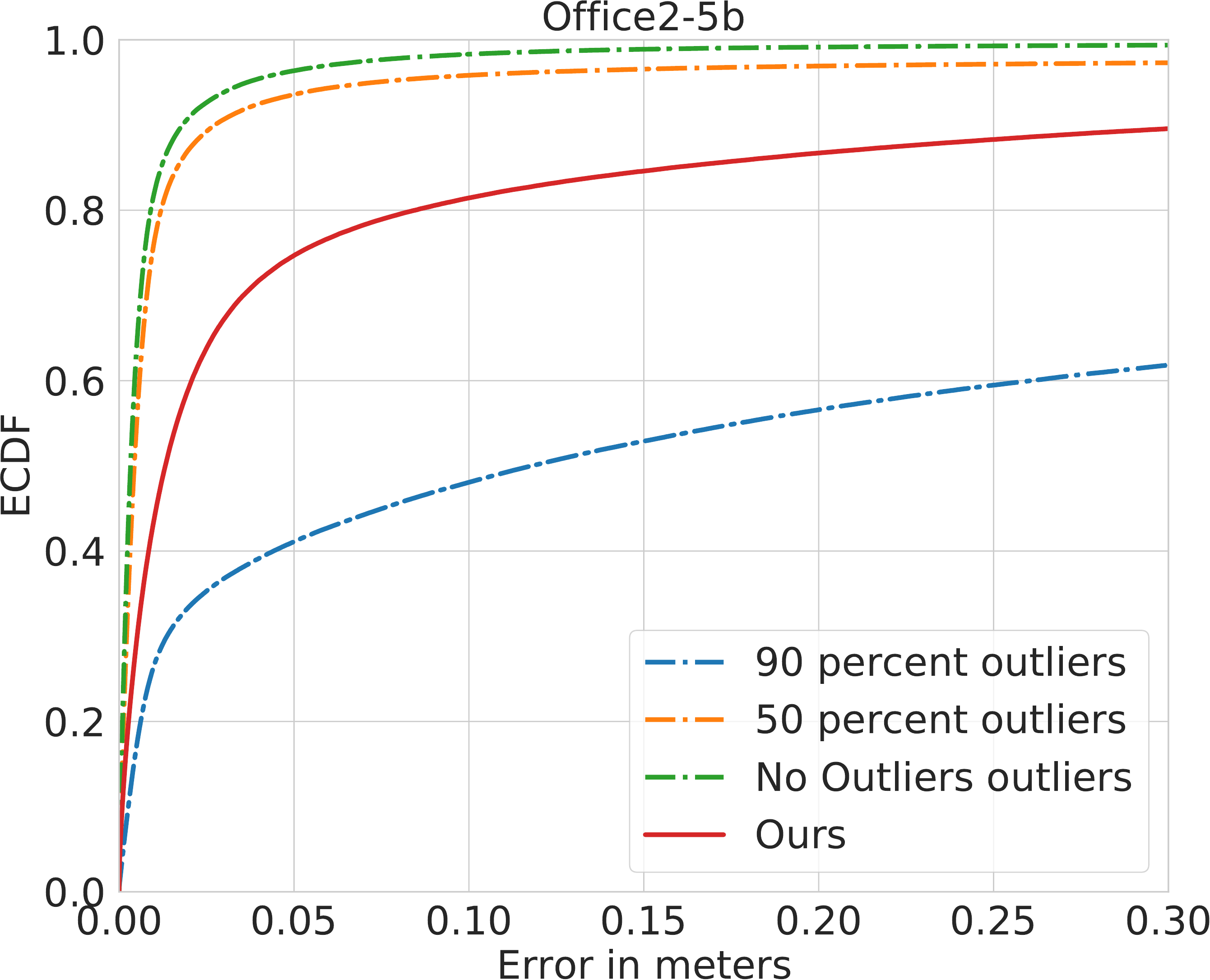}}
    \caption{Quantitative results showing the cumulative distribution of errors in the recovering point positions. We show results obtained on all scenes of the 12 Scenes dataset~\cite{Shotton2013CVPR}, for our approach and when the true neighborhood (of size 50) of each point / line is provided by an oracle. For the latter, we vary the level of contamination by outliers.}%Error distribution of our results and for oracle-provided neighborhoods of different sizes over all scenes from the indoor 12 Scenes dataset~\cite{Shotton2013CVPR}.}
    \label{fig:outliers_indoors_quantitative}
    
\end{figure*}

\begin{figure*}
    \centering
    \includegraphics[width=0.99\linewidth]{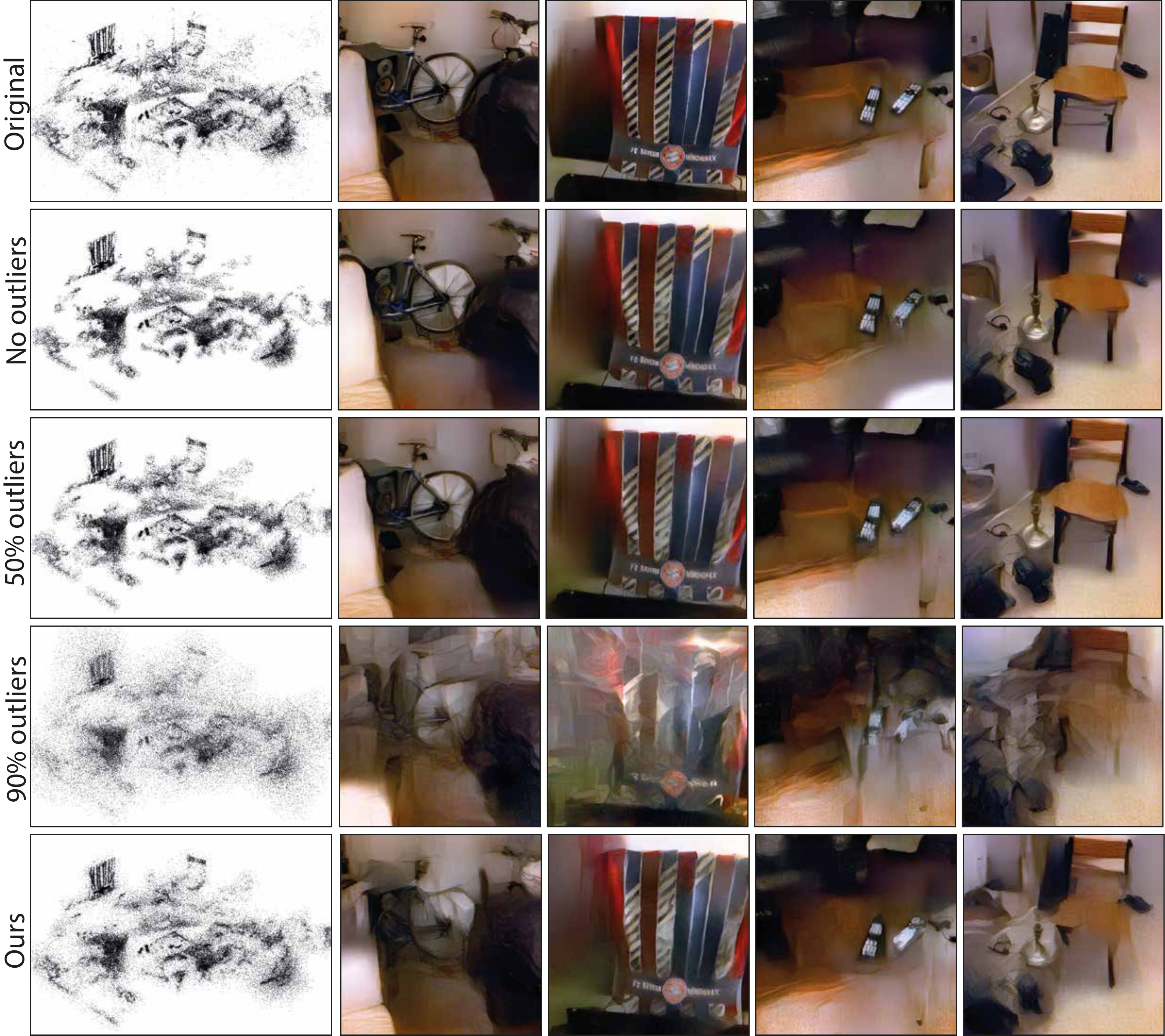}
    \caption{Qualitative results for the 'Apt2-Living' scene from the 12 Scenes dataset~\cite{Shotton2013CVPR}. We show point clouds as well as the images recovered from them using the approach from~\cite{Pittaluga2019CVPR}. Besides the original point cloud and the one recovered by our method, we also show results obtained by using an oracle to provide the neighborhood of each point / line. For these neighborhoods, we vary the percentage of outliers contained in them.} % the impact of outliers in the neighborhood used for estimation of point positions. In rows 2,3,4, it is assumed that the underlying neighborhoods of points are provided by an oracle. Row 5 shows the results from our proposed algorithm.}
    \label{fig:outlier_apt2_living}
\end{figure*}

\begin{figure*}
    \centering
    \includegraphics[width=0.99\linewidth]{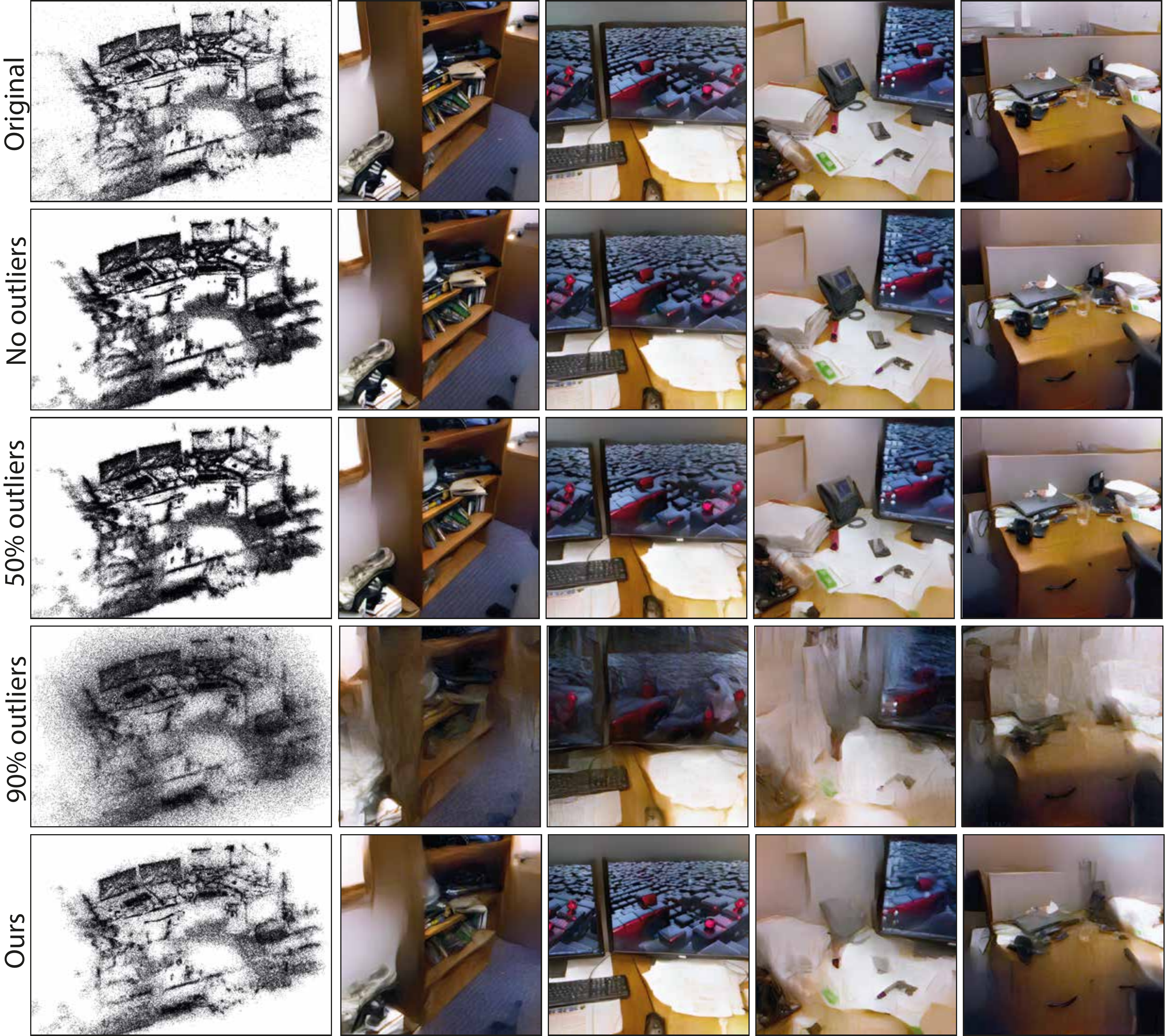}
    \caption{Qualitative results for the 'Office1-Gate362' scene from the 12 Scenes dataset~\cite{Shotton2013CVPR}. We show point clouds as well as the images recovered from them using the approach from~\cite{Pittaluga2019CVPR}. Besides the original point cloud and the one recovered by our method, we also show results obtained by using an oracle to provide the neighborhood of each point / line. For these neighborhoods, we vary the percentage of outliers contained in them.}%{Qualitative results over scene 'Offce1-Gate362' from 12 scenes. Analysing the impact of outliers in the neighborhood used for estimation of point positions. In rows 2,3,4, it is assumed that the underlying neighborhoods of points are provided by an oracle. Row 5 shows the results from our proposed algorithm.}
    \label{fig:outlier_office1_gate362}
\end{figure*}

\begin{figure*}
    \centering
    \includegraphics[width=0.99\linewidth]{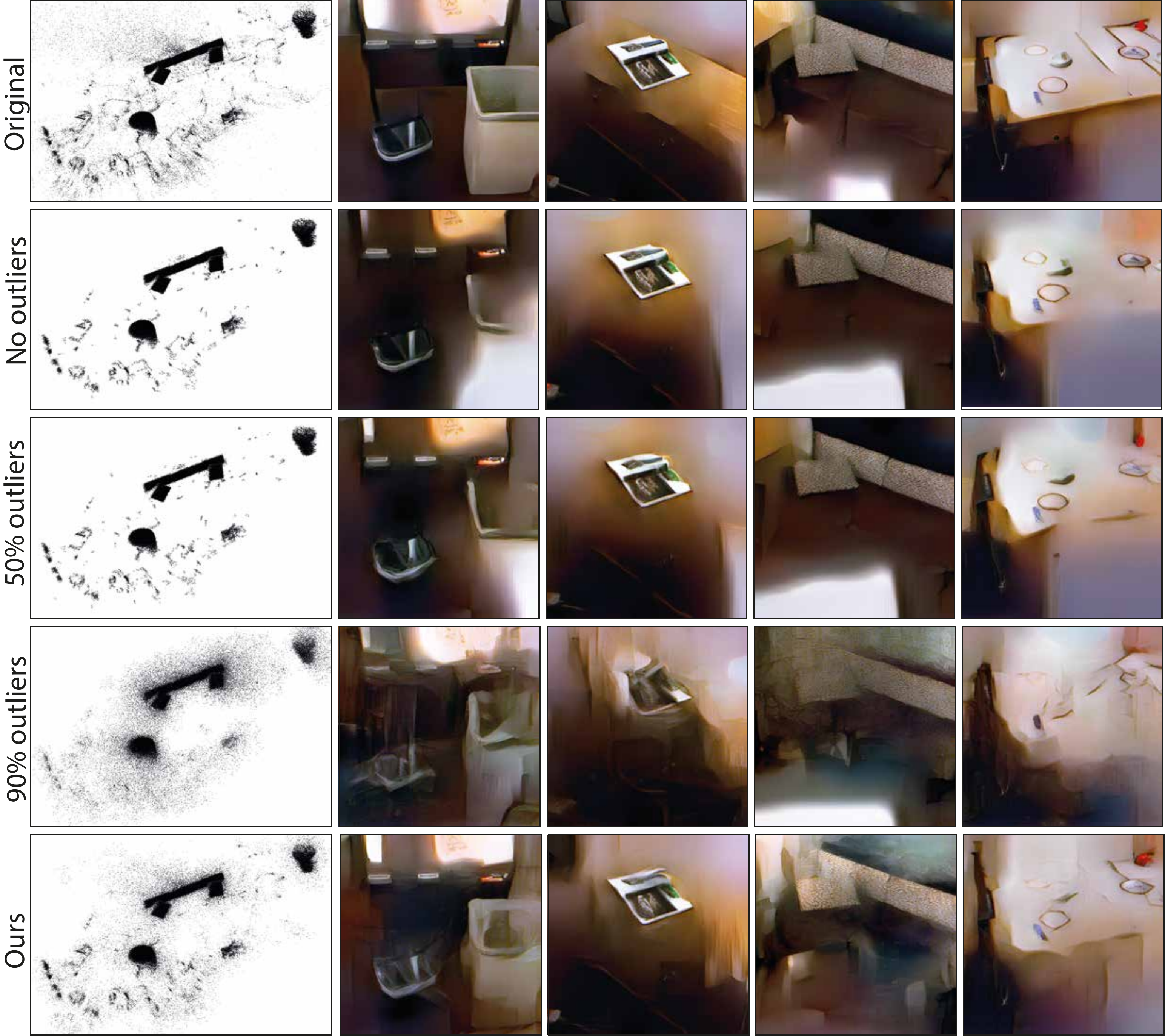}
    \caption{Qualitative results for the 'Office2-5a' scene from the 12 Scenes dataset~\cite{Shotton2013CVPR}. We show point clouds as well as the images recovered from them using the approach from~\cite{Pittaluga2019CVPR}. Besides the original point cloud and the one recovered by our method, we also show results obtained by using an oracle to provide the neighborhood of each point / line. For these neighborhoods, we vary the percentage of outliers contained in them.}%{Qualitative results over scene 'Office2-5a' from 12 scenes. Analysing the impact of outliers in the neighborhood used for estimation of point positions. In rows 2,3,4, it is assumed that the underlying neighborhoods of points are provided by an oracle. Row 5 shows the results from our proposed algorithm.}
    \label{fig:outlier_office2_5a}
\end{figure*}

\begin{figure*}
    \centering
    \includegraphics[width=0.99\linewidth]{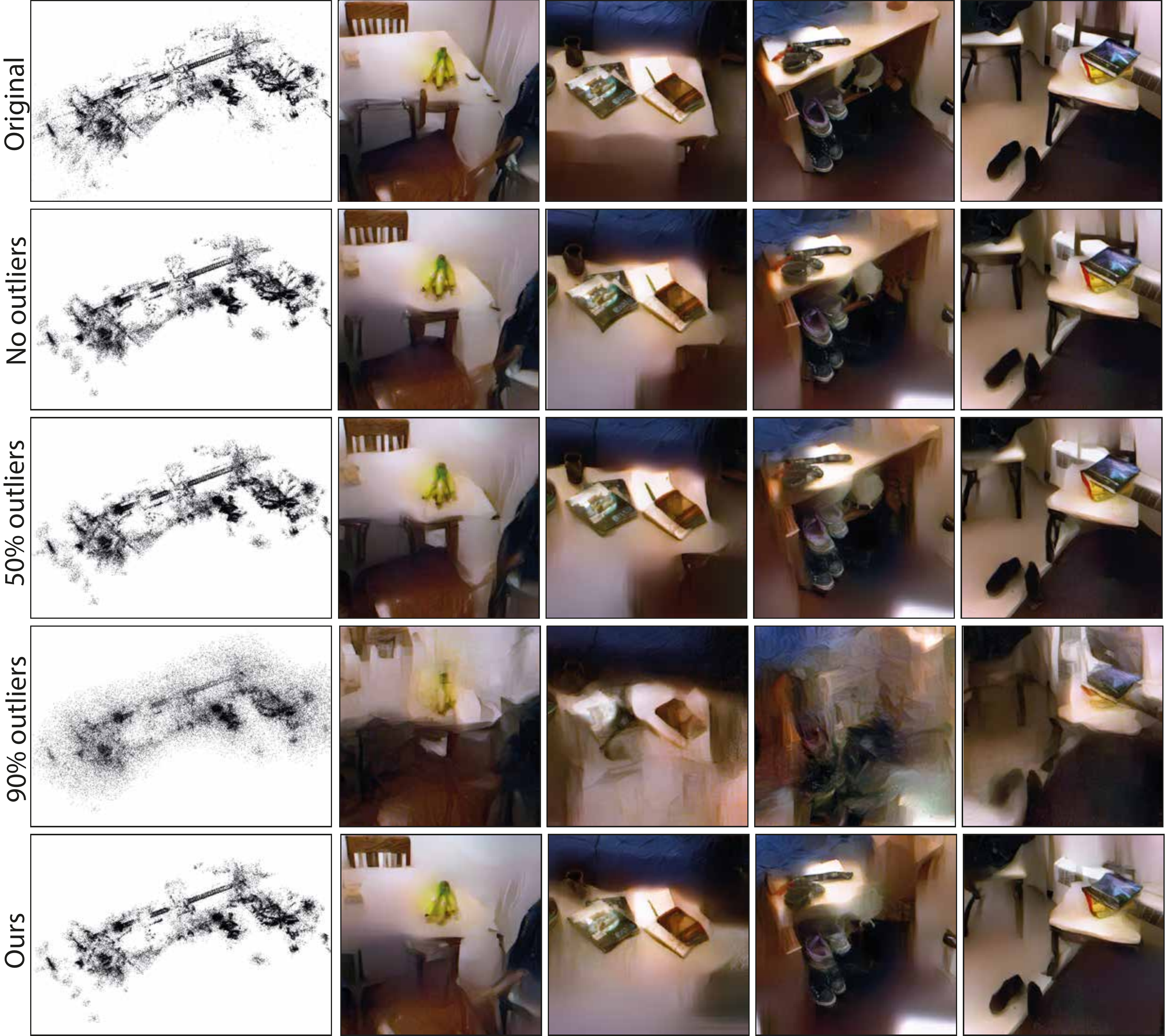}
    \caption{Qualitative results for the 'Apt1-Living' scene from the 12 Scenes dataset~\cite{Shotton2013CVPR}. We show point clouds as well as the images recovered from them using the approach from~\cite{Pittaluga2019CVPR}. Besides the original point cloud and the one recovered by our method, we also show results obtained by using an oracle to provide the neighborhood of each point / line. For these neighborhoods, we vary the percentage of outliers contained in them.}
    \label{fig:outlier_apt1_kitchen}
\end{figure*}

\begin{figure*}
    \centering
    \includegraphics[width=0.99\linewidth]{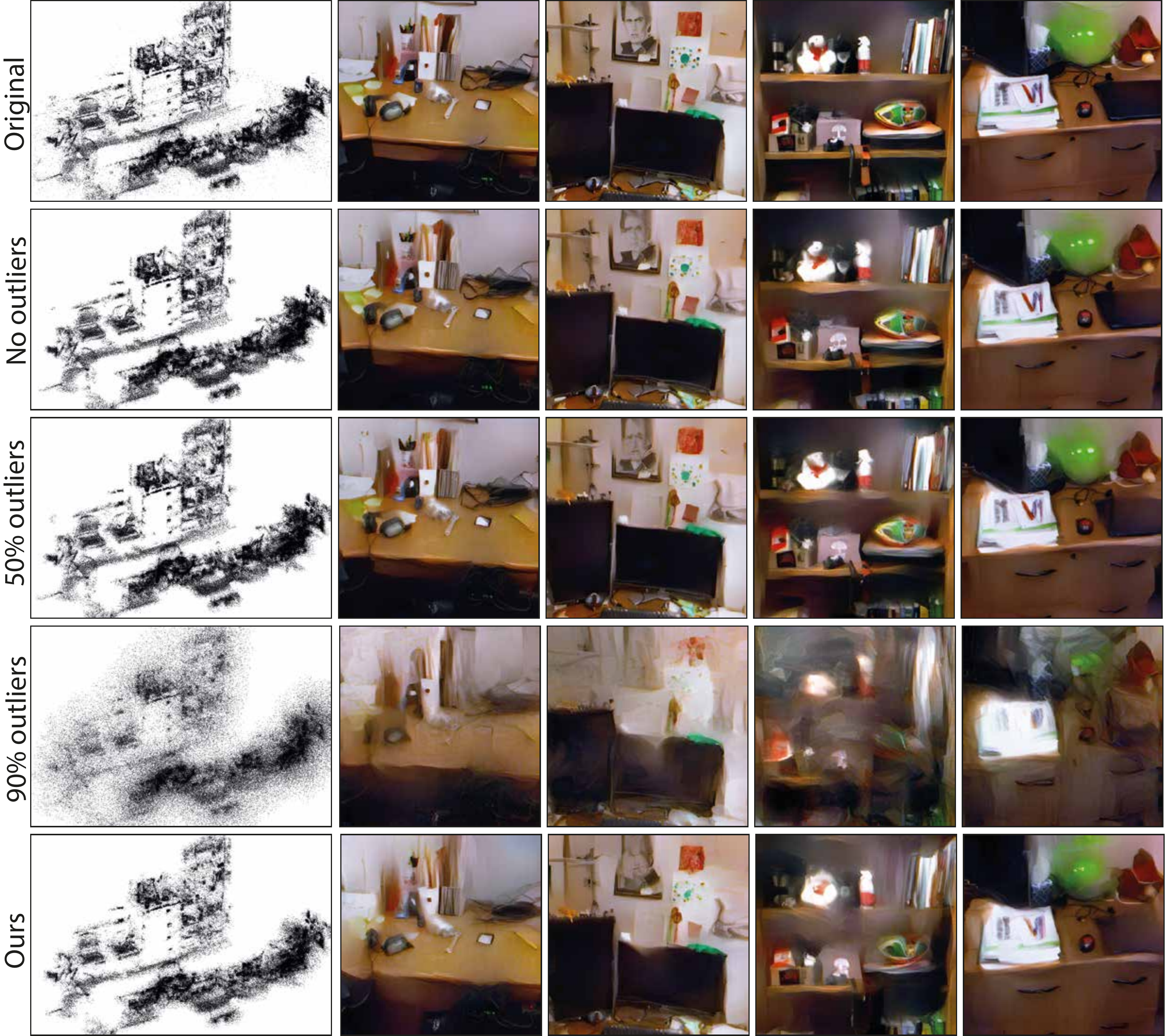}
    \caption{Qualitative results for the 'Office1-Manolis' scene from the 12 Scenes dataset~\cite{Shotton2013CVPR}. We show point clouds as well as the images recovered from them using the approach from~\cite{Pittaluga2019CVPR}. Besides the original point cloud and the one recovered by our method, we also show results obtained by using an oracle to provide the neighborhood of each point / line. For these neighborhoods, we vary the percentage of outliers contained in them.}
    \label{fig:outlier_office1_manolis}
\end{figure*}

\begin{figure*}
    \centering
    \includegraphics[width=0.99\linewidth]{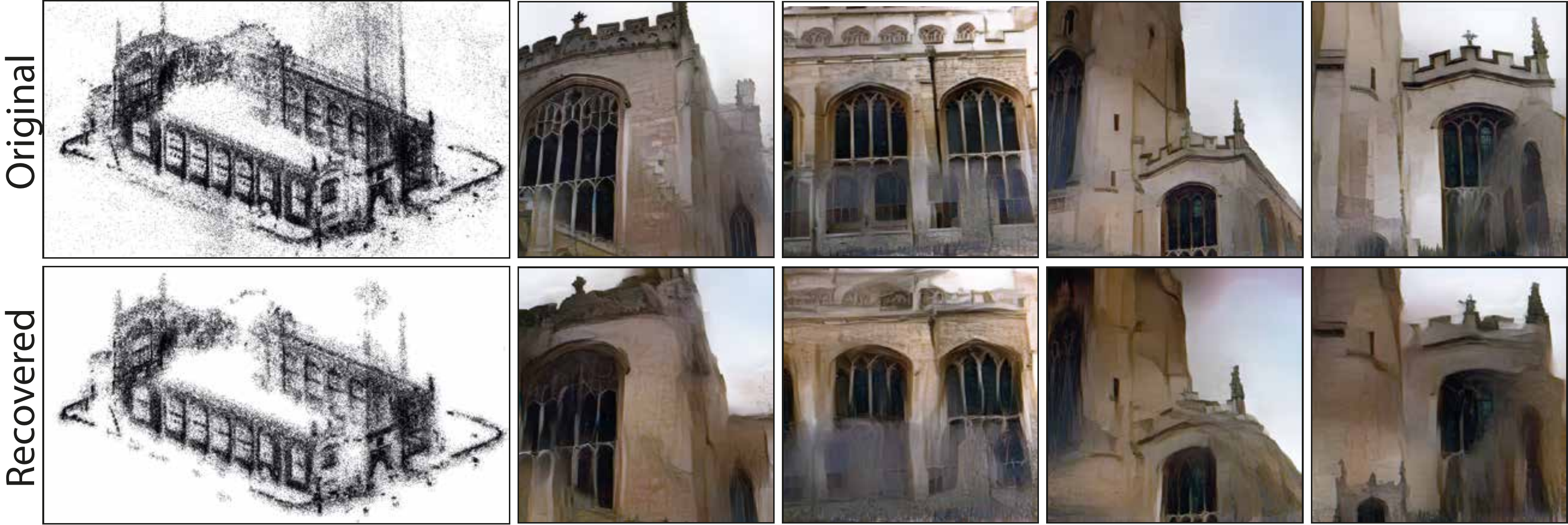}
    \caption{Qualitative results showing the recovered point cloud and the images obtained by applying the inversion method of \cite{Pittaluga2019CVPR} for the 'St Mary's Church' scene from the Cambridge dataset ~\cite{Kendall2015ICCV}. }
    \label{fig:st_marys_qualitative}
\end{figure*}

\begin{figure*}
    \centering
    \includegraphics[width=0.99\linewidth]{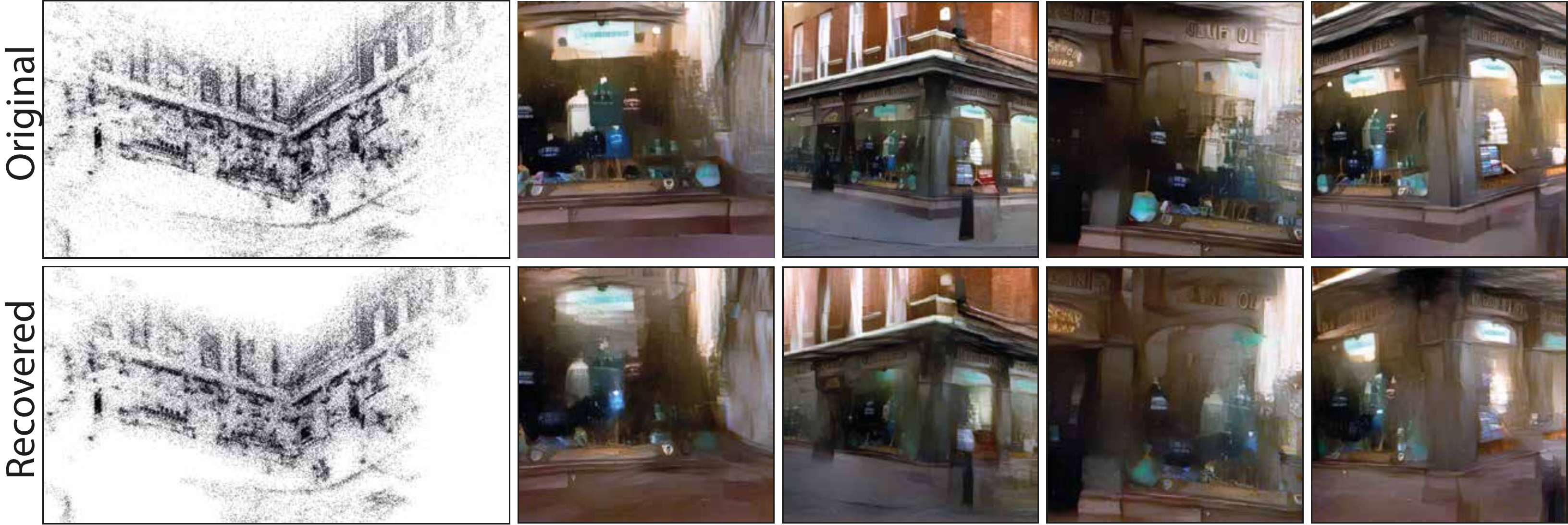}
    \caption{Qualitative results showing the recovered point cloud and the images obtained by applying the inversion method of \cite{Pittaluga2019CVPR} for the 'Shop Facade' scene from the Cambridge dataset ~\cite{Kendall2015ICCV}.}
    \label{fig:shop_facade_qualitative}
\end{figure*}

\PAR{More results.}
In contrast to Fig.~\ref{fig:num_nn_indoors_apt2kitchen}, where we vary the number of true neighbors, we now fix the number of neighbors provided by the oracle to 50. 
As for Fig.~3 in the main paper, we instead vary the percentage of outliers among the neighbors by randomly replacing a fraction of the true neighbors with randomly selected points / lines. 
Fig.~\ref{fig:outliers_indoors_quantitative} quantitatively compares the cumulative distribution of errors in different scenarios for all scenes of the 12 scenes dataset. Figures~\ref{fig:outlier_apt2_living} to~\ref{fig:outlier_office1_manolis} show additional qualitative results for the recovered point clouds and the images obtained from them. 
Figures~\ref{fig:st_marys_qualitative} and ~\ref{fig:shop_facade_qualitative} further show some qualitative results of our method for two of the outdoor scenes considered in our paper.
As can be seen, our approach is able to faithfully recover the 3D point clouds and obtain detailed images as long as the neighborhoods do not contain too many unrelated points / lines. 
This holds both for the neighborhoods provided by the oracle and those estimated by our method. 
In particular, the results show that it is possible to recover image details via the point clouds estimated by our approach. 
As in the main paper, we thus conclude that lifting point clouds to line clouds alone does not guarantee that image details cannot be recovered.

{\small
\bibliographystyle{splncs04}
\bibliography{egbib,torsten}
}

\end{document}